\documentclass[numbers,10pt]{arxiv}

\usepackage[utf8]{inputenc}

\usepackage{tabularx}
\usepackage{nicefrac}             
\usepackage{graphicx, subcaption} 
\usepackage{enumitem}             
\usepackage{array, tabularx}      
\newcolumntype{L}{>{\raggedright\arraybackslash}X} 
\usepackage{fvextra}              
\usepackage[most]{tcolorbox}      
\usepackage{xspace}               
\usepackage{bxcoloremoji}
\usepackage{graphicx}
\usepackage{enumitem}
\usepackage{caption}
\usepackage{xspace}
\usepackage{fvextra}
\usepackage{cleveref}
\usepackage{float} 
\usepackage{subcaption}
\usepackage{booktabs}
\usepackage{adjustbox}
\usepackage{multirow}
\usepackage{siunitx} 

\makeatletter
\DeclareRobustCommand\onedot{\futurelet\@let@token\@onedot}
\def\@onedot{\ifx\@let@token.\else.\null\fi\xspace}

\makeatother

\renewcommand{\paragraph}[1]{\textbf{#1}}

\usepackage{xcolor}
\usepackage{amsmath}
\usepackage{wrapfig}

\definecolor{promptblue}{RGB}{25, 84, 166}
\definecolor{thinkgreen}{RGB}{34, 139, 34}
\definecolor{thinktag}{RGB}{110, 65, 30}  
\definecolor{thinkbg}{RGB}{252, 248, 243}  
\definecolor{boxbg}{RGB}{252, 252, 252}

\tcbset{
    mainbox/.style={
        colback=boxbg,
        colframe=gray!40,
        arc=2mm,
        boxrule=0.8pt,
        left=12pt,
        right=12pt,
        top=12pt,
        bottom=12pt,
        fonttitle=\bfseries\large,
        coltitle=white,
        colbacktitle=gray!60
    }
}

\usepackage[T1]{fontenc}
\usepackage{amsfonts}
\usepackage{amsthm}
\usepackage{amsmath}
\usepackage{nicefrac}
\usepackage{url}
\usepackage{array}
\usepackage{fontawesome5}
\usepackage{amssymb}

\renewcommand{\paragraph}[1]{\sansbf{#1}}

\newcommand{\nyu}{\ensuremath{\textcolor[HTML]{57068C}{\spadesuit}}}
\newcommand{\modal}{\ensuremath{\textcolor[HTML]{00A86B}{\scalebox{1.0}{$\blacktriangle$}}}}
\newcommand{\ucla}{\ensuremath{\textcolor[HTML]{2774AE}{\clubsuit}}}
\newcommand{\uiucu}{\ensuremath{\textcolor[HTML]{D56A1B}{\bigstar}}}
\newcommand{\columbia}{\ensuremath{\textcolor[HTML]{009EFF}{\blacklozenge}}}

\definecolor{takeawaycolor}{RGB}{70, 70, 180}
\definecolor{boxbg}{RGB}{240, 240, 250}
\definecolor{rlratio}{HTML}{2C5282}

\newtcolorbox{takeaway}{%
  enhanced,
  rounded corners,
  breakable,
  colback=boxbg,
  colframe=boxbg,
  boxrule=0pt,
  left=10pt,
  right=10pt,
  top=5pt,
  bottom=5pt,
  toptitle=1pt,
  bottomtitle=1pt,
  arc=12pt,
  before={\vspace{0pt}},
  after={\vspace{0pt}}
}

\definecolor{lawgold}{HTML}{C58A16}
\definecolor{lawcream}{HTML}{FFF9EC}
\definecolor{lawtitlebg}{HTML}{FBECC8}
\definecolor{lawdark}{HTML}{6F4A08}

\newtcolorbox{scalinglaw}{
    enhanced,
    breakable,
    colback=lawcream,
    colframe=lawgold,
    colbacktitle=lawtitlebg,
    coltitle=lawdark,
    fonttitle=\bfseries,
    title={Joint Pretraining--RL Scaling Law},
    boxrule=0.7pt,
    titlerule=0.45pt,
    arc=1.8mm,
    left=3.5mm,
    right=3.5mm,
    top=2mm,
    bottom=2mm,
    toptitle=1.4mm,
    bottomtitle=1.4mm,
    before skip=8pt,
    after skip=8pt,
    drop fuzzy shadow=black!8
}

\title{Understanding Reasoning from Pretraining to Post-Training}

\author[\nyu\,\star]{Jingyan Shen}
\author[\nyu\,\modal\,\star]{Ang Li}
\author[\nyu\,\ucla]{Salman Rahman}
\author[\uiucu]{Yifan Sun}
\author[\columbia]{\\Micah Goldblum}
\author[\nyu]{Matus Telgarsky}
\author[\nyu]{Pavel Izmailov}

\contribution[\ensuremath{\star}]{Equal contribution, correspondence to \texttt{jingyan.s@nyu.edu, al6843@nyu.edu, pi390@nyu.edu}}

\affiliation[\nyu]{\textcolor[HTML]{57068C}{New York University}}
\affiliation[\modal]{\textcolor[HTML]{00A86B}{Modal Labs}}
\affiliation[\ucla]{\textcolor[HTML]{2774AE}{University of California, Los Angeles\quad\quad\quad\quad}}
\affiliation[\uiucu]{\textcolor[HTML]{D56A1B}{University of Illinois Urbana-Champaign}}
\affiliation[\columbia]{\textcolor[HTML]{009EFF}{Columbia University}}
\abstract{Reinforcement learning (RL) has become central to improving large language models (LLMs) on complex reasoning tasks, yet RL post-training is largely studied in isolation from the pretraining that precedes it. As a result, two basic questions remain open: (1) how do pretraining choices (model size, data) shape the returns to RL compute, and (2) what does RL actually do to the model? These questions are difficult to study in the standard LLM setting: pretraining corpora are vast and uncontrolled, making it hard to attribute behaviors to pretraining versus RL, and systematic compute sweeps across both stages are prohibitively expensive. 
To address these challenges, we use chess as a controlled testbed for studying reasoning across the full pretraining-to-post-training pipeline. 
We follow the standard LLM training pipeline by pretraining language models from 5M to 1B parameters on human chess games, supervised fine-tuning on synthetic reasoning traces, and running RL on chess puzzles with verifiable rewards. Using this framework, we establish a scaling law connecting pretraining and RL: the post-RL performance at given RL compute level is well-predicted from the pretraining loss, and slope of the RL reward curves improves approximately linearly with the pretraining tokens. 
Beyond scaling, we find that RL does not simply sharpen the SFT policy: on easy puzzles it amplifies correct moves the SFT policy already preferred, while on hard puzzles it surfaces correct moves that were nearly absent under SFT. 
We further test whether our findings beyond chess by training a 1B language model on math domain, where the same predictive pattern emerges: longer-pretrained checkpoints reach higher post-RL performance and improve faster under RL.
In sum, we provide a quantitative account of the pretraining-to-RL interface and a controlled testbed for studying the science of reasoning across the full pretraining-to-post-training pipeline.

\coloremojicode{1F917} \textbf{Models \& Datasets}: \href{https://huggingface.co/collections/pavelslab-nyu/pre2post-chess}{huggingface.co/pavelslab-nyu/pre2post-chess} \\
\faGithub\ \textbf{Code}: \href{https://github.com/pavelslab-nyu/pre2post-chess}{github.com/pavelslab-nyu/pre2post-chess}
}

\theoremstyle{plain}

\theoremstyle{definition}

\theoremstyle{remark}

\begin{document}
\maketitle

\section{Introduction}
\label{sec:intro_new}

The standard pipeline for training large language models (LLMs) consists of large-scale pretraining followed by post-training, typically supervised fine-tuning (SFT) and reinforcement learning (RL) with verifiable rewards~\citep{guo2025deepseek, lambert2024tulu, yu2025dapo, zeng2025simplerl}.
As LLMs continue to scale, two perspectives on where to
invest additional compute have begun to diverge. One emphasizes the pretrained prior: scaling model size, data,
and compute to produce stronger base models from human text~\citep{kaplan2020scaling, hoffmann2022training}.
The
other emphasizes experience: using RL to learn from environmental interaction and outcome-based feedback, thereby eliciting or developing capabilities beyond direct imitation. 
This view is reflected in arguments for an \textit{era of experience}~\citep{silver2025welcome}, and exemplified by AlphaZero, which famously removed the imitation cold-start used by earlier AlphaGo and learned stronger policies from self-play alone~\citep{silver2016mastering, silver2017mastering}.
For LLM reasoning, an experience-only approach is not yet practical as their action spaces are enormous and rewards for a randomly-initialized policy are initially extremely sparse.
Thus, RL is invariably initialized from a pretrained prior, so the relevant question is not whether to use a prior, but how good the prior needs to be. 
Concretely, how should a fixed compute budget be divided between improving the pretrained model and further optimizing it with RL?
Prior work has developed quantitative scaling laws for pretraining~\citep{kaplan2020scaling, hoffmann2022training} and has studied post-training RL scaling and recipes separately~\citep{olmo2025olmo, khatri2025art}, but there is no quantitative characterization of how pretraining interacts with RL scaling.

\begin{figure}[t]
    \centering
    \includegraphics[width=\linewidth]{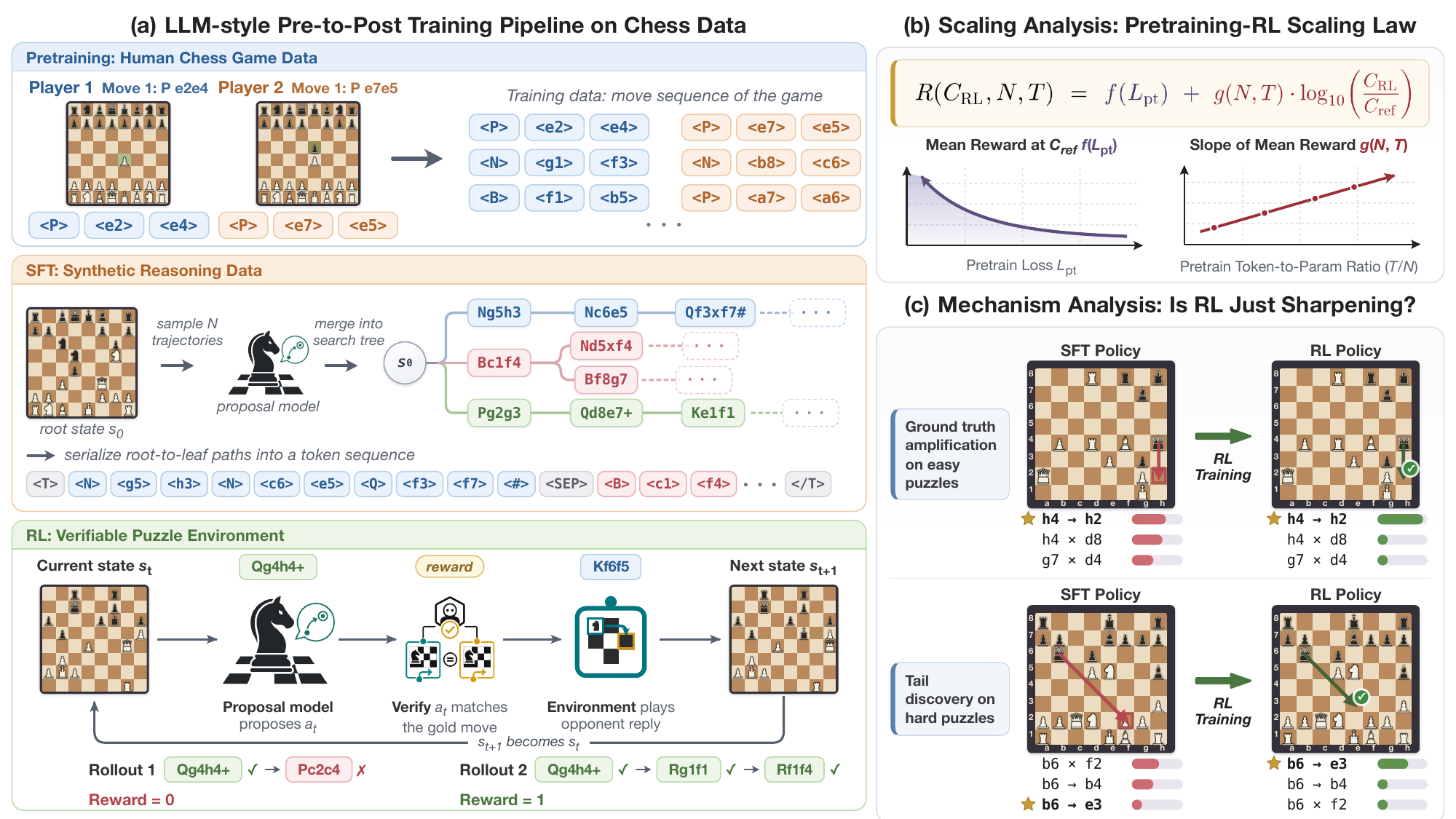}
    \caption{\textbf{Overview.} (a) We introduce a synthetic framework for studying reasoning from pretraining to post-training in the chess domain. (b) Using this framework, we establish a joint pretraining–RL scaling law, showing that pretraining performance provides predictive signal for RL performance under a fixed compute budget. (c) Through mechanistic analysis of policy evolution, we show that RL can surface moves that were nearly absent under the SFT policy.
    }
    \label{fig:overview}
\end{figure}

A related question concerns what RL actually does to the pretrained policy it inherits.
Recent work points to strikingly different views.
\citet{yue2025does} argues that RL primarily sharpens reasoning patterns the base model already prefers, observing that base models match or exceed RL-tuned ones in pass@$k$ when $k$ is large.
\citet{yuan2025f} argues the opposite, showing that RL composes pretrained skills into new ones.
\citet{sun2025rl} reports both behaviors, with ``grokking'' transitions on some problems and outright failures on others. These competing views have direct implications for compute allocation: if RL mostly sharpens, we should invest more compute in pretraining; if RL can genuinely discover, we should invest more in RL.

Studying these questions directly in reasoning LLMs trained on natural language is challenging. 
Systematic sweeps over pretraining and RL compute for language models are prohibitively expensive at frontier scale, while massive and heterogeneous pretraining corpora make it hard to attribute behaviors to pretraining versus RL.
Moreover, evaluation typically provides only final-answer correctness, leaving the policy’s behavior at individual reasoning steps largely opaque. 
These obstacles make it difficult to isolate how pretraining and RL interact.

Therefore, we use chess as a controlled testbed for studying these questions, with a training pipeline designed to mirror standard LLM training: pretraining on tokenized sequences of human game moves, supervised fine-tuning on synthetic reasoning traces, and RL with verifiable rewards. Chess has a compact, explicit action space, and move quality can be verified exactly via game outcomes or strong engines, providing ground truth at each step of a reasoning trajectory.
Human play data is plentiful and controllable: we can vary the amount, quality, and composition of the pretraining corpus (e.g., by filtering on player Elo) without making complex data-mixture choices.
Task-specialized small models can already reach nontrivial chess performance~\citep{ruoss2024amortized, zhang2024human, silver2017mastering}, making compute sweeps across pretraining and RL both affordable and informative: the models are sensitive enough for changes in scale and RL compute to produce measurable differences in performance and policy behavior.
Our goal is not to build the strongest chess model, but to use chess as a tractable setup for isolating how pretraining scale and verifiable RL interact.

Using this framework (Fig.~\ref{fig:overview}), we pretrain models from 5M to 1B parameters and sweep 36 pretraining--RL combinations. This enables us to quantify how pretraining choices shape subsequent RL scaling, and to inspect how RL changes the inherited policy at the level of individual moves. We further test whether the same patterns transfer beyond chess using a 1B language model trained on math-domain text. Our contributions are summarized as follows:
\begin{itemize}
    \item \textbf{A controlled chess testbed for pretraining-to-post-training studies.}
    We instantiate the standard LLM training pipeline in chess, with pretraining on human games, SFT on reasoning traces, and RL with verifiable rewards. 
    This setting enables systematic sweeps over pretraining and RL compute, and detailed analysis of policy reasoning traces (Section~\ref{sec:framework}).

    \item \textbf{A joint pretraining--RL scaling law.}
    We find that pretraining loss predicts the post-RL downstream performance, measured by pass@1, while the local RL slope grows approximately linearly with log pretraining token count. Combining this law with a Chinchilla-style pretraining loss scaling law lets us score hypothetical recipes defined by number of model parameters $N$, pretraining tokens $T$, and the amount of compute $C_{\text{RL}}$ allocated to reinforcement learning.
    We trace a compute-optimal frontier, and find that the optimal allocation shifts toward a larger RL fraction as total compute grows (Section~\ref{sec:experiments}).

    \item \textbf{Mechanism of RL policy change.}
    We analyze policy evolution and reasoning dynamics. We find that on easy puzzles, RL primarily amplifies correct moves the SFT policy already preferred; on hard puzzles, it can surface correct moves that were nearly absent, but also reinforces incorrect moves. These heterogeneous effects connect to the observation that RL improves pass@1 without consistently improving pass@$k$ (Section~\ref{sec:mechanism}).

    \item \textbf{Evidence of transfer beyond chess.}
    Across checkpoints of a fixed 1B language model pretrained on 10B–200B tokens of math-domain text, longer pretraining is associated with higher performance at fixed RL compute and a steeper local RL scaling slope (Section~\ref{sec:transfer}).
\end{itemize}

In sum, we provide a quantitative study of the interface between pretraining and post-training, and a practical way to reason about compute allocation across the two stages.

\section{Framework: Chess as a Testbed for Reasoning}
\label{sec:framework}

Our testbed, illustrated in Fig.~\ref{fig:overview}, mirrors the standard language model training pipeline and consists of three components: pretraining on a large-scale corpus of human games (§\ref{sec:pretraining}), a synthetic reasoning-trace generator for supervised fine-tuning (§\ref{sec:sft}), and RL on a verifiable chess-puzzle environment (§\ref{sec:rl}).

\textbf{Chess Representation.} Following prior work~\citep{zhang2024human}, we represent each chess game as an alternating sequence of the two players' moves serialized into tokens.
Drawing on the SAN and UCI conventions, we encode each move with four tokens, $\langle \texttt{piece} \rangle ;
\langle \texttt{source} \rangle ;
\langle \texttt{destination} \rangle ;
\langle \texttt{flag} \rangle$,
where $\langle \texttt{flag} \rangle$ marks special cases such as promotion, castling, en passant, check, or checkmate.
Any valid prefix of the resulting sequence determines a unique board state, and the full vocabulary has size $|\mathcal{V}| = 81$.
Examples are provided in Fig.~\ref{fig:overview}.

\subsection{Pretraining on Human Game Trajectories}

\label{sec:pretraining}

In the pretraining stage, the model learns the distribution of plausible move sequences from large-scale human play.
We collect game trajectories from Lichess\footnote{\url{https://database.lichess.org/}}, yielding a corpus that spans a wide range of player strengths and game outcomes.
The corpus can be subsampled along axes such as player Elo and game length, giving fine-grained control over data composition.
On this corpus, we train an autoregressive policy over tokenized games using the standard next-token prediction objective.

\subsection{Supervised Fine-Tuning with Synthetic Reasoning Traces}
\label{sec:sft}

In post-training, we train the model on chess puzzles: the solver must select the unique best move at each step, and each move can be verified exactly against the ground-truth solution line.

\textbf{Verifiable chess puzzle environment.} As shown in Fig.~\ref{fig:overview}, each puzzle specifies an initial board state $s_0$ together with a ground-truth solution line, formulating puzzle solving as a multi-step interactive decision problem in which the model acts only as the solver, rather than generating moves for both sides.
At each step $t$, the model observes the current solver state $s_t$ and outputs a candidate move $a_t$.
The environment then checks $a_t$ against the ground-truth best move: a mismatch terminates the episode immediately, while a match either completes the puzzle or triggers the corresponding opponent response, producing the next solver state $s_{t+1}$.
The model thus plays one move at a time: on a correct move, the environment appends the opponent's reply $o_t$ to the context, and the model conditions on it when generating $a_{t+1}$.
Following~\citet{ruoss2024amortized}, a trajectory is considered successful only if \emph{the model selects the correct solver move at every step} and thereby completes the full solution line.
We present an example puzzle game in Fig.~\ref{fig:puzzle_game_example}.

During pretraining, the model only observes move sequences and does not see explicit reasoning traces.
Prior chess models have augmented these base policies with inference-time search algorithms such as MCTS and beam search to recover planning behavior at test time~\citep{silver2017mastering,zhang2024human}.
We instead pursue an approach that mirrors how language models reason: rather than attaching an external search procedure, we elicit reasoning in context, training the model to produce its own reasoning trace before committing to a move.
We emphasize that for our models, the reasoning happens in chess move tokens, without natural language.

\textbf{Synthetic reasoning trace construction.} Motivated by prior work on CoT generation~\citep{long2023large}, we synthesize the CoT as a serialization of possible game continuations (Fig.~\ref{fig:overview}).
Intuitively, the reasoning trace is a set of plausible game continuations sampled from the pretrained model, slightly reordered to follow a tree-traversal structure.
Given an input board $s_0$, we sample $K$ continuations $\tau_1,\dots,\tau_K$ from a proposal policy.
Because these continuations share opening moves, we merge them by common prefixes into a tree of positions rooted at $s_0$, storing each shared prefix once.
The tree has $m \le K$ leaves, one per distinct continuation, and we write $\tilde{\tau}_{1},\dots,\tilde{\tau}_{m}$ for the corresponding root-to-leaf paths.
We serialize the tree in depth-first order to form the reasoning trace:
$r=\texttt{<T>}\;\tilde{\tau}_{1}\;\texttt{<sep>}\;\tilde{\tau}_{2}\;\texttt{<sep>}\;\cdots\;\tilde{\tau}_{m}\;\texttt{</T>}$,
where \texttt{<sep>} separates consecutive paths, and $\texttt{<T>}\texttt{</T>}$ are the thinking tags delimiting the reasoning.
The proposal policy $p_{\theta_{\mathrm{prop}}}$ is itself a pretrained model from Section~\ref{sec:pretraining}, so the resulting traces remain close to the model's own distribution rather than coming from an external search procedure.
We provide an example trace in Table~\ref{tab:chess_reasoning_example} and full construction details in Appendix~\ref{app:synthetic_trace}.

After producing the synthetic reasoning trace $r$, we train the model to commit to the best solution continuation $\tau^\star$ from $\{\tilde{\tau}_{1},\tilde{\tau}_{2},\dots,\tilde{\tau}_{m}\}$.
We train on the concatenated sequence $w=(r,\tau^\star)$.
Here, $\tau^\star=(a_1,o_1,a_2,o_2,\dots,a_H)$ consists of the player's moves and the environment moves.
Since opponent moves are provided by the environment at inference time, we mask opponent-move tokens from the loss and train only on reasoning $r$ and the model moves.

\subsection{Reinforcement Learning with Verifiable Rewards}
\label{sec:rl}

Starting from the SFT policy, we optimize the model on the puzzle environment with a binary outcome reward
$R(\zeta, s_0) = \mathbf{1}[a_1 = a_1^\star, \dots, a_H = a_H^\star]$,
where $\zeta$ is the full trajectory (reasoning trace followed by the executed move sequence) and $(a_1^\star, \dots, a_H^\star)$ is the ground-truth solution line.
In words, the model receives reward $1$ only if every executed move matches the corresponding ground-truth move, and $0$ otherwise, so a single mistake anywhere in the line yields no reward.
We optimize the policy with Group Relative Policy Optimization (GRPO)~\citep{shao2024deepseekmath}; algorithm details are in Appendix~\ref{app:algorithm}.

\subsection{Experimental Setup}

\label{sec:datasets_and_models}

We collect a 54B-token pretraining corpus of Blitz and Rapid human games played on Lichess in 2022, from which our scaling sweeps draw varying token budgets.
For post-training, we use 156K quality-filtered Lichess puzzles, spanning five difficulty bins (B1--B5, from easiest to hardest).
For evaluation, we curate a benchmark of 1,480 tactical puzzles spanning the same difficulty bins, balanced for theme diversity and solution length.
Since current models rarely solve B5 puzzles, all aggregate pass@$k$ results in the paper are reported over B1--B4, and we retain B5 for the difficulty-stratified mechanism analysis in Section~\ref{sec:policy_change}.
The three datasets are mutually disjoint at the board-position level to prevent contamination.
All models use the dense Qwen3~\citep{yang2025qwen3} base architecture, trained at 10 scales: $\{5\mathrm{M},\allowbreak 10\mathrm{M},\allowbreak 20\mathrm{M},\allowbreak 32\mathrm{M},\allowbreak 50\mathrm{M},\allowbreak 100\mathrm{M},\allowbreak 200\mathrm{M},\allowbreak 410\mathrm{M},\allowbreak 680\mathrm{M},\allowbreak 1\mathrm{B}\}$.
Full details on datasets, reasoning trace construction, model architectures, and training configurations are provided in Appendix~\ref{appendix:implementation-details}.

\section{Scaling Analysis: From Pretraining to Post-Training}
\label{sec:experiments}

We begin by analyzing pre-RL scaling behavior in our chess setup, then study how pretraining choices interact with RL scaling through two questions:
\begin{itemize}
    \item \textbf{RQ1: Compute allocation.} At different total compute levels, what final performance frontier is induced by different allocations between pretraining and RL?
    \item \textbf{RQ2: Predicting RL scaling.} Can pretraining properties, such as model size, number of pretraining tokens, and pretraining loss, predict RL scaling behavior in a given compute regime?
\end{itemize}

\subsection{Pre-RL Analysis: Scaling Behavior Before RL}
\label{sec:exp_pretraining_scaling}

We sweep 11 pretraining compute budgets from $6.5 \times 10^{16}$ to $6.5 \times 10^{19}$ FLOPs across 10 model sizes, corresponding to training runs of approximately $200$M to $52$B tokens.
Following the methodology in prior scaling-law studies~\citep{hoffmann2022training,roberts2026test}, we report IsoFLOP curves for validation loss on held-out human games, along with pass@$1$ and pass@$16$ on the downstream puzzle benchmark in Fig.~\ref{fig:pretrain_scaling}.
The results show that, within each model size, downstream benchmark performance continues to improve with additional pretraining over the compute range we study.
However, under fixed FLOPs, an optimal parameter-token allocation exists, and this optimum closely tracks validation loss on human games, pass@$1$, and pass@$16$. A functional-form fit of the Chinchilla law is provided in Appendix~\ref{sec:appendix_chichilla_fit}. 

For SFT, we compare training on the target move sequence alone against training on synthetic reasoning traces followed by the target answer, using the same number of puzzle samples across models.
Fig.~\ref{fig:comparison_pre_sft_rl} compare the two settings across model sizes and pretraining FLOPs.
SFT without reasoning traces improves pass@1 at every scale.
At larger scales, however, its pass@$16$ gain over the pretrained baseline shrinks, while SFT with reasoning traces gives a higher pass@16, especially on harder puzzles. 
Because SFT with reasoning traces lets larger models sample a broader set of correct solutions, we adopt it for all subsequent RL experiments, using it at every model size for consistency.
Moreover, for a fixed model size, stronger pretrained checkpoints consistently achieve higher post-SFT performance, and this ordering is preserved across the compute range we study.
This suggests that additional pretraining provides stronger pre-RL initializations. 
We include more discussions in Appendix~\ref{app:sft_appendix}.

\subsection{RQ1: What Is the Pretraining-RL Compute Tradeoff?}
\label{sec:rl_results}

\begin{figure}[t]
    \centering
    \includegraphics[width=\linewidth]{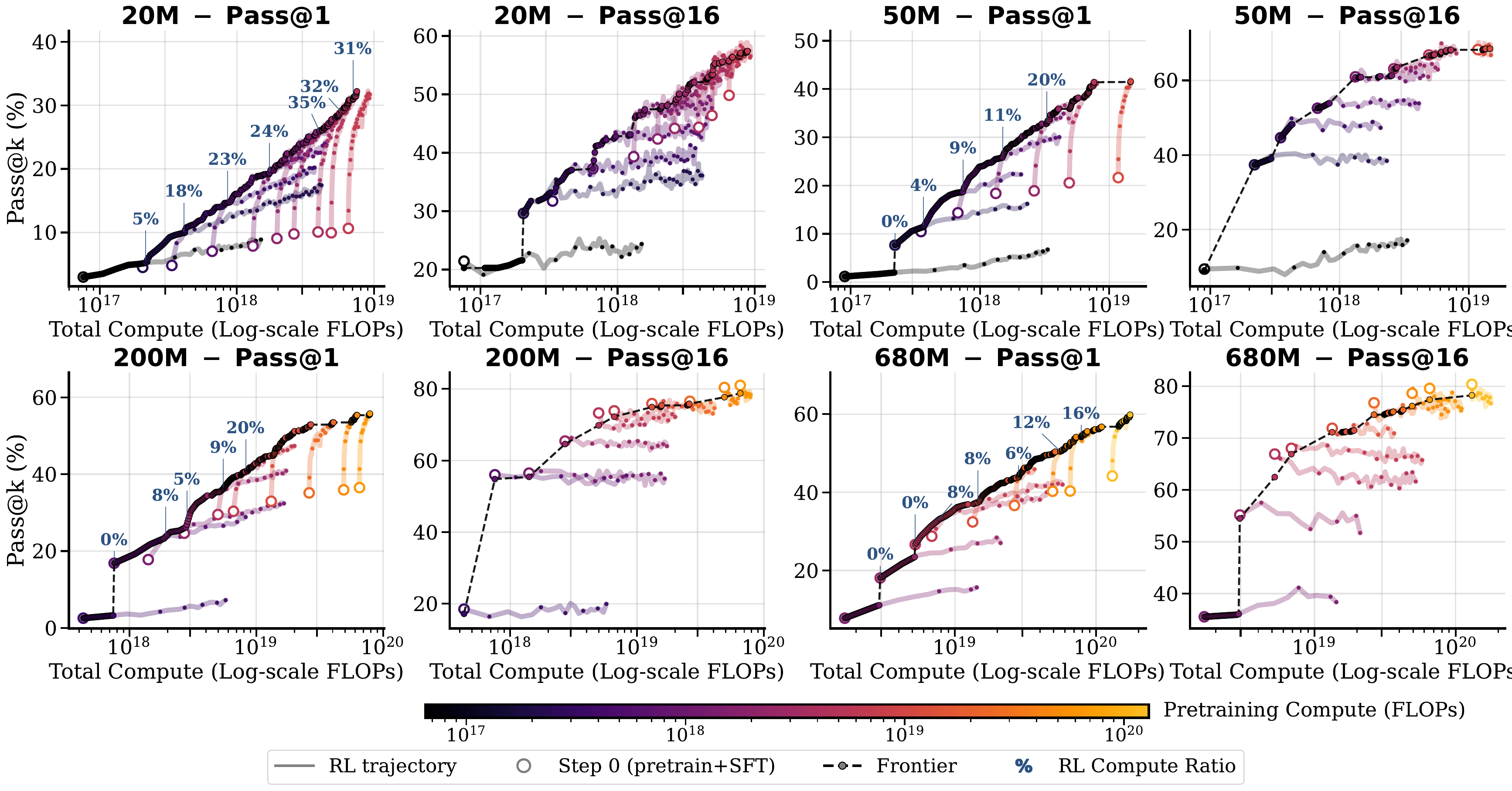}
    \caption{\textbf{Empirical frontier of puzzle benchmark performance 
across pretraining-RL sweeps for four model sizes (20M, 50M, 
200M, 680M).} Each point is a checkpoint evaluated on the puzzle benchmark. Color 
indicates pretraining compute; open rings mark the pre-RL baseline. Black curves mark the Pareto frontier. For the pass@$1$ curves, we label the \textcolor{rlratio}{fraction of RL compute} at which each run first reaches the frontier (excluding the smallest and largest compute ranges). For pass@1, RL consistently improves performance and the labeled RL ratio shows an increasing trend as total compute increases across models. For pass@16, gains from RL are smaller and additional pretraining is often more effective.}
    \label{fig:per_model_rl_frontier_flops}
\end{figure}

Under a fixed total compute budget and model size $N$, pretraining on more tokens (larger $T$) improves the initialization but leaves less compute for RL.
For each model size and total budget, we evaluate the final performance obtained from each available checkpoint after spending the remaining compute on RL, and define the fixed-budget frontier as the best result over these choices.
A frontier point selected at an earlier checkpoint (smaller $T$) indicates that additional RL compute is more valuable than further pretraining, whereas a point selected at larger $T$ indicates that the initialization with longer pretraining is worth the reduction in RL compute.
We evaluate the frontier for 4 model sizes using pretrained checkpoints from the IsoFLOP sweeps in Section~\ref{sec:exp_pretraining_scaling}.
For each model size in $\{20\mathrm{M}, 50\mathrm{M}, 200\mathrm{M}, 680\mathrm{M}\}$, we select checkpoints spanning 8-11 pretraining compute levels, apply the fixed SFT recipe, and perform RL from each resulting SFT policy for $1000$ to $5000$ steps\footnote{For reference, 2000 RL steps take approximately 160 H200 GPU-hours for a 50M model.}.
Our primary analysis measures all stages' compute in FLOPs\footnote{We note that equal FLOPs do not necessarily translate to equal wall-clock time across training stages. While prior work often measures RL compute using wall time or training steps~\citep{khatri2025art}, we use FLOPs as the primary allocation unit to isolate algorithmic compute. Wall time additionally reflects stage-specific systems factors that can differ substantially between pretraining and RL. Thus, wall time is important for measuring realized cost, but is a noisier unit for studying the algorithmic compute-allocation tradeoff between pretraining and RL.
}, using the estimation derived in Appendix~\ref{sec:appendix_flops_accounting}.

The results are shown in Fig.~\ref{fig:per_model_rl_frontier_flops}.
Across all training runs, RL substantially improves pass@$1$ performance, and pass@$1$ continues to increase with additional RL training.
We also mark the Pareto frontier across all model sizes and highlight the RL compute fraction at each frontier point.
For a fixed model, the frontier initially selects a high pretraining fraction (i.e., low RL fraction), suggesting that RL is strongly initialization-limited: under the same total budget, the extra RL compute from starting earlier does not compensate for the weaker pretrained policy.
However, the pretraining fraction decreases as the budget grows, indicating diminishing marginal downstream returns from additional pretraining.
Once the initialization is sufficiently strong, a larger fraction of compute is better spent on RL.
For instance, for the 20M model, the RL compute ratio increases from $5\%$ to $32\%$ along the frontier, indicating that additional RL becomes more useful in the higher-compute regime.

In contrast, pass@$16$ exhibits more mixed behavior.
For the smallest 20M model, pass@$16$ improves sharply at the beginning of RL training and then increases more gradually.
For larger models, however, the pass@$16$ curves remain nearly flat, and in some cases even degrade slightly with additional RL training.
In this regime, additional pretraining can lead to higher pass@$16$ performance than allocating the same compute to more RL.
This limited pass@$16$ improvement from RL is consistent with prior findings~\citep{yue2025does}.
We further analyze policy change through fine-grained categorization in Section~\ref{sec:policy_change}, providing additional insight into the mixed effects of RL.

\subsection{RQ2: Can Pretraining Properties Predict RL Scaling Behavior?}
\label{sec:exp_pre_to_rl_scaling}

Motivated by the frontier behavior above, we further study whether the observed learning trends can be parameterized by a functional form.
Prior work~\citep{khatri2025art} proposes an RL scaling law that models the performance of a fixed model as a sigmoid function of RL compute (see Eq.~\ref{eq:prior_sigmoid} in Appendix~\ref{app:local_rl_scaling_fit}), with the upper asymptote representing the RL ceiling: the performance the model would reach with unlimited RL compute.
However, the sigmoid only plateaus after very long RL training, so its ceiling cannot be identified reliably without runs long enough to reach that saturation.
Under our fixed total compute budgets, most RL runs instead cover only the early, non-saturated part of the curve.
We therefore take a first-order Taylor expansion of the sigmoid law in the non-saturated RL regime, writing the local RL scaling fit as
\begin{align}
    R_{N,T}(C)
    =
    R^{\mathrm{ref}}_{N,T}
    +
    B_{N,T}(\log_{10} C-\log_{10} C_{\mathrm{ref}}),
\label{eq:log-linear}
\end{align}
with derivations in Appendix~\ref{app:local_rl_scaling_fit}.
Eq.~\eqref{eq:log-linear} is log-linear in RL compute: anywhere along the fitted line, each $10\times$ increase in RL compute adds the same fixed increment $B_{N,T}$ to the reward.

\begin{figure}
    \centering
    \includegraphics[width=\linewidth]{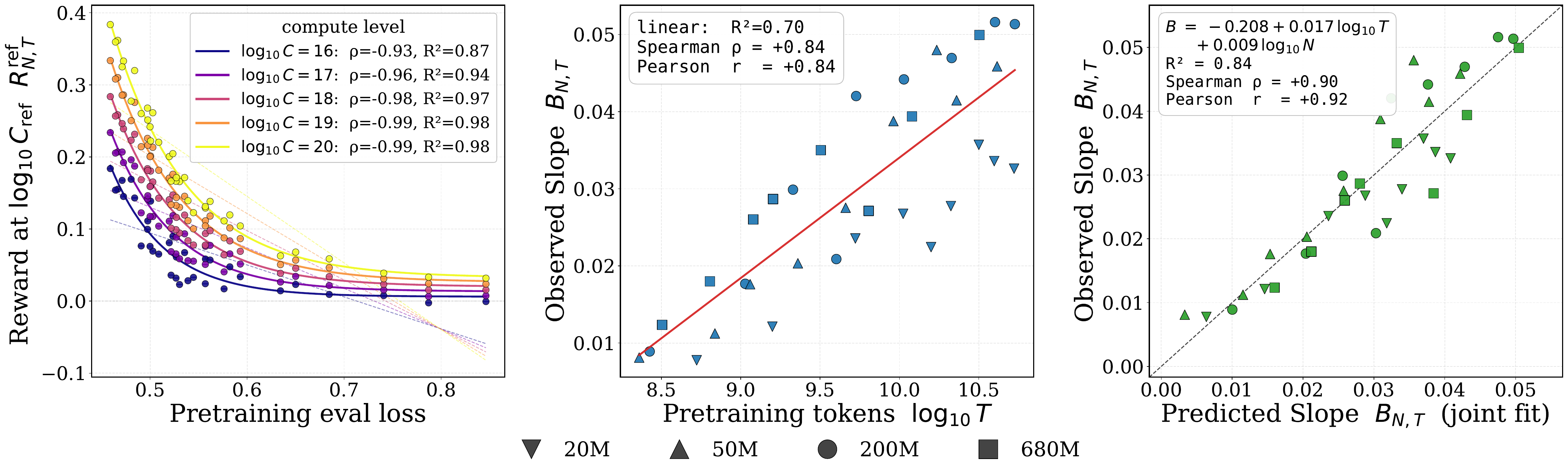}
    \caption{\textbf{Pretraining properties predict local RL 
scaling behavior.} (a) $R^{\mathrm{ref}}_{N,T}$, the fitted 
post-RL performance (pass@1 metric) at a reference RL compute level, versus pretraining validation loss. Each curve corresponds to a different reference compute level; the fit tightens as RL compute increases ($\rho{=}{-}0.93$ to ${-}0.99$). (b) $B_{N,T}$, the local slope measuring performance gain per decade of RL compute, versus pretraining tokens. More tokens predict faster RL improvement. (c) Joint fit of the slope using both \(\log T\) and $\log N$, achieving higher \(R^2\).}
    \label{fig:log_linear_fitting}
\end{figure}

\textbf{Fitting procedure.} 
For each RL run obtained in Section~\ref{sec:rl_results}, we collect the pass@\(1\) metric on the downstream benchmark after \(C\) FLOPs of RL compute as samples of \(R_{N,T}(C)\) for different values of $C$.
We fit per-run coefficients $(R^{\mathrm{ref}}_{N,T}, B_{N,T})$ using ordinary least squares to the observations $(x, y) = (\log_{10} C - \log_{10}C_{\mathrm{ref}}, R_{N,T}(C))$:
\begin{align*}
    (R^{\mathrm{ref}}_{N,T}, B_{N,T}) = 
    \arg \min_{a, b} \frac 1 {N_{C}} \sum_{i=1}^{N_C}
    \big(R_{N,T}(C_i) - a - b
    (\log_{10} C_i-\log_{10} C_{\mathrm{ref}})\big)^2,
\label{eq:log-linear}
\end{align*}
where $N_{C}$ is the number of observations.
The reference compute level $C_{\mathrm{ref}}$ defines a linear shift in the features of our linear fit, and does not affect the quality of the fit itself or the slope parameter $B_{N, T}$; it does affect the reference reward parameter $R^{\mathrm{ref}}_{N,T}$, which should be interpreted as the fitted reward at a reference compute level $C_{\mathrm{ref}}$.
We then ask whether the parameters $\phi_{N,T}=\{R^{\mathrm{ref}}_{N,T},B_{N,T}\}$ can be predicted from properties of the pretrained checkpoint, namely the model size $N$, the number of pretraining tokens $T$, and the pretraining validation loss $L_{\mathrm{pt}}(N,T)$.
In this analysis, we evaluate only on the B3--B4 benchmark with intermediate difficulty, where the models we consider do not saturate the evaluation.


\textbf{Connecting pretraining to the fitted log-linear law.}
Fig.~\ref{fig:log_linear_fitting} (left) shows representative fits relating the post-RL performance to pretraining loss. Since our RL compute spans $10^{16}$ to $10^{20}$, we evaluate different choices of $C_{\mathrm{ref}}$ within this range. 
The pass@1 performance at a given RL compute level, \(R^{\mathrm{ref}}_{N,T}\), is strongly predicted by pretraining validation loss: across choices of \(C_{\mathrm{ref}}\), lower pretraining loss consistently corresponds to higher post-RL performance, and the relationship becomes increasingly monotone at larger reference compute \(C_{\mathrm{ref}}\). 
For example, the Spearman correlation increases from $|\rho|=0.93$ to $|\rho|=0.99$ as \(\log_{10} C_{\mathrm{ref}}\) increases from \(16\) to \(20\). 
We thus fix $C_{\mathrm{ref}} = 10^{20}$ FLOPs as a constant across all runs (Appendix~\ref{app:cref-sensitivity}).
This dependence \(f(L_{\mathrm{pt}}(N,T)) \approx R^{\mathrm{ref}}_{N,T}\) between the pretraining loss and the RL reward is better captured by a nonlinear fit than by a linear one as we show in Fig.~\ref{fig:a_fitting}. 
By comparison, as shown in Fig~\ref{fig:R0_vs_L}, \(R_0\), the post-SFT performance before RL, is less tightly predicted by pretraining loss than post-RL performance at high RL compute. 
We observe a similar but weaker trend for post-SFT pass@$k (k>1)$ metrics, shown in Fig.~\ref{fig:sft_passk_vs_loss}. 
In Appendix~\ref{app:20m_ceiling}, we additionally analyze the sigmoid law fit on the 20M model runs (the only model size where the fit is sufficiently determined by data), where it shows that the asymptotic performance ceiling is similarly predictable from pretraining loss (see Fig~\ref{fig:20m_ceiling}).

As presented in Fig.~\ref{fig:log_linear_fitting} (middle), the per-run slope parameter $B_{N,T}$ shows a positive linear correlation with pretraining tokens $\log_{10}T$ (Pearson $r{=}{+}0.84$). 
We therefore fit $B_{N,T}$ against $\log_{10}T$ alone, tokens per parameter $\log_{10}(T/N)$, and a joint linear model using both $\log_{10}T$ and $\log_{10}N$. 
Among these, the joint model attains the lowest RMSE and highest $R^2$ in predicting the observed slope, so we adopt it as our parameterization of $g(N,T) \approx B_{N,T}$. 
As shown in Fig.~\ref{fig:log_linear_fitting} (right), the joint fit assigns a larger coefficient to $\log_{10}T$ than to $\log_{10}N$. 
Together, these observations indicate that the RL improvement rate is largely shaped by the amount of pretraining data exposure, with model size providing a weaker positive correction: at a fixed token budget $T$, larger models improve RL performance slightly faster with RL compute. 
We compare slope parameterizations in Fig.~\ref{fig:b_fitting} and report fits on other benchmark subsets in Figs.~\ref{fig:b_fitting_b1} and~\ref{fig:b_fitting_b2}. Across these subsets, the estimated slope remains positively correlated with pretraining tokens overall. However, we also observe failure cases when stronger models approach saturation on the easy benchmarks (i.e. the sigmoid reward curves approach a plateau), which systematically compresses their local slope estimates. We therefore interpret the observed association as a local empirical trend over the compute range studied where the benchmarks are not saturated, rather than as a global relationship that should hold across all training regimes.

Jointly, these observations motivate the following scaling law. We report the full statistical analysis, including parameterization of $f$ and $g$, in Appendix~\ref{app:joint-rl-scaling-law}. 
Leave-one-out validation on our existing runs shows that the fitted parameterization predicts held-out RL trajectories well when the observed pretraining loss is provided (Fig.~\ref{fig:loo_observed}). 

\begin{scalinglaw}
\vspace{-1mm}
\[
\small
R(C_{\mathrm{RL}},N,T)
=
f\!\left(L_{\mathrm{pt}}(N,T)\right)
+
g(N,T)
\left(
\log_{10} C_{\mathrm{RL}}
-
\log_{10} C_{\mathrm{ref}}
\right)
\]
\vspace{-4mm}
\end{scalinglaw}

\textbf{Extrapolating the compute-optimal frontier.} 
Combining our joint pretraining-RL law with the Chinchilla loss prediction $L(N,T)$ allows us to evaluate hypothetical training recipes $(N,T,C_{\mathrm{RL}})$ without actually training them.
We consider a ladder of 13 model sizes from 20M to 2B parameters. 
For each size, we evaluate 260 total-compute budgets between $10^{17}$ and $10^{21}$ FLOPs. 
For each model size $N$ and total-compute budget $C$, we search over 400 candidate splits of the budget between pretraining and RL, each a choice of $(T, C_{\mathrm{RL}})$.
We then predict the pretraining loss \(L_{\mathrm{pt}}(N,T)\) using the Chinchilla law and estimate the post-RL rewards \(R(C_{\mathrm{RL}},N,T)\) using our joint law.
Based on the grid search, we identify the best predicted reward $R_N^\star(C)$ achievable by model size $N$ at budget $C$.
The global frontier is then obtained by maximizing $R_N^\star(C)$ over model sizes at each budget $C$. See Appendix~\ref{sec:sim-frontier} for more details.

\begin{figure}[tbp]
    \centering
    \includegraphics[width=\linewidth]{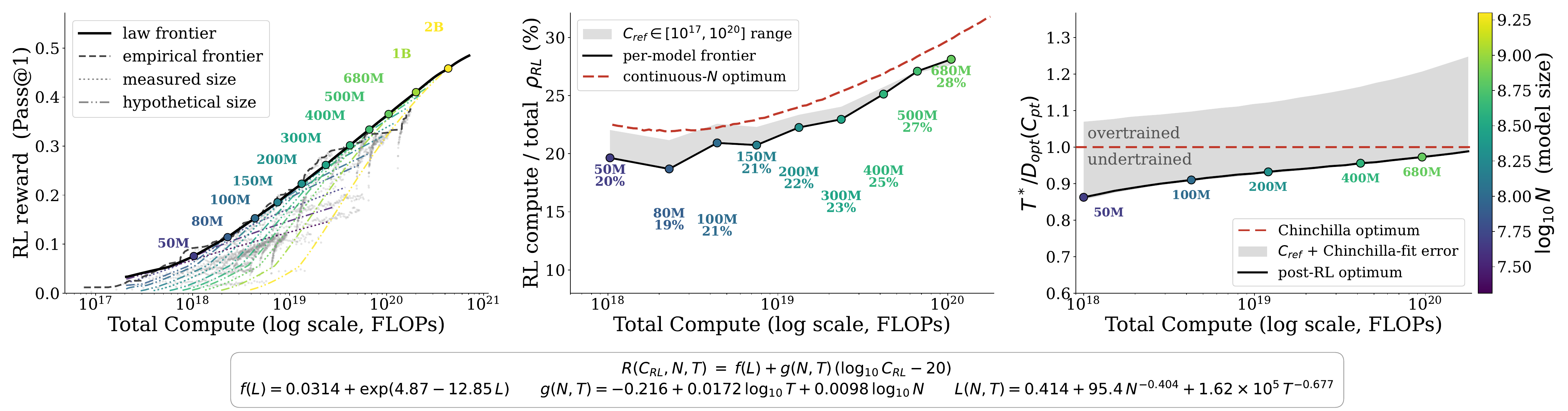}
    \caption{\textbf{Extrapolated compute-optimal frontier across model sizes using the fitted law.} \textbf{(a)} Simulated compute-optimal reward ($R_N^\star(C)$) for each model size. The black curve is the global simulated frontier, dots mark model-size transitions, and the dashed gray curve is the empirical frontier. The two frontiers closely agree. \textbf{(b)} Compute-optimal RL share ($\rho_{\mathrm{RL}}=C_{\mathrm{RL}}^\star/C_{\mathrm{total}}$). The optimal RL fraction increases with total compute, from approximately 20\% at 50M to 28\% at 680M; the red dashed curve shows the continuous-(N) optimum. \textbf{(c)} Optimal pretraining tokens relative to the Chinchilla allocation at the same pretraining compute. The results do not show systematic deviation from Chinchilla scaling.
}
    \label{fig:law_runs_frontier}
\end{figure}

The resulting frontier is shown in Fig.~\ref{fig:law_runs_frontier}. For each model size, we locate the budget at which it first attains the global fitted frontier and report the corresponding RL-compute share. 
We also compute the pretraining token count $T^\star$ selected at that frontier point and compare it to the Chinchilla-optimal token count under the same pretraining compute budget, $D_{\mathrm{opt}}(C_{\mathrm{pt}})$ tokens. The ratio $T^\star/D_{\mathrm{opt}}(C_{\mathrm{pt}})$ measures the deviation from Chinchilla allocation: values above one indicate more pretraining tokens than the Chinchilla optimum, while values below one indicate fewer. The results indicate that the compute-optimal RL share increases with total compute, while the corresponding pretraining token allocation does not significantly deviate from Chinchilla scaling.

\begin{takeaway}
\textcolor{takeawaycolor}{\textbf{Takeaway:}} 
(1) Pretraining affects RL in two ways: lower pretraining loss predicts higher downstream reward (pass@1) at a fixed RL compute level, and larger pretraining data scale improves the rate at which reward grows with RL compute. 
(2) RL contributes more directly to pass@1 gains, whereas pass@$k$ remains more sensitive to pretraining scale.
(3) For pass@1 at fixed model size, in the low total-compute regime, pretraining is the dominant contributor to the final performance. 
As the budget increases, RL should scale proportionally with pretraining.
\end{takeaway}

\section{Mechanism Analysis: Policy Evolution During RL Post-training}
\label{sec:mechanism}

In this section, we study how RL training reshapes the policy beyond its pre-RL prior. Prior work disagrees on whether post-training primarily elicits capabilities already present in the base model~\citep{yue2025does} or induces qualitatively new behaviors~\citep{sun2025rl,yuan2025f}. Our framework provides direct access to both the policy distribution over moves at each state and the structured reasoning traces, enabling us to study policy evolution along two dimensions:

\begin{itemize}
    \item \textbf{RQ1: How does RL reshape the move distribution at each state?}
    We analyze how probability mass is redistributed across candidate moves and characterize whether RL amplifies already-preferred moves, surfaces low-probability correct moves, or reinforces incorrect modes of the predictive distribution.

    \item \textbf{RQ2: How does RL reshape the chain-of-thought reasoning that produces those moves?}
    We examine how the structure and content of the reasoning traces change after RL, and whether these changes correspond to improved decision-making.
\end{itemize}

\subsection{RQ1: How Does RL Change the Move Policy?}
\label{sec:policy_change}

For any puzzle board state $s$, we define the induced move policy $\pi_{\theta}$ as a probability distribution over the legal move set $\mathcal A(s)$. For the pretraining model, the unnormalized score for move $a$ is
$\widetilde{\pi}_{\theta_{\mathrm{pre}}}(a\mid s)=\prod_{j=1}^{|\tau(a)|} p_{\theta_{\mathrm{pre}}}(x^{(a)}_j\mid s,x^{(a)}_{<j})$,
where $\tau(a)$ is the token serialization of the move. For SFT and RL models, the move score should be marginalized across reasoning traces:
$\widetilde{\pi}_{\theta}(a\mid s)=\sum_r \pi_{\theta}(r)\,\widetilde{\pi}_{\theta}(a\mid s,r)$,
where $\widetilde{\pi}_{\theta}(a\mid s,r)$ is obtained by scoring the token serialization of move $a$ conditioned on trace $r$. The induced move policy $\pi_{\theta}(a\mid s)$ is then obtained by normalizing $\widetilde{\pi}_{\theta}(a\mid s)$ over the legal move set $\mathcal A(s)$. We sample 128 reasoning traces per puzzle to evaluate the marginal distribution and provide more details in the Appendix~\ref{sec:appendix_policy_evolution}.

\textbf{RL is not well explained by a uniform temperature scaling of SFT.} Inspired by prior work~\citep{karan2025reasoning}, we test whether the RL policy is a power transformation of the SFT policy, $\pi_{\mathrm{RL}}(a\mid s)\propto \pi_{\mathrm{SFT}}(a\mid s)^\alpha$, where $\alpha>1$ corresponds to sharpening. Exact power scaling implies that centered RL log-probabilities are linear in centered SFT log-probabilities with slope $\alpha$, so we fit this slope by zero-intercept regression, both globally and per state (Appendix~\ref{app:alpha_power_fitting}). Table~\ref{tab:beta_fitting_statistics} shows the fitted global slope increases during RL, indicating that RL sharpens the SFT policy on average. However, the fit achieves only moderate $R^2$ and the per-state slopes vary substantially, indicating that RL reshapes the policy in state-dependent ways. 

\begin{figure}[tbp]
    \centering
    \includegraphics[width=1.0\linewidth]{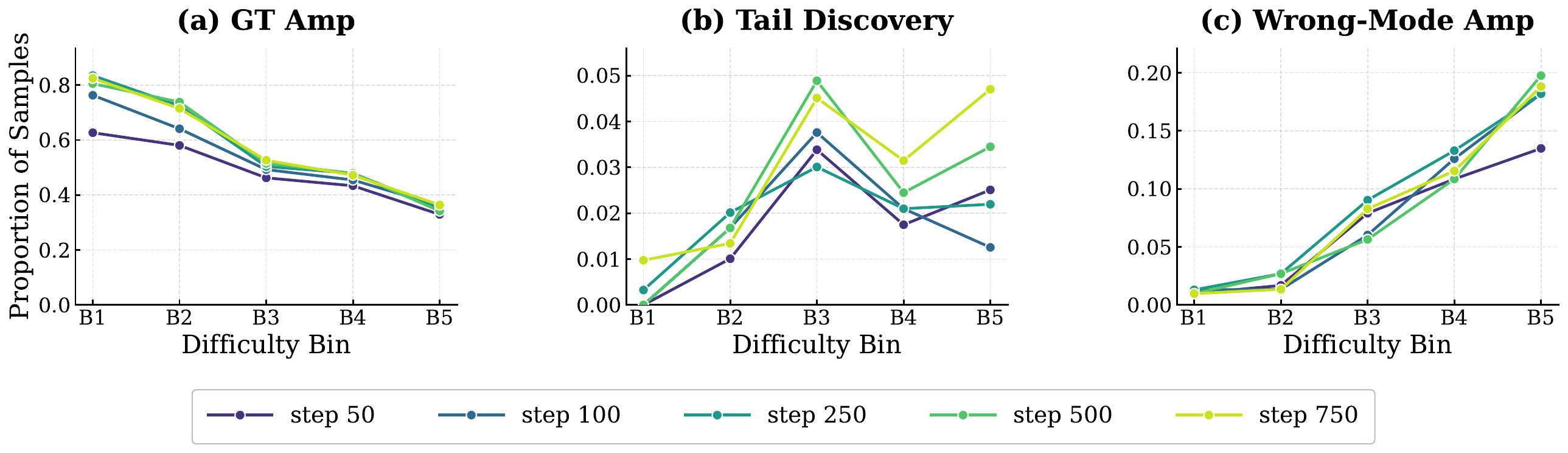}
    \caption{\textbf{RL reshapes the move policy in qualitatively 
different ways across puzzle difficulty.} Each panel shows the 
proportion of puzzle states assigned to one policy-update 
category (Table~\ref{tab:policy_update_taxonomy}) across 
difficulty bins B1-B5, at RL training steps until 750. On easy 
puzzles, ground-truth amplification (a) dominates. On harder 
puzzles, tail discovery (b) and wrong-mode amplification (c) 
both increase, showing that RL simultaneously surfaces 
previously absent correct moves and reinforces incorrect ones.}
    \label{fig:policy_categorization}
\end{figure}

We therefore investigate \emph{how probability mass is redistributed} at individual states. We categorize local policy changes by how the ground-truth move transitions between the top-$k$ sets of the initial policy $\pi_{\theta_0}$ and the updated policy $\pi_{\theta_1}$. Table~\ref{tab:policy_update_taxonomy} gives the formal definitions for the full categorization. In Fig.~\ref{fig:policy_categorization}, we highlight three major categories of policy changes:

\begin{itemize}
\item \textit{Ground-truth amplification}: the correct move is already in the top-$k$ set and is further reinforced by training.
\item \textit{Tail discovery}: training promotes a correct move from the low-probability tail, defined as probability below $\epsilon_{\mathrm{tail}} = 0.05$, into the top-$k$ set.
\item \textit{Wrong-mode amplification}: the correct move remains outside the top-$k$ set, while the initially preferred wrong move is further reinforced.
\end{itemize}

We set $k=3$ in practice and present additional results in Appendix~\ref{app:policy_evolution}.

\textbf{On easier puzzles RL training mostly amplifies correct moves the SFT policy already preferred, while on harder puzzles it both surfaces moves that were nearly absent under SFT and reinforces incorrect ones.}  In our experiments, we instantiate $\theta_0$ as the SFT policy and $\theta_1=\theta_t$ as the RL policy at training step $t$. Fig.~\ref{fig:policy_categorization} shows the resulting category proportions across B1-B5 puzzle test sets. Fig.~\ref{fig:puzzle_example} presents a qualitative example of how RL recovers the ground-truth move from the tail distribution for a hard puzzle on the B5 test set.

\begin{takeaway}
\textcolor{takeawaycolor}{\textbf{Takeaway:}} 
RL produces limited pass@$k$ gains because probability mass is redistributed in several ways: RL strengthens correct modes and discovers some correct tail moves, but also amplifies wrong modes on harder tasks. Mitigating wrong-mode amplification is therefore important for improving RL beyond pass@$1$.
\end{takeaway}

\subsection{RQ2: How Does RL Change the Dynamics of Chain-of-Thought Reasoning?}
\label{sec:cot}

As a complementary analysis, we examine how the structure of reasoning traces evolves over the course of RL training. Because our CoT format comprises explicit move sequences, each rollout can be reconstructed as a prefix tree rooted at the puzzle state (Section~\ref{sec:sft}), with each node corresponding to a move, enabling us to probe both the structure and quality of the model's reasoning.  In Fig.~\ref{fig:cot_evolution}, Fig.~\ref{fig:cot_length_app} and Fig.~\ref{fig:cot_target_continuation_app}, we compare two representative RL runs for the 20M and 50M models pretrained under matched compute, in terms of reasoning tree structure, move quality and search behavior.

\textbf{Reasoning trace quality improves during RL, but deeper search remains challenging.}
The parsed traces show that models primarily expand search breadth rather than depth: the width-to-depth ratio and branching factor increase while maximum search depth stays roughly flat. Meanwhile, the quality of proposed moves improves for both the model's own moves and its predicted opponent responses, and the model becomes more likely to surface the ground-truth move in its CoT and commit to the best candidate it has considered. However, Fig.~\ref{fig:cot_target_continuation_app} shows that the model still struggles to recover continuations requiring more than 5 moves, suggesting that RL improves candidate generation and selection faster than long-horizon search. These structured search features may guide future SFT data construction toward examples that encourage deeper, more systematic search.

\section{Transfer to Text: A Qualitative Case Study in Math}
\label{sec:transfer}
To test whether the scaling law identified in our chess setting also transfers to natural language modeling, we pretrain a 1B-parameter OLMo-2~\citep{olmo20242} model on a 200B-token mixed pretraining corpus consisting of 70\% Nemotron-CC-Math-v1\footnote{\url{https://huggingface.co/datasets/nvidia/Nemotron-CC-Math-v1}}~\citep{mahabadi2025nemotron} and 30\% Dolma3, the mid-training corpus used for OLMo-3~\citep{olmo2025olmo}. 
We train a single main run with a linear learning rate schedule with constant warmup; we use 5B Dolma3 tokens for learning rate annealing at different points along the run, producing 14 checkpoints between 10B and 200B tokens. We then perform one epoch of supervised fine-tuning on NuminaMath-CoT\footnote{\url{https://huggisngface.co/datasets/AI-MO/NuminaMath-CoT}}~\citep{numina_math_datasets}, followed by RL on a 24.9K-problem mixed training corpus drawn from GSM8K~\citep{gsm8k}, MATH~\citep{math}, and DeepScaler~\citep{luo2025deepscaler}. We evaluate on a held-out validation set of 500 problems drawn from the training distribution, as well as on the GSM8K and MATH test sets. All results are reported using pass@1, estimated from 16 sampled completions per problem at temperature \(0.7\). Extra experiment details are in Appendix~\ref{app:olmo}.

\begin{figure}[t]
    \centering
    \includegraphics[width=1.0\linewidth]{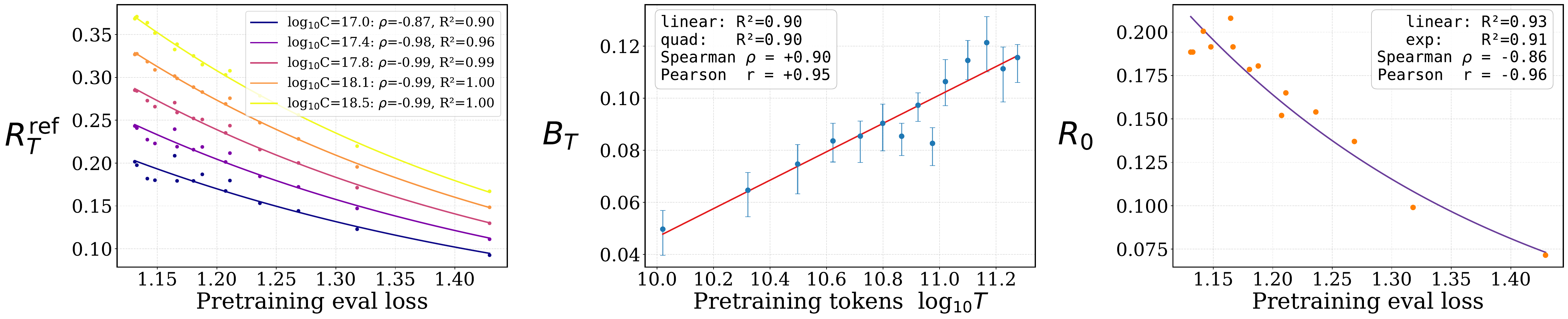}
    \caption{\textbf{The predictive pattern extends to the math domain.}
Across 1B OLMo-2 checkpoints from 10B to 200B pretraining tokens, lower pretraining loss consistently predicts higher post-RL performance $R_{T}^{\mathrm{ref}}$, with the fit tightening as RL compute increases. The slope $B_T$ increases nearly linearly with $\log_{10}T$. The $R_{T}^{\mathrm{ref}}$ relationship with pretraining loss becomes tighter as RL compute increases, mirroring the chess results (Fig.~\ref{fig:log_linear_fitting}).}
    \label{fig:skyeasy_law_fitting}
\end{figure}

We report the fitting on the held-out set in Fig.~\ref{fig:skyeasy_law_fitting}, and additional results in Fig.~\ref{fig:gsm_math_law_fitting}. Overall, the experiment provides early evidence that the pretraing-to-post-training scaling structure identified in chess extends to modern language model training. 
Despite differences in task format, data distribution, and training recipe, we find a similar pattern: the post-RL performance level at high RL compute is well-predicted from the pretraining loss, and the slope of the RL reward curves improves approximately linearly with the pretraining tokens. 


\section{Related Work}

\textbf{Reinforcement Learning for Reasoning.}
RL with verifiable rewards has become standard for improving reasoning in language models~\citep{guo2025deepseek, lambert2024tulu, yu2025dapo, zeng2025simplerl}, yet what RL actually does to the pretrained policy remains debated. Prior work has variously argued that RL primarily amplifies existing reasoning patterns~\citep{yue2025does}, improving pass@1 while sometimes reducing large-$k$ coverage; composes pretrained skills into new ones~\citep{yuan2025f}; or does both depending on the problem regime~\citep{sun2025rl,zhang2025interplay}. The difficulty of resolving this in natural language stems from the enormous action space, ambiguous token-level actions, and lack of step-level supervision. Our chess setting provides engine-based supervision at every board state and a clean definition of actions in token space, enabling us to decompose RL's effect by puzzle difficulty. We show that RL amplifies existing correct moves on easy puzzles, while surfacing buried moves and reinforcing incorrect ones on hard puzzles.

\textbf{Scaling Laws from Pretraining to Post-training.}
Scaling laws have long been used as quantitative tools to predict language-model performance from training compute. \citet{kaplan2020scaling} showed that language-model loss follows power laws in model size, data, and compute, and \citet{hoffmann2022training} refined the compute-optimal allocation between model size and training tokens. Subsequent work has extended scaling analysis beyond standard pretraining loss, including overtrained models and downstream top-1 error~\citep{gadre2025language}, test-time sampling and pass@$k$~\citep{roberts2026test}, inference-aware scaling~\citep{10.5555/3692070.3693840}, synthetic pretraining data~\citep{qin2025scaling}, and fixed-data SFT scaling~\citep{zhang2024scaling}. More recent work also characterizes scaling during RL post-training: \citet{khatri2025art} fit sigmoidal RL scaling curves, while \citet{cheng2026isocompute} study compute-optimal sampling within RL. 
However, these works either treat downstream use as given or treat the pretrained initialization as fixed. 
\citet{huang2025best} and \citet{chen2025coverage} show that coverage, the
probability mass a policy places on high-quality responses, characterizes
Best-of-$N$ and pass@$k$ performance, and the latter further prove that
next-token prediction implicitly optimizes coverage, thereby connecting
pretraining to post-training. 
We complement this
account empirically: we directly study the relationship between pretraining properties with RL scaling behavior in a controlled
domain, and find that pretraining loss is strongly predictive of the post-RL
pass@1 level.

\textbf{Allocating Compute between Pretraining and RL.}
An important question in multi-stage language-model training is how much pretraining is needed to make RL effective, and how limited compute should be allocated between pretraining and RL post-training. Empirically, \citet{qi2025evolm} study tradeoffs across pretraining, continued pretraining, SFT, and RL, showing that post-training gains can saturate as pretraining and RL compute increase. Recent work further studies early RL during pretraining~\citep{bansal2026rl}. However, these works mainly characterize stage-wise effects or training recipes, rather than directly modeling the pretraining-vs.-RL compute allocation problem. Our work directly varies compute across pretraining and RL, establishing a quantitative scaling law for how pretraining compute changes the subsequent RL learning curve.
\section{Conclusions}
\label{sec:conclusions}

We used chess as a testbed for studying how pretraining influences RL dynamics and how RL reshapes the inherited policy. We established a joint scaling law: pretraining loss predicts post-RL performance level, while pretraining data scale is closely associated with the slope of RL improvement. The resulting compute-allocation frontier suggests a tradeoff between pretraining and RL. As the total budget grows, the optimal pretraining fraction tends to decrease, indicating that RL should take an increasingly large share of compute. Meanwhile, starting RL too early from weakly pretrained checkpoints gives limited gains in our setting, suggesting that RL remains initialization-dependent and requires sufficient pretraining exposure before it becomes effective. The behavior also differs between pass@1 and pass@16. Mechanistically, RL is not uniform sharpening: its effect varies systematically with puzzle difficulty, amplifying correct moves already preferred by the SFT policy on easy puzzles, while surfacing buried moves but sometimes reinforcing incorrect ones on hard puzzles. This mixed redistribution explains why RL can improve pass@1 without consistently improving pass@16. Finally, we observe a similar qualitative pattern in the math domain on a 1B language model, suggesting our findings extend beyond chess.

There are several exciting directions for future work.
First, the scaling framework can be used to study when to switch from pretraining to RL, or more generally how to allocate compute across $(N, T, C_{\text{RL}})$. Second, improving RL beyond pass@1 likely requires methods that reduce wrong-mode amplification and expand the support of correct solutions, rather than only sharpening the current policy. Third, these results might motivate better ways to combine pretraining and RL. Since RL gives limited gains when started from weakly pretrained checkpoints but pure RL can also amplify wrong modes on harder states, a fixed two-stage recipe may be suboptimal. Future work could study interleaving strategies that decide when additional pretraining data is more valuable than additional RL updates. More broadly, our setting also provides a controlled testbed for studying synthetic data design, self-play, transcendence~\citep{zhang2024transcendence}, and weak-to-strong generalization~\citep{burns2023weak}.

\section*{Acknowledgement}

We thank Vatsal Baherwani, Sean McLeish, Vadim Bereznyuk, Timur Garipov and Yulin Chen for their insightful feedback on the draft. This work was also supported in part by NYU IT High Performance Computing resources, services, and staff expertise.

\bibliographystyle{plainnat}
\small\bibliography{ref}

\newpage
\appendix

\section{Discussions and Limitations}
\label{appendix:discussion-limitation}

Our testbed mirrors the standard LLM pipeline (pretraining on human data, SFT on reasoning traces, RL with verifiable rewards), and the phenomena we study (how pretraining properties shape RL scaling, how RL reshapes the inherited policy) are pipeline-level dynamics rather than chess-specific ones. 
That said, chess differs from natural language in ways that limit direct transfer: the vocabulary is small (81 tokens), verification is exact, and reasoning is not entangled with world knowledge or fluency. Our scaling exponents and category proportions should therefore be read as characterizing the structure of the pretraining-to-RL interface in a controlled setting, not as quantitative predictions for language models. Additionally, our RL environment uses puzzles with unique designated solutions and binary rewards, which is a restricted form of verification compared to the partial-credit or open-ended rewards common in language tasks. Furthermore, our models reach at most 1B parameters. The scaling trends we identify (e.g., the decreasing optimal pretraining fraction) may behave differently at larger scale, and verifying this is beyond our current compute budget. Besides, our structured CoT format uses a specific tree-based serialization and data-generation choices. Different reasoning trace formats could yield different CoT evolution dynamics under RL, which is worth exploring as a future direction. We also discuss the limitations for the fitted law in Section~\ref{app:limitations_law_fitting}.

\section{Extended Related Work}

\textbf{Scaling Laws for Language Models from Pretraining to Post-training.}
Scaling laws have long been used as quantitative tools to predict language-model performance from training compute. \citet{kaplan2020scaling} showed that language-model loss follows power laws in model size, data, and compute, and \citet{hoffmann2022training} refined the compute-optimal allocation, finding that model size and training tokens should scale roughly equally. Subsequent work has extended scaling analysis beyond standard pretraining loss: \citet{gadre2025language} show that overtrained language models scale reliably in both validation loss and average downstream top-1 error, while \citet{roberts2026test} incorporate test-time sampling through pass@$k$ and show that overtraining can become compute-optimal when inference compute is accounted for. Other work studies inference-aware scaling~\citep{10.5555/3692070.3693840} and synthetic pretraining data~\citep{qin2025scaling}. On the post-training side, \citet{zhang2024scaling} study scaling laws for fixed-data SFT, rather than on-policy RL, and find that fine-tuning performance depends more strongly on model scale than on pretraining-data scale. More recent work has also begun to characterize scaling behavior during RL post-training: \citet{khatri2025art} fit sigmoidal RL scaling curves and identify stable training recipes, while \citet{cheng2026isocompute} study compute-optimal allocation for sampling within RL. However, both treat the pretrained initialization as fixed. Across these lines of work, pretraining scaling laws treat downstream use as given, while post-training scaling laws treat the initialization as given. Neither asks how pretraining compute changes the shape of the RL scaling curve. Our work fills this gap: we show that pretraining loss predicts post-RL performance level and pretraining data scale predicts the rate of RL improvement in terms of average reward.

\textbf{Allocating Compute between Pretraining and RL.}
An important question in multi-stage language-model training is how much pretraining is needed to make RL effective, and how limited compute should be allocated between pretraining and RL post-training. \citet{chen2025coverage} theoretically connect pretraining and post-training through coverage, arguing that the probability mass assigned to high-quality responses determines whether post-training and test-time scaling can succeed. Empirically, \citet{qi2025evolm} study tradeoffs across pretraining, continued pretraining, SFT, and RL, showing that post-training gains can saturate as pretraining and RL compute increase. Recent work further studies early RL during pretraining~\citep{bansal2026rl}, front-loading reasoning data into pretraining~\citep{akter2025front}, and how RL gains depend on pretraining headroom and task difficulty~\citep{zhang2025interplay}. These works provide important evidence that pretraining and RL are coupled, but they mainly characterize stage-wise effects or training recipes rather than directly modeling the pretraining-vs.-RL compute allocation problem. Our results show that for large-$k$ pass@$k$, extended pretraining remains consistently beneficial. For pass@$1$, additional pretraining also improves performance, but its marginal benefit decreases as the total compute budget increases. Our findings also offer a complementary explanation for the saturation observed in prior multi-stage studies \citep{qi2025evolm}. Since post-RL performance is predictable from pretraining loss, diminishing returns in pretraining loss naturally translate into diminishing marginal gains after RL. Thus, saturation from increased pretraining need not imply a conflict between pretraining and RL. Rather, RL inherits part of its attainable performance level from the pretrained policy, while its learning rate is shaped by pretraining data scale. This perspective turns stage-wise saturation into a predictable consequence of the pretraining-to-RL scaling relationship.

\textbf{Controlled Testbeds for Studying Reasoning.} A line of work has studied controlled synthetic settings that isolate specific aspects of reasoning~\citep{ye2024physics, zhang2025interplay, yuan2025f, sun2025rl}. We offer a complementary testbed based on chess that more closely mirrors language-model training: models are first pretrained on human behavior data and then improved through RL with verifiable rewards. Chess requires multi-step planning, supports exact verification at every move, and yields nontrivial scaling regimes across both pretraining and RL, making it a useful environment for studying reasoning from pretraining to post-training. Prior chess language models study pretraining or amortized search without RL~\citep{feng2023chessgpt, ruoss2024amortized, zhang2024human}, RL from scratch without pretraining on human data~\citep{silver2017mastering}, or reasoning evolution from a fixed-size model~\citep{dionisopoulosreasoning}. None study how pretraining scale shapes RL scaling behavior. Our work fills this gap across model sizes from 5M to 1B parameters.

\section{Implementation Details}
\label{appendix:implementation-details}

\subsection{Datasets}
\label{sec:appendix_datasets}

\textbf{Pretraining datasets.} We collect Blitz and Rapid games played on Lichess in 2022. Unless otherwise noted, all games start from the standard chess initial position and follow standard legal-move rules. To ensure game quality, we filter out games shorter than 10 plies and sample games to balance the average Elo rating of the two players from 800 to 3000, yielding a corpus of 54B tokens.

\textbf{Post-training datasets.}
We construct the post-training dataset from Lichess puzzles, retaining only puzzles with average player Elo above 800 and popularity above 100 to improve data quality. The resulting dataset contains 156K puzzles. We partition puzzles into five Elo-based difficulty bins, denoted as B1 through B5 in increasing order of difficulty with Elo ranging
from 800 to 3000. We further balance the dataset by solution length and theme coverage using a greedy sampler. In our experiments, we sample 33K puzzles uniformly from Elo 800 to 2400 for SFT training and 62.3K puzzles with an easy-skewed distribution (70\% from B1 and B2, 30\% from B3 to B5) for RL training.

\textbf{Test datasets.}
Table~\ref{tab:chess_eval_datasets} presents details of the test benchmark. For the Lichess puzzle subsets, we first assign puzzles to Elo bins and remove puzzles with solution lines longer than 12 moves. Within each bin, we use an approximate greedy sampler to balance theme coverage and solution length. Each candidate is ranked by the average current frequency of its theme tags in the selected set, with ties broken by the frequency of its solution length. We select lower-frequency candidates first and update these counts after each batch, yielding difficulty-stratified test sets with broader coverage over tactical themes and solution depths.

\textbf{Decontamination.} To prevent contamination, we discard from the pretraining corpus any game whose trajectory passes through a position that also appears in the post-training or test sets. For each game we replay its move sequence from the initial position and compare every reached position against the set of post-training and test board states, matching on a normalized FEN consisting of the piece placement and side to move. Any game with a match is removed in full. For the post-training and test sets, which are puzzles defined by a starting position, we deduplicate on the starting board position. 

\begin{table}[htbp]
\centering
\small
\caption{\textbf{Test benchmark details.} The Lichess puzzle subsets evaluate tactical performance across Elo difficulty, puzzle themes, and solution depths.}
\begin{tabular}{lrl}
\toprule
\textbf{Dataset} & \textbf{Size} & \textbf{Description} \\
\midrule
Puzzle \texttt{B1} & 308 & Lichess puzzles with Elo rating in $(800,1200]$. \\
Puzzle \texttt{B2} & 298 & Lichess puzzles with Elo rating in $(1200,1600]$. \\
Puzzle \texttt{B3} & 267 & Lichess puzzles with Elo rating in $(1600,2000]$. \\
Puzzle \texttt{B4} & 287 & Lichess puzzles with Elo rating in $(2000,2400]$. \\
Puzzle \texttt{B5} & 320 & Lichess puzzles with Elo rating above $2400$. \\
\bottomrule
\end{tabular}
\label{tab:chess_eval_datasets}
\end{table}

\subsection{Models}
\label{sec:appendix_models}

Table~\ref{tab:model_architectures} presents the architecture details of our model family. We do not tie the input and output embeddings. In addition, motivated by prior work showing the effectiveness of deeper designs for small language models~\citep{liu2024mobilellm}, we use relatively deep architectures at each model scale.

  \begin{table}[htbp]                                                              
  \centering                                                                                                              
  \caption{\textbf{Model architecture details.} All models use a Qwen-style
  decoder-only Transformer architecture with grouped-query attention.}
  \begin{tabular}{lrrrrrrr}                                                        
  \toprule                                                                         
  \textbf{Model} & \textbf{Params} & \textbf{Layers} & \textbf{Hidden Size} &
  \textbf{Intermediate Size} & \textbf{Heads (Q / KV)} & \textbf{Head Size}  \\    
  \midrule                                                        
  5M   & 5.1M   & 6  & 256  & 768  & 2 / 2   & 128 \\                              
  10M  & 11.8M  & 6  & 384  & 1024 & 4 / 4   & 128 \\                              
  20M  & 20.5M  & 6  & 512  & 1536 & 4 / 4   & 128 \\                              
  32M  & 31.6M  & 8  & 512  & 1536 & 8 / 4   & 128 \\                              
  50M  & 47.3M  & 12 & 512  & 1536 & 8 / 4   & 128 \\                              
  100M & 101.6M & 12 & 768  & 2304 & 12 / 4  & 128 \\             
  200M & 203.1M & 24 & 768  & 2304 & 12 / 4  & 128 \\                              
  410M & 411.3M & 28 & 1024 & 3072 & 16 / 4  & 128 \\                              
  680M & 678.6M & 30 & 1280 & 3840 & 20 / 4  & 128 \\                              
  1B   & 1.03B  & 32 & 1536 & 4608 & 24 / 4  & 128 \\                              
  \bottomrule                                                                      
  \end{tabular}                                                                    
  \label{tab:model_architectures}                                                  
  \end{table} 

\subsection{Algorithms}
\label{app:algorithm}

\textbf{SFT objective.} Let $M_t\in\{0,1\}$ be the loss mask, where $M_t=1$ if $w_t$ belongs to the synthetic trace or a model move, and $M_t=0$ otherwise. The SFT objective is
$
\mathcal{L}_{\mathrm{SFT}}(\theta_{\mathrm{sft}})
=
\mathbb{E}_{(s,w)\sim\mathcal{D}_{\mathrm{SFT}}}
\left[
-\sum_{t=1}^{|w|} M_t \log \pi_{\theta_{\mathrm{sft}}}(w_t\mid s,w_{<t})
\right]$.

\textbf{RL algorithm.} We optimize the policy using GRPO as formalized below. For each state $s_0$, we sample a group of trajectories $\zeta_1,\dots,\zeta_G\sim\pi_{\theta_{\mathrm{old}}}(\cdot\mid s_0)$, compute rewards $r_i=R(\zeta_i,s_0)$, and normalize rewards within the group to obtain advantages $A_i=\frac{r_i-\mathrm{mean}(\{r_j\}_{j=1}^{G})}{\mathrm{std}(\{r_j\}_{j=1}^{G})}$. The GRPO objective is {\footnotesize \begin{align*}
\mathcal{J}_{\mathrm{GRPO}}(\theta_{\mathrm{rl}})
={}&
\mathbb{E}_{s_0,\{\zeta_i\}_{i=1}^{G}}
\Bigg[
\frac{1}{G}\sum_{i=1}^{G}
\frac{1}{|o_i|}\sum_{t=1}^{|o_i|}
\min\!\Big(
\rho_{i,t}A_i,\,
\operatorname{clip}\bigl(\rho_{i,t},1-\epsilon,1+\epsilon\bigr)A_i
\Big)
\Bigg] -
\beta D_{\mathrm{KL}}
\!\left(
\pi_{\theta_{\mathrm{rl}}}\,\|\,\pi_{\mathrm{ref}}
\right).
\end{align*} } where $\rho_{i,t}=\frac{\pi_{\theta_{\mathrm{rl}}}(\zeta_{i,t}\mid s_0,\zeta_{i,<t})}{\pi_{\theta_{\mathrm{old}}}(\zeta_{i,t}\mid s_0,\zeta_{i,<t})}$ is the importance sampling weight and $\beta$ is the KL coefficient.
 
\subsection{Training Configurations}
\label{sec:appendix_training}

\paragraph{Pretraining configurations.} All models are pretrained for one epoch over their assigned pretraining-token budget. All model scales share the optimization setup listed in Table~\ref{tab:pretraining_config}.

\paragraph{SFT training configurations.}
We initialize from the corresponding pretrained checkpoint and extend the context length to 3072 tokens. SFT is performed on 4 NVIDIA H100 GPUs for 3 epochs using AdamW with learning rate $3\times10^{-4}$, weight decay $0.01$, and Adam betas $(0.9,0.95)$. The learning rate follows a cosine schedule with a warmup ratio of $0.01$ and a minimum learning rate of $1\times10^{-5}$. We use an effective batch size of 524,288 tokens per optimizer step, apply gradient clipping with maximum norm $1.0$, and train with standard cross-entropy loss. During SFT, prompt tokens are masked from the loss, so optimization is applied only to the response tokens.

\paragraph{RL training configurations.} All experiments are implemented using the \texttt{verl} framework~\citep{sheng2024hybridflow} with hyperparameters: learning rate $10^{-5}$, KL coefficient $\beta = 0.001$, clip ratio $\epsilon=0.2$ and no entropy regularization. We set group size $G = 8$. For response sampling, we fix the sampling temperature $1.0$ and a maximum response length of $3072$ tokens. In \texttt{verl}, we set both the training batch size and mini-batch size to 256 prompts. A simple rule-based reward function is used, assigning reward $1$ to correct answers and $0$ otherwise, without incorporating any format-related signals. RL is performed on 8 NVIDIA H200 GPUs.

\begin{table}[t]
\centering
\small
\caption{\textbf{Pretraining configurations.}}
\begin{tabular}{ll}
\toprule
\textbf{Configuration} & \textbf{Value} \\
\midrule
Hardware & 8 NVIDIA H200 GPUs \\
Context length & 1024 \\
Epochs & 1 \\
Optimizer & AdamW \\
Peak learning rate & $1\times10^{-3}$ \\
Minimum learning rate & $1\times10^{-4}$ \\
Learning-rate schedule & Cosine decay \\
Warmup ratio & $0.05$ \\
Adam betas & $(0.9, 0.95)$ \\
Weight decay & $0.1$ \\
Dropout & $0.0$ \\
Max gradient norm & $1.0$ \\
Per-device batch size & 32 sequences \\
Gradient accumulation steps & 2 \\
Effective batch size & 512 sequences \\
Tokens per optimizer step & 524,288 \\
\bottomrule
\end{tabular}
\label{tab:pretraining_config}
\end{table}

\subsection{Puzzle Game Example}

\begin{figure}
    \centering
    \includegraphics[width=0.8\linewidth]{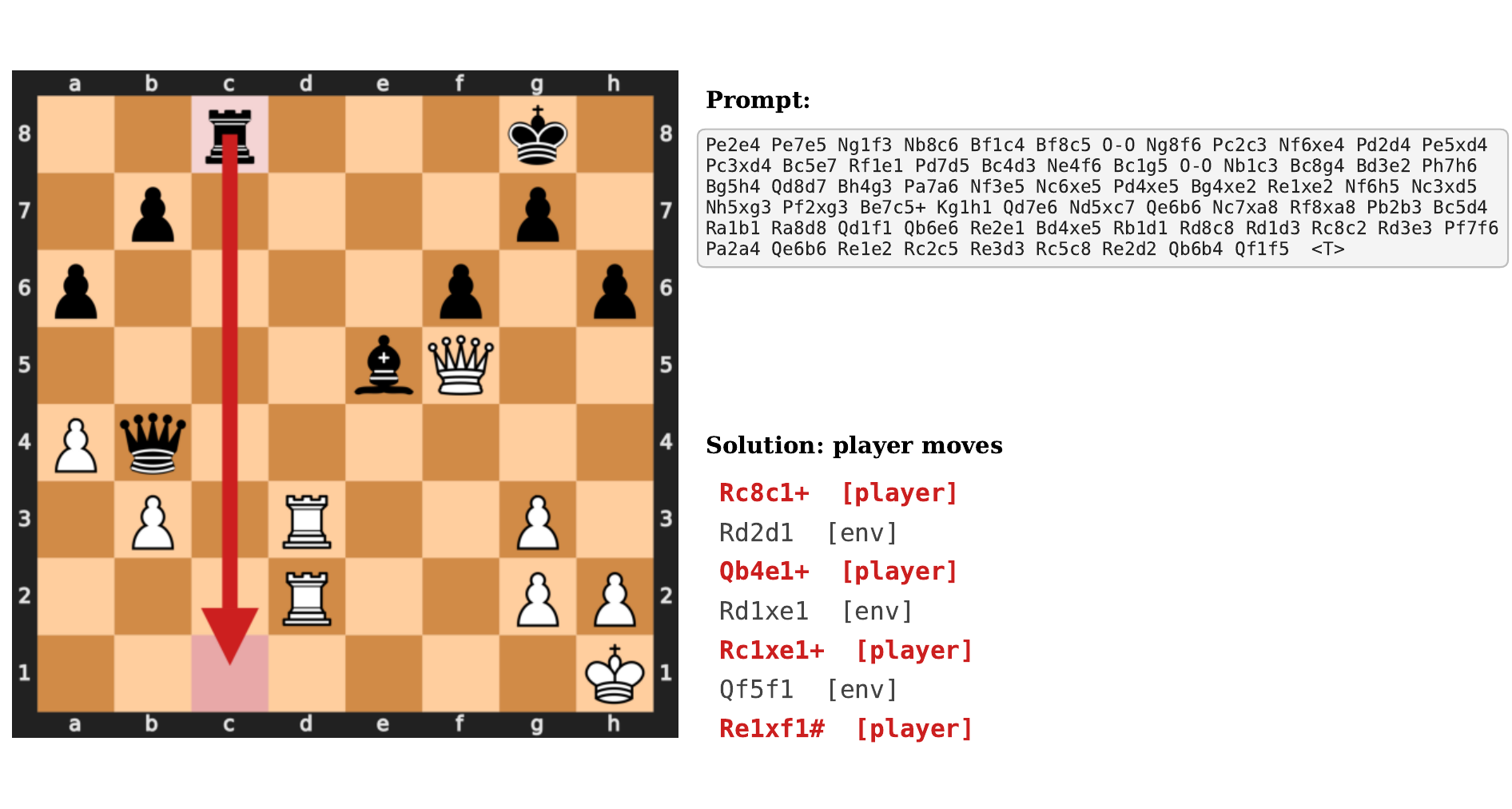}
    \caption{\textbf{An example of a chess puzzle.} The prompt consists of the move sequence representing the board state, optionally followed by a <T> token that triggers reasoning. The player is required to figure out all correct moves in the continuation to solve this puzzle. The best solution line is unique for all puzzle games.}
    \label{fig:puzzle_game_example}
\end{figure}

Fig.~\ref{fig:puzzle_game_example} shows an example of a chess puzzle game. Each puzzle provides a starting position, which we convert into a move sequence for the model. To solve the puzzle, the player must identify every correct move in the continuation.

\newcommand{\tok}[1]{\texttt{\footnotesize #1}}

\begin{table}[t]
\centering
\caption{An example of reasoning trace generated by the model. Given an input prompt, the model generates several possible continuations, each containing both the player’s and the opponent’s moves. These continuations are concatenated using the <sep> token. The model is then expected to generate the final answer move, ideally by selecting the best move from the reasoning traces. The tokens <T> and </T> indicate the reasoning format.}
\label{tab:chess_reasoning_example}
\renewcommand{\arraystretch}{1.35}
\begin{tabularx}{\linewidth}{>{\bfseries}p{0.16\linewidth} X}
\hline
Input &
{\ttfamily\footnotesize
Pe2e4 Pc7c5 Ng1f3 Nb8c6 Pd2d4 Pc5xd4 Bf1c4 Pg7g6 Nf3g5 Pe7e6 Pc2c3 Pd7d5 Pe4xd5 Pe6xd5 Qd1f3 Ng8f6 Bc4xd5 Nf6xd5 <T>
}
\\
\hline
Reasoning &
{\ttfamily\footnotesize
Ng5h3 Bf8g7 \textless sep\textgreater{} Ng5h3 Nc6e5 Qf3xf7\# Ne5xf7 \textless sep\textgreater{} Ng5h3 Bc8g4 Qf3xf7\# \textless sep\textgreater{} Ng5h3 Qd8d6 Bc1f4 Qd6e7+ \textless sep\textgreater{} Ng5h3 \textless sep\textgreater{} Ng5xf7 Nd5e7 Bc1f4 Rh8g8 Nf7xd8 Ne7f5 Qf3xc6+ \textless sep\textgreater{} Bc1f4 Nd5xf4 Qf3xf4 Bc8f5 Qf4e5+ Bf8e7 \textless sep\textgreater{} Bc1f4 Bf8g7 Ng5xf7 Rh8f8 \textless sep\textgreater{} Bc1f4 Bf8e7 Bf4e5 \textless sep\textgreater{} Qf3xf7\# \textless sep\textgreater{} O-O Nd5f6 Nc6e5 Qf7xd5 Ne5c6 \textless sep\textgreater{} Qf3xf7\# \textless sep\textgreater{} Pg2g3 Qd8f6 Qf3xf6 Rh8g8 \textless sep\textgreater{} Pg2g3 Qd8e7+ Ke1f1 Bc8e6 Ng5xe6 Qe7xe6 \textless sep\textgreater{} Qf3xf7\# \textless/T\textgreater{}
}
\\
\hline
Answer &
{\ttfamily\footnotesize
Qf3xf7\#
}
\\
\hline
\end{tabularx}
\end{table}

\subsection{Additional Details of Synthetic Reasoning Trace Generation}
\label{app:synthetic_trace}

\textbf{Synthetic trace construction.}
We visualize the procedure in Fig.~\ref{fig:overview}. Starting from an input board $s_0$, we first use a proposal policy $p_{\theta_{\mathrm{prop}}}$ to generate $K$ rollout traces $\tau_1,\dots,\tau_K \sim p_{\theta_{\mathrm{prop}}}(\cdot \mid s_0)$, where $K$ is controlled by the budget. The proposal policy is either the model pretrained in Section~\ref{sec:pretraining} or a policy constructed from a chess engine. For each rollout $\tau_k$, we sample a length budget $L_k$ and generate moves until either the game terminates or the sampled budget $L_k$ is reached.

We then convert each rollout $\tau_k$ into a move sequence $(a_{k,1},o_{k,1},\dots,)$ by discarding invalid or unparsable suffixes. The resulting move sequences are merged by shared prefixes into a prefix tree $\mathcal{G}$ rooted at $s_0$. Each non-root node $u$ is labeled by a move $a_u$, and its board state $s_u$ is obtained by applying the moves along the path from the root to $u$. In principle, a verifier can assign scores to nodes, which may be used to rank, prune, or truncate $\mathcal{G}$ under the serialization budget. In our implementation, we use the simplest rule: we discard illegal moves, preserve the proposal policy's sampling order when ordering children in the prefix tree, and ensure that the best-of-$K$ continuation is included in the serialized trace.

Finally, we serialize the constructed tree by depth-first traversal (DFS). Let $(u_1,\dots,u_m)$ denote the retained leaf nodes of $\mathcal{G}$ ordered by DFS. For each leaf node $u_j$, let $\tilde{\tau}_{u_j}$ denote the rendered move sequence along the path from the root to $u_j$. The final synthetic reasoning trace is obtained by concatenating these root-to-leaf paths in DFS order:
$r=\texttt{<T>}\;\tilde{\tau}_{u_1}\;\texttt{<sep>}\;\tilde{\tau}_{u_2}\;\texttt{<sep>}\;\cdots\;\tilde{\tau}_{u_m}\;\texttt{</T>}$.

After producing the synthetic trace, the model is trained to commit to a single continuation. We choose this continuation from the same rollout set used to construct the trace. Specifically, among the valid parsed continuations, we select the best one according to a verifier, such as Stockfish\footnote{\url{https://github.com/official-stockfish/stockfish}}, and use it as the supervised answer following the trace. Let the selected continuation be $\tau^\star=(a_1,o_1,a_2,o_2,\dots,a_H)$, where $a_t$ is the model move and $o_t$ is the opponent move. We train on the concatenated sequence $w=(r,\tau^\star)$. Since opponent moves will be produced by the environment rather than by the model policy during inference, we mask the opponent-move tokens in $\tau^\star$ out of the loss and apply supervision only to $r$ and model moves in $\tau^\star$. Let $M_t\in\{0,1\}$ be the loss mask, where $M_t=1$ if $w_t$ belongs to the synthetic trace or a model move, and $M_t=0$ otherwise. The SFT objective is
$
\mathcal{L}_{\mathrm{SFT}}(\theta_{\mathrm{sft}})
=
\mathbb{E}_{(s,w)\sim\mathcal{D}_{\mathrm{SFT}}}
\left[
-\sum_{t=1}^{|w|} M_t \log p_{\theta_{\mathrm{sft}}}(w_t\mid s,w_{<t})
\right]$.

\textbf{Implementation details for synthesizing reasoning traces.} Following the procedure described in Section~\ref{sec:sft}, we use a fixed proposal policy, a pretrained 200M model, to generate synthetic reasoning traces. 
To diversify the reasoning paths, for each puzzle, we branch at the first move: we sample $n = 8$ first-move candidates from the policy model. From each candidate we then roll out continuations under the same policy, where the number of traces per candidate $m$ and the trajectory length $l$ are drawn from rounded log-normal distributions truncated to $[2, 10]$ and $[1, 20]$, with means $8$ and $5$ and shape parameters $\sigma = 0.2$ and $\sigma = 0.4$, respectively. The two means roughly correspond to $1/3$ of the legal action space and the typical puzzle solution length in the SFT dataset. We did not tune these values.

\section{FLOP Estimation}
\label{sec:appendix_flops_accounting}

We estimate training compute in model FLOPs using the standard dense language
model approximation
\begin{equation}
    C_{\mathrm{train}}(N,T) \approx 6NT,
    \label{eq:train_flops}
\end{equation}
where $N$ is the number of active model parameters and $T$ is the number of
tokens processed during training. This convention is widely used in language
model scaling analyses \citep{kaplan2020scaling, hoffmann2022training}.

This approximation is appropriate for the dense Qwen3 models used in our
experiments because they are dense decoder-only Transformer language models:
all non-embedding Transformer parameters are active for every processed token.
Thus, their dominant training cost comes from dense matrix multiplications in
attention and feed-forward layers, which is precisely the regime targeted by
the $6NT$ approximation. We use $N$ as the active dense parameter count of the
trained model.

\paragraph{Pretraining.}
Let $T_{\mathrm{pt}}$ denote the number of pretraining tokens. We estimate
pretraining FLOPs as $C_{\mathrm{pt}}=6N T_{\mathrm{pt}}$.

\paragraph{Supervised fine-tuning.}
Let $\mathcal{D}_{\mathrm{sft}}$ denote the SFT dataset, let $\ell_i$ denote
the tokenized length of example $i$, and let $E_{\mathrm{sft}}$ denote the
number of SFT epochs. The total number of SFT training tokens is
$T_{\mathrm{sft}}=E_{\mathrm{sft}}\sum_{i \in \mathcal{D}_{\mathrm{sft}}} \ell_i$.
Therefore, we estimate SFT FLOPs as $C_{\mathrm{sft}}=6N E_{\mathrm{sft}}
    \sum_{i \in \mathcal{D}_{\mathrm{sft}}} \ell_i$.
In our experiments, all SFT examples are padded or truncated to a fixed context
length $L_{\mathrm{sft}}$. Thus, if $n_{\mathrm{sft}}$ examples are used per
epoch, the estimate simplifies to $C_{\mathrm{sft}}=6N E_{\mathrm{sft}} n_{\mathrm{sft}} L_{\mathrm{sft}}$.

\paragraph{Reinforcement learning.}
For GRPO, let $n_{\mathrm{prompt}}$ denote the total number of training prompts processed
over the entire RL stage and let $g$
denote the group size, i.e., the number of sampled responses per prompt. For prompt $q$, let $L_{\mathrm{prompt}}^{(q)}$ denote the prompt
length and let $L_{\mathrm{resp}}^{(q,i)}$ denote the response length of sample $i$. Define
$L_{\mathrm{rl}}^{(q,i)}=L_{\mathrm{prompt}}^{(q)} + L_{\mathrm{resp}}^{(q,i)}$.
The total number of rollout tokens is then computed as
\[
T_{\mathrm{rollout}} = \sum_{q=1}^{n_{\mathrm{prompt}}}
\sum_{i=1}^{g}
L_{\mathrm{rl}}^{(q,i)} .
\]

We estimate RL compute as the sum of policy rollout generation, reference-model
log-probability evaluation, and policy optimization. Policy rollout generation
and reference-model evaluation each require one forward pass over the rollout
tokens, while policy optimization is estimated using the standard training FLOP
approximation. Thus,
\begin{align*}
C_{\mathrm{rl}}
&=
2N T_{\mathrm{rollout}}
+
2N T_{\mathrm{rollout}}
+
6N T_{\mathrm{rollout}} \\
&=
10N T_{\mathrm{rollout}} \\
&=
10N
\sum_{q=1}^{n_{\mathrm{prompt}}}
\sum_{i=1}^{g}
L_{\mathrm{rl}}^{(q,i)} .
\end{align*}
If no reference model is used, the second $2N T_{\mathrm{rollout}}$ term is
omitted and $C_{\mathrm{rl}}$ is reduced to $8NT_{\mathrm{rollout}}$.

\section{Pretraining Law Fitting}
\label{sec:appendix_chichilla_fit}

\begin{figure}
    \centering
    \includegraphics[width=\linewidth]{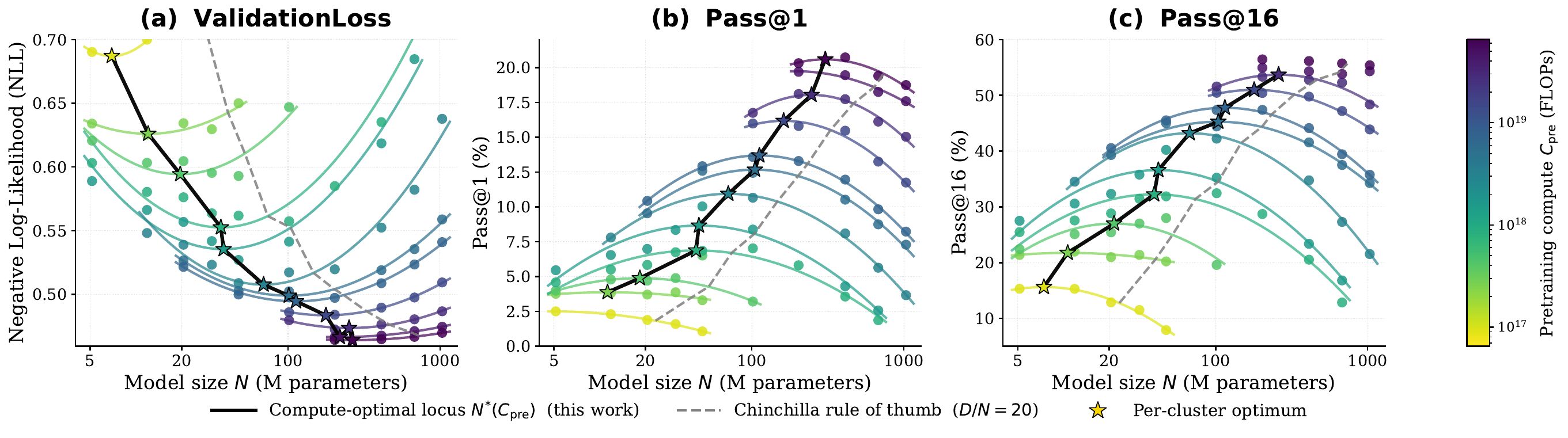}
    \caption{\textbf{IsoFLOP curves for pretrained models across a sweep of 11 compute budgets and 10 model sizes.} (a) Pretraining validation loss (NLL) on held-out human games. (b) Pass@1 and (c) Pass@16 on the puzzle benchmark. The starred points mark the compute-optimal model size at each budget. The solid line connects these optima, while the dashed line represents the Chinchilla allocation (D/N = 20). Over the plotted range, the chess-optimal locus lies at smaller models and more tokens than this rule, with the two converging at higher compute.}
    \label{fig:pretrain_scaling}
\end{figure}

For pretraining, we sweep 11 compute budgets from $6.5 \times 10^{16}$ to $6.5 \times 10^{19}$ FLOPs across 10 model sizes, corresponding to training runs of approximately $200$M to $52$B tokens, following the setup of isoFLOP methodology in prior scaling-law studies~\citep{hoffmann2022training}. Fig.~\ref{fig:pretrain_scaling} reports isoFLOP curves on validation loss on held-out human games, along with pass@$1$ and pass@$16$ on the downstream puzzle benchmark (including all puzzles from B1 to B5). Note that for pretraining evaluation, all models just generate move sequences without reasoning. We observe a clear valley in validation loss at each budget, confirming that an optimal parameter-token allocation exists under fixed FLOPs. 
Comparing this optimum to the Chinchilla-style compute-optimal allocation for natural language modeling (grey dashed line in Fig.~\ref{fig:pretrain_scaling}), our results indicate that chess pretraining favors smaller models trained on more tokens than the language-modeling baseline over the compute range studied. 
Moreover, the isoFLOP optima for pass@1 and pass@16 on the puzzle benchmark closely track the validation-loss optimum on human games. Within each model size, downstream performance continues to improve with additional pretraining over the compute range we study.

To complement the IsoFLOP analysis, we refit the parametric form of \citet{hoffmann2022training} to our pretraining sweep:
\begin{equation}
L(N, D) = E + \frac{A}{N^{\alpha}} + \frac{B}{D^{\beta}},
\label{eq:chinchilla}
\end{equation}
where $N$ is the number of total parameters, $D$ is the number of training tokens, and $(E, A, B, \alpha, \beta)$ are fitted parameters. We optimize the Huber loss between predicted and observed validation loss using L-BFGS over $n = 64$ $(N, D, L)$ triples drawn from our sweep, restricting to runs with validation loss below 1.0 to exclude undertrained outliers.

Table~\ref{tab:chinchilla_fit} reports the fitted parameters alongside the \citet{hoffmann2022training} values for language modeling. First, the parameter exponent $\alpha = 0.4006 \pm 0.0399$ remains close to the language-modeling estimate ($\alpha \approx 0.34$), suggesting that returns to model scale are similar across domains. Second, the data exponent $\beta = 0.6789 \pm 0.0291$ is more than twice the language-modeling estimate ($\beta \approx 0.28$), indicating that loss falls substantially faster with additional training tokens in the chess setting holding model size fixed. The RMSE is 0.0097 for the fitting. The implied compute-optimal allocation,
\begin{equation}
N_{\text{opt}}(C) \propto C^{\beta/(\alpha+\beta)} = C^{0.63}, \quad D_{\text{opt}}(C) \propto C^{\alpha/(\alpha+\beta)} = C^{0.37},
\end{equation}
implies that the compute-optimal model size grows more rapidly than the token budget. This is consistent with the IsoFLOP curves (Fig.~\ref{fig:pretrain_scaling}): over the compute range we study, the chess optima use more tokens per parameter than the $D/N=20$ rule, but this gap narrows as compute grows.

\begin{table}[h]
\centering
\caption{\textbf{Chinchilla functional form parameters fit to our chess pretraining sweep, compared to the language-modeling estimates of \citet{hoffmann2022training}.}}
\label{tab:chinchilla_fit}
\begin{tabular}{lcc}
\toprule
Parameter & Chess (ours) & Language~\citep{hoffmann2022training} \\
\midrule
$E$ & $0.412 \pm 0.009$ & — \\
$\alpha$ & $0.401 \pm 0.040$ & $\approx 0.34$ \\
$\beta$ & $0.679 \pm 0.029$ & $\approx 0.28$ \\
$\beta / (\alpha + \beta)$ & 0.63 & $\approx 0.46$ \\
\bottomrule
\end{tabular}
\end{table}

\section{SFT Performance Comparisons}
\label{app:sft_appendix}

\begin{figure}
    \centering
    \includegraphics[width=\linewidth]{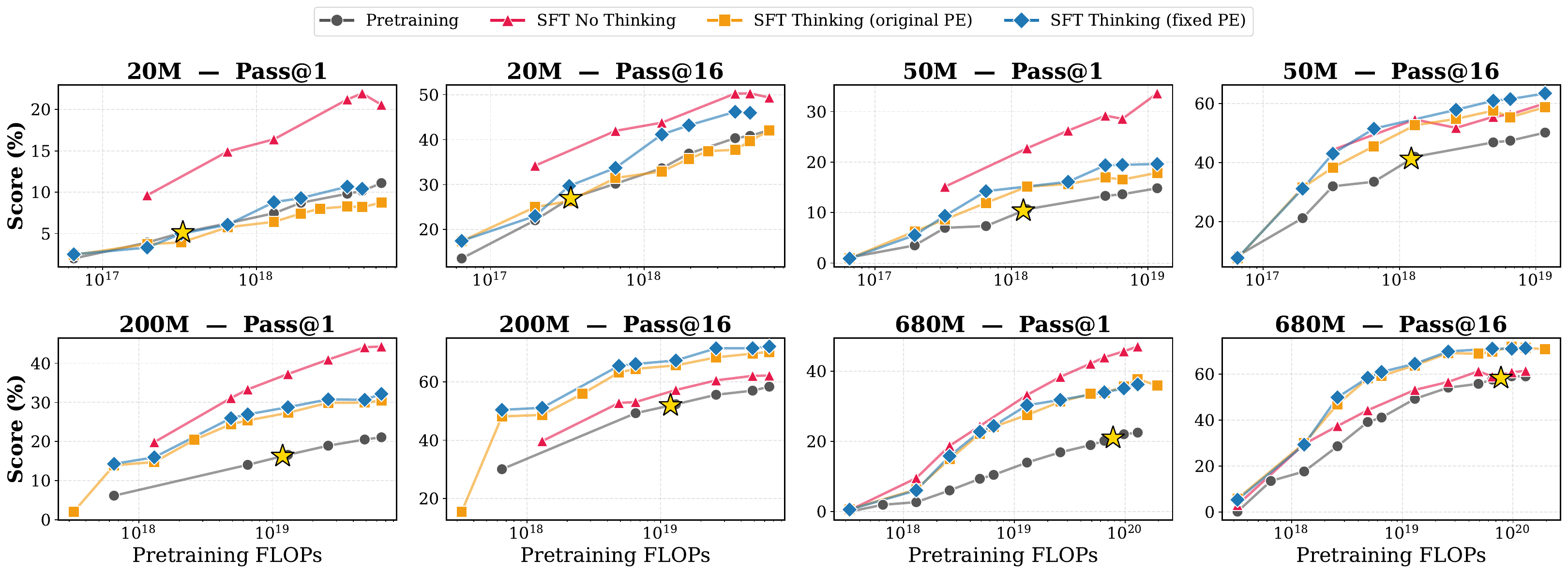}
    \caption{\textbf{Effect of SFT with and without reasoning traces across four model sizes.} Each panel shows puzzle benchmark performance (B1--B5, macro-averaged) as a function of pretraining compute. The star marks the compute-optimal pretraining budget from the isoFLOP fit. The blue line shows results obtained with an alternative positional encoding implementation.}
    \label{fig:comparison_pre_sft_rl}
\end{figure}

For SFT, we compare two settings: (1) \emph{SFT without reasoning} directly fine-tunes the model on the target move sequence, with the loss masked on opponent moves; (2) \emph{SFT with reasoning traces} trains the model on synthetic intermediate reasoning traces followed by the target answer. We train all SFT variants using the same number of puzzle samples across models. Fig.~\ref{fig:comparison_pre_sft_rl} compares performance across model sizes and pretraining FLOPs. For a fixed model size, stronger pretrained models consistently achieve higher post-SFT performance, indicating that pretraining quality transfers to SFT. SFT without reasoning traces improves pass@1 at every scale, and for small models (20M, 50M) it also improves pass@$16$. At larger scales (200M, 680M), however, its pass@16 gain over the pretrained baseline shrinks, while SFT with reasoning traces gives the highest pass@16, especially on harder puzzles. 
This suggests that learning on reasoning traces can enrich the broader distribution of candidates the model can generate. 
In the paper, we use the same SFT at every model size for consistency.

To better understand why SFT with reasoning traces underperforms SFT without reasoning on pass@$1$, we examine two aspects of our training setup. First, our current CoT format requires the model to predict both player and opponent moves. While the player's moves in a puzzle are optimal, the opponent's moves are not necessarily optimal. The model may therefore consider valid alternative continuations that do not match the prescribed puzzle continuation. Our ablation intervenes during generation by replacing the model-generated opponent moves with the ground-truth opponent moves and allowing the model to continue reasoning from the edited trace. This intervention improves pass@$1$. RL may improve the model's ability to consider multiple possible branches for opponent moves. Second, although both settings use the same number of training updates, answer tokens constitute a much smaller fraction of the supervised tokens in the reasoning-trace setting and therefore receive less weight under token-averaged loss. This difference in effective answer supervision may also contribute to the gap. We leave the exploration of alternative reasoning-trace formats and data-generation choices, such as using different proposal models, to future work.

\textbf{Implementation note.} In our current SFT experiments, the intended RoPE extension was not applied, requiring the models to learn to adapt positions beyond the pretraining context length during SFT. 
Fig.~\ref{fig:comparison_pre_sft_rl} shows that using the positional encoding extension implementation improves SFT performance, particularly for smaller models.


\section{Joint Pretraining--RL Scaling Law}
\label{app:joint-rl-scaling-law}

This appendix describes our statistical model used to relate pretraining scale, pretraining loss, and downstream RL compute.  The goal is to obtain a simple predictive law for the reward obtained after RL, conditioned on a pretrained model with parameter count $N$, pretraining tokens $T$, and RL compute $C_{\mathrm{RL}}$ as we discuss in Section~\ref{sec:exp_pre_to_rl_scaling}.
\subsection{Interpretation of the Local RL Scaling Fit}
\label{app:local_rl_scaling_fit}

Prior work~\citep{khatri2025art} models the performance of a fixed model under RL compute with a sigmoid compute--performance curve:
\begin{align}
R^{\mathrm{sig}}(C)-R_0
=
(A-R_0)
\frac{1}{1+(C_{\mathrm{mid}}/C)^{\gamma}},
\label{eq:prior_sigmoid}
\end{align}
where \(R_0\) is the initial performance, \(A\) is the asymptotic reward ceiling, \(C_{\mathrm{mid}}\) controls the transition location, and \(\gamma\) controls the sigmoid steepness. This form explicitly models saturation, but fitting the full curve requires sufficiently long RL runs to identify both the transition point and the plateau. In our experiments, running every configuration long enough to reach the plateau regime would require substantially more RL compute, and in our available compute range we found the asymptote \(A\) to be weakly identified. 
We therefore focus on the local, non-saturated regime, which is also scientifically useful: it isolates how pretraining scale affects RL improvement before the curve is dominated by asymptotic saturation.

Let \(x=\log_{10} C\) and \(x_{\mathrm{ref}}=\log_{10} C_{\mathrm{ref}}\). A first-order Taylor expansion of Eq.~\eqref{eq:prior_sigmoid} around \(x_{\mathrm{ref}}\) gives
\[
R^{\mathrm{sig}}(x)
\approx
R^{\mathrm{sig}}(x_{\mathrm{ref}})
+
\left.
\frac{dR^{\mathrm{sig}}}{dx}
\right|_{x=x_{\mathrm{ref}}}
(x-x_{\mathrm{ref}}).
\]
Thus, locally, the sigmoid reduces to a log-linear form:
\begin{align*}
R_{N,T}(C)
=
R^{\mathrm{ref}}_{N,T}
+
B_{N,T}
\left(
\log_{10} C-\log_{10} C_{\mathrm{ref}}
\right),
\label{eq:local_log_linear}
\end{align*}
where \(R^{\mathrm{ref}}_{N,T}=R_{N,T}(C_{\mathrm{ref}})\), and \(B_{N,T}\) is the local reward slope.
To interpret the local slope, differentiate Eq.~\eqref{eq:prior_sigmoid} with respect to \(x=\log_{10}C\). This gives
\[
\frac{dR^{\mathrm{sig}}}{dx}
=
\gamma\ln(10)(A-R_0)
\frac{(C_{\mathrm{mid}}/C)^\gamma}
{\left[1+(C_{\mathrm{mid}}/C)^\gamma\right]^2}.
\]
Using
\[
\frac{R^{\mathrm{sig}}(C)-R_0}{A-R_0}
=
\frac{1}{1+(C_{\mathrm{mid}}/C)^\gamma},
\qquad
A-R^{\mathrm{sig}}(C)
=
(A-R_0)
\frac{(C_{\mathrm{mid}}/C)^\gamma}
{1+(C_{\mathrm{mid}}/C)^\gamma},
\]
the local slope can be rewritten as
\begin{align*}
\frac{dR^{\mathrm{sig}}}{dx}
=
\gamma\ln(10)
\frac{R^{\mathrm{sig}}(C)-R_0}{A-R_0}
\bigl(A-R^{\mathrm{sig}}(C)\bigr).
\end{align*}
Therefore, at the reference point,
\begin{align}
B_{N,T}
\approx
\gamma\ln(10)
\frac{R^{\mathrm{ref}}_{N,T}-R_0}{A-R_0}
\bigl(A-R^{\mathrm{ref}}_{N,T}\bigr).
\label{eq:raw_slope_decomposition}
\end{align}

We note that $B_{N,T}$ is a local estimate and can therefore be sensitive to the portion of the RL trajectory used for fitting. For a fixed ($\gamma$), the local slope follows a concave quadratic: it is small when the reference point is close to either the initial performance ($R_0$) or the saturation level ($A$), and largest in the intermediate regime. Consequently, runs evaluated very early in RL or after approaching saturation can yield systematically smaller and less predictable slope estimates. On the B3-B4 benchmarks, most models remain in the non-saturated regime, making these benchmarks better suited to our slope analysis. We examine these effects empirically in Section~\ref{app:param}. 

\subsection{Parameterizations}
\label{app:param}

We now describe the parameterizations used in the joint pretraining--RL scaling law. 
For each pretrained model indexed by model size $N$ and pretraining tokens $T$, we first fit its RL trajectory as a log-linear function of RL compute:
\begin{align}
R_{N,T}(C_{\mathrm{RL}})
=
R_{N,T}^{\mathrm{ref}}
+
B_{N,T}
\left(
\log_{10} C_{\mathrm{RL}} - \log_{10} C_{\mathrm{ref}}
\right),
\end{align}
where \(R_{N,T}^{\mathrm{ref}}\) is the reward at the reference RL compute level \(C_{\mathrm{ref}}\), and \(B_{N,T}\) is the reward gain per decade of RL compute. 
In practice, we fit the log-linear form with least-squares regression.
We then parameterize the fitted quantities
\(\phi_{N,T}=\{R_{N,T}^{\mathrm{ref}}, B_{N,T}\}\)
using pretraining properties: model size \(N\), number of pretraining tokens \(T\), and pretraining validation loss \(L_{\mathrm{pt}}(N,T)\).

\begin{figure}
    \centering
\includegraphics[width=0.8\linewidth]{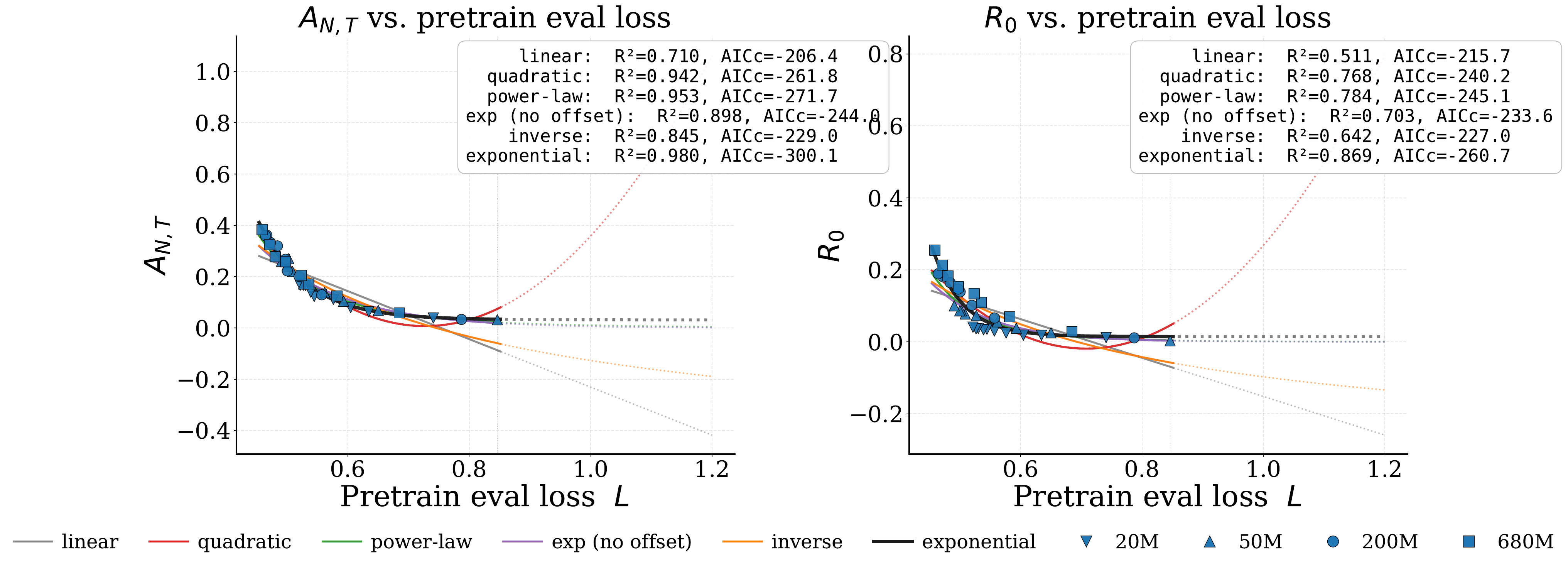}
    \caption{\textbf{Comparison of different parameterization choices for $f$.}}
    \label{fig:a_fitting}
\end{figure}

\begin{figure}
    \centering
    \includegraphics[width=1.0\linewidth]{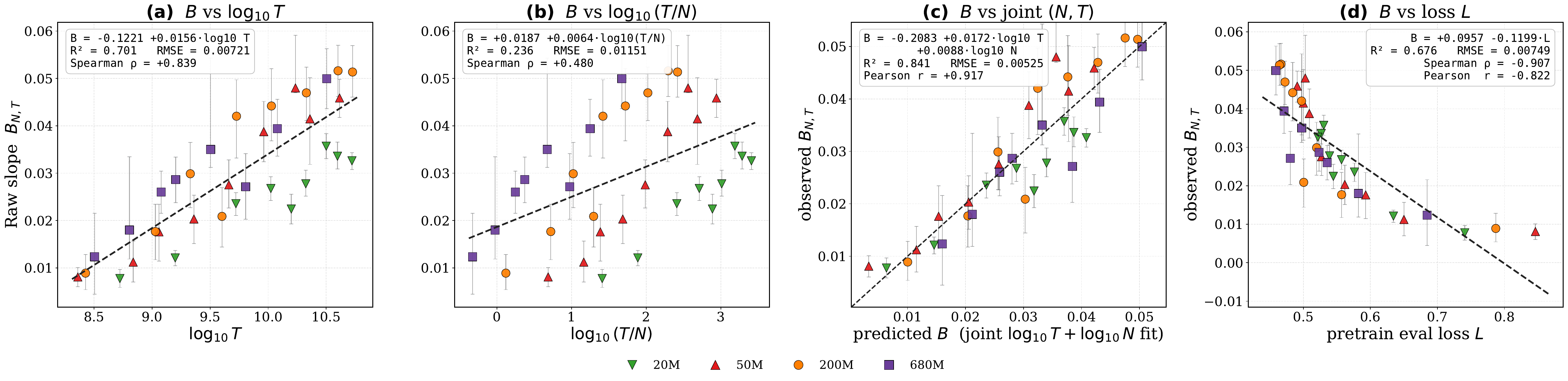}
    \caption{\textbf{Comparison of token-only, tokens-per-parameter, loss-based, and joint fits for the RL slope on the B3--B4 benchmark.} We observe a clear monotonic relationship between log-scale pretraining tokens and the RL slope, with a Spearman correlation of $0.84$. The joint fit achieves a lower RMSE and higher $R^2$, with a substantially 2$\times$ larger coefficient on log-scale tokens than on model size. Pretraining loss is also a reasonably strong predictor of the slope.}
    \label{fig:b_fitting}
\end{figure}

\begin{figure}
    \centering
    \includegraphics[width=1.0\linewidth]{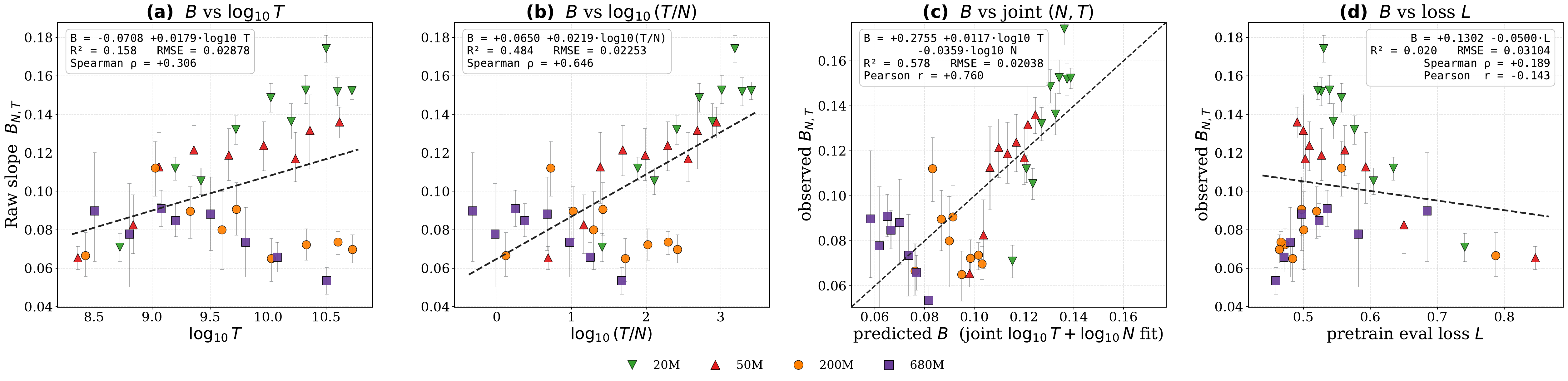}
    \caption{\textbf{Comparison between token-only, token-per-parameter and loss fitting for RL slope on B1 benchmark.} On the easiest benchmark, larger and more extensively pretrained models, such as the 200M and 680M models, quickly approach saturation, resulting in low local slope estimates.}
    \label{fig:b_fitting_b1}
\end{figure}

\begin{figure}
    \centering
    \includegraphics[width=1.0\linewidth]{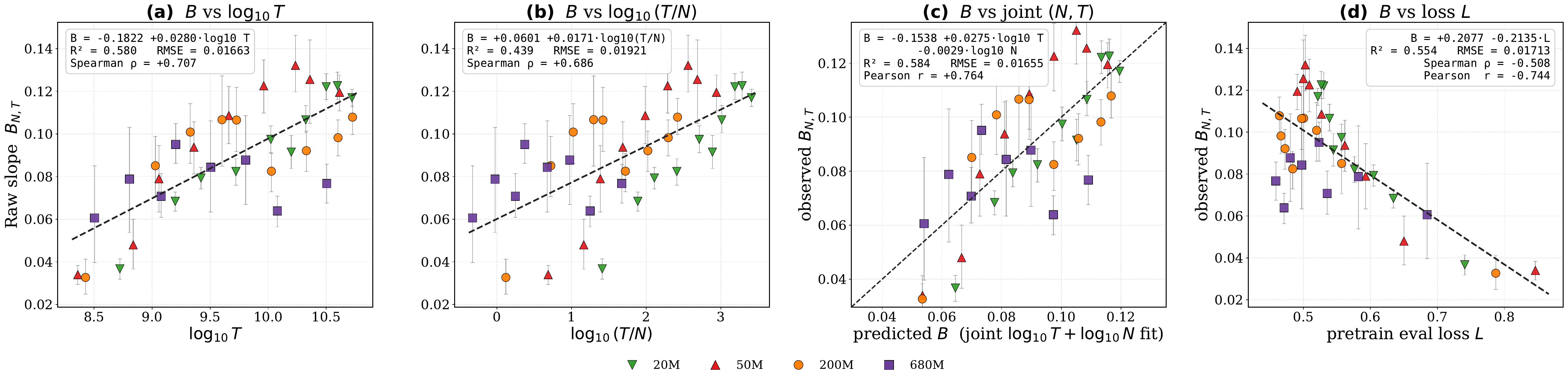}
    \caption{\textbf{Comparison of token-only, tokens-per-parameter, and loss-based fits for the RL slope on the B2 benchmark.} On B2, the 200M and 680M models also exhibit some degree of saturation, and pretraining loss is a weaker predictor of the local RL slope.}
    \label{fig:b_fitting_b2}
\end{figure}

We first examine the empirical dependence of \(R_{N,T}^{\mathrm{ref}}\) on pretraining validation loss. 
As discussed in Section~\ref{sec:exp_pre_to_rl_scaling}, \(R_{N,T}^{\mathrm{ref}}\) is strongly monotonic in \(L_{\mathrm{pt}}\), and this relationship becomes tighter as the RL compute slice increases. 
We compare alternative functional forms in Fig.~\ref{fig:a_fitting}. 
Among the candidate forms considered, we adopt an exponential parameterization, which provides the best tradeoff between fit quality, simplicity, and extrapolation behavior:
\begin{align}
f(L_{\mathrm{pt}})
=
\alpha_f + \beta_f \exp(-\gamma_f L_{\mathrm{pt}}),
\qquad \gamma_f > 0 .
\end{align}
This form is monotone decreasing in pretraining loss and avoids the pathological behavior of linear or quadratic fits outside the observed range. 
Moreover, since the Chinchilla loss surface includes an irreducible-loss term, \(L_{\mathrm{pt}}\) is bounded below; consequently, \(f(L_{\mathrm{pt}})\) has a finite ceiling rather than growing without bound.

Next, we parameterize the RL slope \(B_{N,T}\).
We compare token-only, token-per-parameter and a joint form over $N$ and $T$. 
As illustrated in the benchmark-specific results (Fig.~\ref{fig:b_fitting_b1} and Fig.~\ref{fig:b_fitting_b2}), stronger pretrained models on easier benchmarks have already been close to saturation, leading to small estimated slopes. 
At the other extreme, runs observed only during the initial phase of RL may not yet exhibit a stable local scaling trend. 
These effects introduce noise into both the slope estimates and their downstream prediction.
We observe a clear monotonic relationship between log-scale pretraining tokens and the RL slope on B3-B4 benchmarks (Fig.~\ref{fig:b_fitting}), with a
Spearman correlation of 0.84.
However, the joint fit achieves a lower RMSE and higher $R^2$.
We also note that pretraining loss predicts well on B3-B4 benchmarks but is substantially less reliable on B2. 
Therefore, the joint form explains the RL slope most robustly:
\begin{align}
g(N,T)
=
\alpha_g + \beta_g \log_{10} T + \gamma_g \log_{10} N.
\end{align}
The fitted coefficient on $\log_{10} T$ is larger than that on $\log_{10} N$, indicating that the RL slope is driven primarily by the amount of pretraining data, while model size provides a weaker correction. 

Combining these two parameterizations yields the final \textbf{joint pretraining--RL scaling law}:
\begin{align}
R(C_{\mathrm{RL}}, N, T)
=
f\!\left(L_{\mathrm{pt}}(N,T)\right)
+
g(N,T)
\left(
\log_{10} C_{\mathrm{RL}} - \log_{10} C_{\mathrm{ref}}
\right).
\label{eq:loss_conditioned_joint_law}
\end{align}

When the observed pretraining validation loss is available, the measured \(L_{\mathrm{pt}}\) can be directly used, which evaluates the quality of the fitted maps \(f\) and \(g\). 
When only the pretraining configuration \((N,T)\) is available, we instead predict \(L_{\mathrm{pt}}(N,T)\) using the Chinchilla loss surface (Eq.~\eqref{eq:chinchilla}) and substitute it into Eq.~\eqref{eq:loss_conditioned_joint_law}. 
This gives a fully predictive scaling law from \((N,T,C_{\mathrm{RL}})\) to post-RL reward.

\subsection{Leave-one-out Fitting Validation}
We validate the joint law with leave-one-out (LOO) prediction over the 36 runs with observed pretraining evaluation losses. 
For each held-out run, we refit $f$ and $g$ on the remaining 35 runs and predict the full held-out RL curve. The Chinchilla loss surface $L(N,T)$ is not refit inside the LOO loop.

\begin{figure}
    \centering
\includegraphics[width=1.0\linewidth]{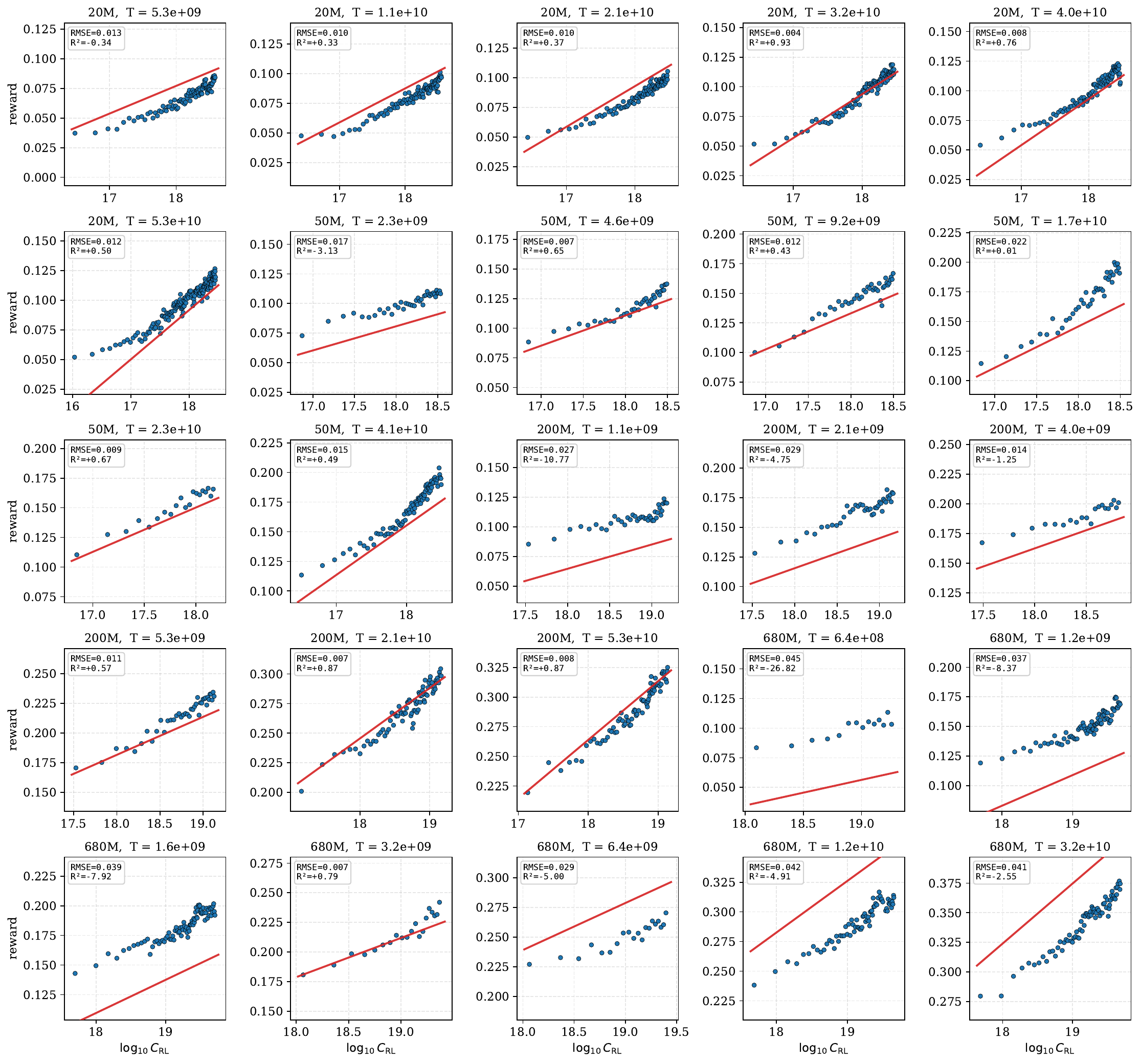}
    \caption{\textbf{LOO validation fitting for existing RL runs with Chinchilla-predicted loss.}}
    \label{fig:loo_chinchilla}
\end{figure}

\begin{figure}
    \centering
\includegraphics[width=1.0\linewidth]{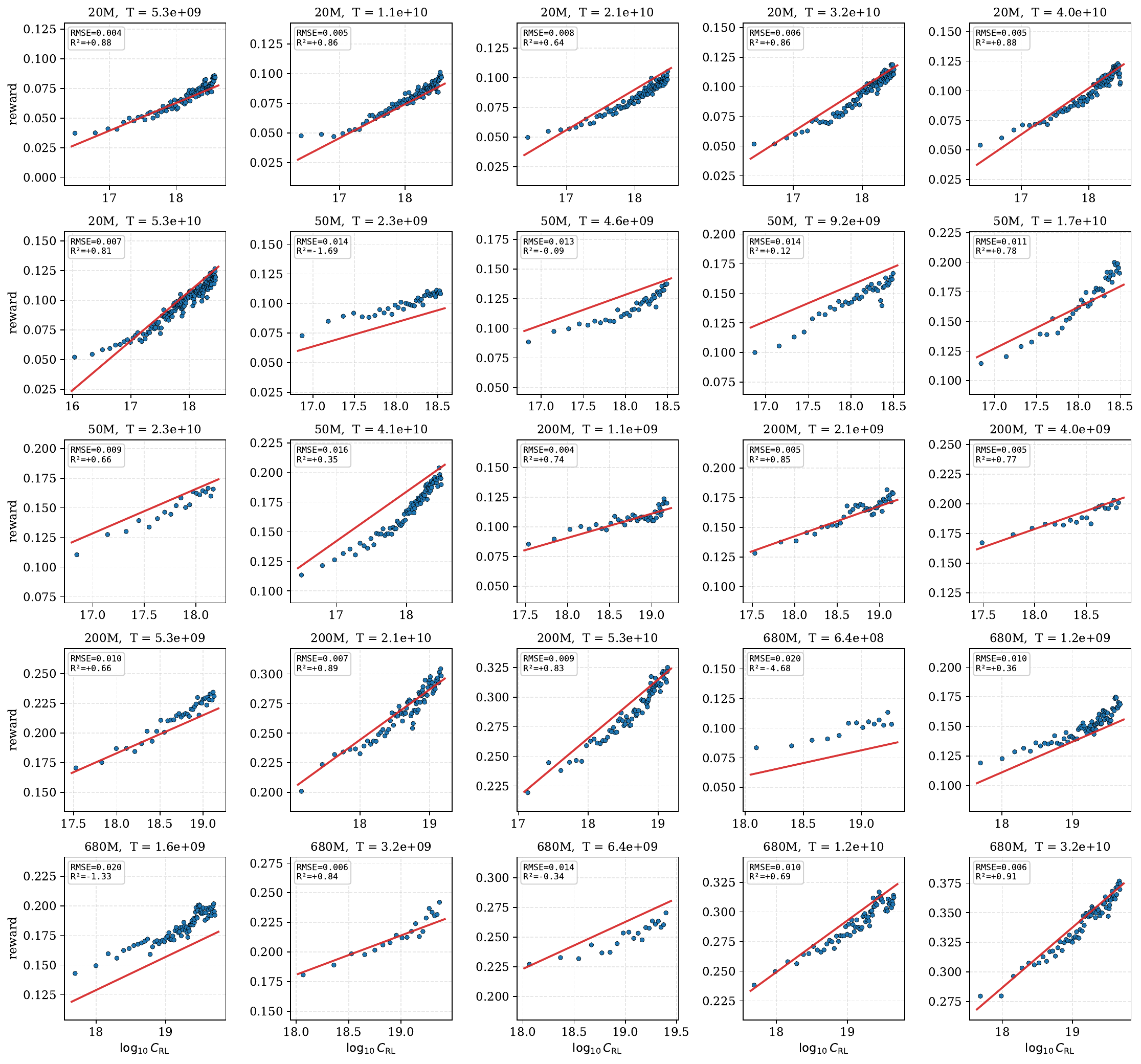}
    \caption{\textbf{LOO validation fitting for existing RL runs with observed loss.}}
    \label{fig:loo_observed}
\end{figure}

We report two validation modes. In the first, called Chinchilla-$L$ LOO, the held-out loss is predicted from $(N,T)$ using Eq.~\eqref{eq:chinchilla}. This corresponds to the practical extrapolation setting where only the proposed pretraining configuration is known. In the second, called observed-$L$ LOO, the held-out run's measured pretraining evaluation loss is supplied to $f$. This isolates the error from the $f$ and $g$ parameterizations, removing the additional error from predicting $L$. The metrics are reported in Table~\ref{tab:loo-validation}. We report qualitative fitting examples for these two modes in Fig.~\ref{fig:loo_chinchilla} and Fig.~\ref{fig:loo_observed}.

\begin{table}[t]
\centering
\caption{\textbf{Aggregate leave-one-out validation metrics. } Chinchilla-$L$ uses predicted pretraining loss from $(N,T)$; observed-$L$ uses the measured held-out pretraining loss.}
\label{tab:loo-validation}
\begin{tabular}{lrrr}
\toprule
Metric & Chinchilla-$L$ & Observed-$L$ & Difference \\
\midrule
Overall LOO RMSE on $R$ & 0.0194 & \textbf{0.0102} & +0.0092 \\
Overall LOO MAE on $R$ & 0.0153 & \textbf{0.0078} & +0.0075 \\
Median per-run $R^2$ & -0.86 & \textbf{+0.65} & -- \\
Mean per-run $R^2$ & -3.76 & \textbf{-0.67} & -- \\
Predicted $R^{\text{ref}}_{N,T}$ vs. actual: Pearson $r$ & +0.972 & $\mathbf{+0.989}$ & -- \\
Predicted $R^{\text{ref}}_{N,T}$ vs. actual: $R^2$ & +0.940 & $\mathbf{+0.977}$ & -- \\
Predicted $B_{N,T}$ vs. actual: Pearson $r$ & +0.890 & +0.890 & -- \\
\bottomrule
\end{tabular}
\end{table}

The absolute error is small: the strict Chinchilla-$L$ mode obtains RMSE $0.0194$ in reward. The negative mean per-run $R^2$ should not be interpreted as a failure of the fit. Some runs have nearly flat observed RL trajectories, so their within-run variance is tiny; dividing by this small variance makes $R^2$ unstable even when the absolute residuals are only around $0.02$--$0.03$. For this reason, RMSE and MAE are more informative than mean per-run $R^2$ in this validation.

\textbf{The comparison between the two LOO modes suggests that pretraining-loss prediction error accounts for a nontrivial fraction of the total error.} Moving from observed-$L$ to Chinchilla-$L$ increases RMSE by $0.0099$, about 49\% of the strict LOO RMSE. The remaining error is attributable to residual variation in the $f$ and $g$ fits, with $g(N,T)$ being the harder component to predict: $f(L)$ explains the reference reward more cleanly than $g(N,T)$ explains the RL slope.

\subsubsection{Leave-one-model-size-out validation}
\label{app:lmso-validation}

To test whether the joint law extrapolates to unseen model sizes, we perform \emph{leave-one-model-size-out} (LMSO) validation: for each of the four measured sizes $\{20\mathrm{M},50\mathrm{M},200\mathrm{M},680\mathrm{M}\}$, we hold out \emph{all} runs of that size and refit the entire pipeline on the remaining data.  
Concretely, we (i)~drop every pretraining row of the held-out size from the Chinchilla data and refit $L(N,T)$; (ii)~refit $f$ and $g$ on the remaining joint-law runs; (iii)~predict each held-out run's $A^{\rm ref}_{N,T}$, $B_{N,T}$, and full RL trajectory using the refit surface.
Table~\ref{tab:lmso-per-fold} reports the per-fold metrics and Table~\ref{tab:lmso-aggregate} compares the LMSO aggregate to the run-level LOO baseline.

\begin{table}[t]
\centering
\caption{\textbf{Per-fold LMSO validation on B3B4.}  For each held-out size, we report the RMSE of the held-out RL predictions, and the refit joint-law coefficients. }
\label{tab:lmso-per-fold}
\begin{tabular}{lrrrrr}
\toprule
Held-out & RMSE($R$) & RMSE($A^{\rm ref}$) & RMSE($B$) & $\beta_g$ ($T$) & $\gamma_g$ ($N$) \\
\midrule
20M   & 0.013 & 0.027 & 0.011 & +0.019 & +0.004 \\
50M   & 0.016 & 0.028 & 0.008 & +0.017 & +0.011 \\
200M  & 0.017 & 0.021 & 0.006 & +0.016 & +0.009 \\
680M & 0.048 & 0.045 & 0.009 & +0.017 & +0.014 \\
\bottomrule
\end{tabular}
\end{table}

\begin{table}[t]
\centering
\caption{\textbf{Aggregate LMSO metrics vs the run-level LOO baseline.}  Aggregated across all 36 held-out run, LMSO on $R$ is only $25\%$ larger than the run-level LOO baseline, and predicted $R^{\rm ref}_{N,T}$ correlates with observed at Pearson $r{=}{+}0.95$.}
\label{tab:lmso-aggregate}
\begin{tabular}{lrr}
\toprule
Metric & Run-level LOO & LMSO \\
\midrule
Overall RMSE on $R$                       & 0.0194 & 0.0242 \\
Overall MAE on $R$                        & 0.0153 & 0.0183 \\
Predicted $R^{\rm ref}_{N,T}$: Pearson $r$ & $+0.972$ & $+0.954$ \\
Predicted $B_{N,T}$: Pearson $r$          & $+0.890$ & $+0.792$ \\
\bottomrule
\end{tabular}
\end{table}

\begin{table}[htbp]
\centering
\small
\caption{\textbf{Fitting results on the out-of-sample model sizes (32M and 410M).}}
\begin{tabular}{c c c c c c}
\toprule
$N$ & $T$ (B tokens) &
$\Delta R^{\rm ref}$ (fit $-$ law) &
$\Delta B$ (fit $-$ law) &
RMSE (obs-$L$) &
RMSE (Chin-$L$) \\
\midrule
32M  & 6.87 & +0.001 & +0.0014 & 0.0060  & 0.0086 \\
32M  & 13.7 & +0.005 & +0.0048 & 0.0054$^{*}$ & 0.0054 \\
410M & 1.98 & +0.019 & +0.0080 & 0.0142  & 0.0315 \\
410M & 2.63 & +0.032 & +0.0083 & 0.0265  & 0.0372 \\
\midrule
Overall & -- & -- & -- & 0.0169 & 0.0232 \\
\bottomrule
\end{tabular}
\label{tab:out-of-sample-fitting}
\end{table}

We interpret the 680M-held-out RMSE as our best empirical estimate of the joint law's extrapolation error at the top of the observed size range.  Predictions beyond 680M (e.g., the 1B and 2B extrapolations in Sec.~\ref{sec:rl_results}) should therefore be treated as extrapolations of at least this magnitude. We also validate the fitting results on two model sizes (32M and 410M) that were not used to fit the joint law and evaluated its predictions on them, obtaining low prediction errors (Table~\ref{tab:out-of-sample-fitting}.

\subsection{Choice of reference compute}
\label{app:cref-sensitivity}

The reference compute $C_{\mathrm{ref}}$ controls where the per-run log-linear RL trajectories are anchored.  Table~\ref{tab:cref-sensitivity} compares several choices.  We use $C_{\mathrm{ref}}=10^{20}$ because it is the best compromise across the two validation modes: it attains the lowest Chinchilla-$L$ LOO RMSE (tied with $10^{19}$) and is within $2\%$ of the best observed-$L$ RMSE, while also yielding a stable fit for $f(L)$.  No single anchor is best on both criteria, and the choice matters little for the RL-share trend, which varies by only a few points across this range.

\begin{table}[t]
\centering
\caption{\textbf{Sensitivity to the reference compute $C_{\mathrm{ref}}$.} The selected value is $\log_{10}C_{\mathrm{ref}}=20$.  Parameters are for the offset-exponential form $f(L)=\gamma_f+\alpha_f\,e^{-\beta_f L}$; the offset $\gamma_f$ is positive at every anchor, so the fitted reference reward stays above zero at large loss.}
\label{tab:cref-sensitivity}
\begin{tabular}{rrrrrrr}
\toprule
$\log_{10}C_{\mathrm{ref}}$ & $\alpha_f$ & $\beta_f$ & $\gamma_f$ & $R^2_f$ & LOO RMSE, Chinchilla-$L$ & LOO RMSE, observed-$L$ \\
\midrule
18 & 214.95 & 14.58 & 0.0198 & 0.972 & 0.0214 & 0.0159 \\
19 & 157.50 & 13.57 & 0.0257 & 0.981 & 0.0201 & 0.0125 \\
20 & $\mathbf{129.85}$ & $\mathbf{12.85}$ & $\mathbf{0.0314}$ & $\mathbf{0.980}$ & $\mathbf{0.0201}$ & $\mathbf{0.0102}$ \\
21 & 114.82 & 12.32 & 0.0368 & 0.975 & 0.0213 & 0.0100 \\
\bottomrule
\end{tabular}
\end{table}

\subsection{Post-SFT pass@$k$ as an auxiliary validation signal}
\label{app:sft-passk}

\begin{figure}
    \centering
    \includegraphics[width=0.6\linewidth]{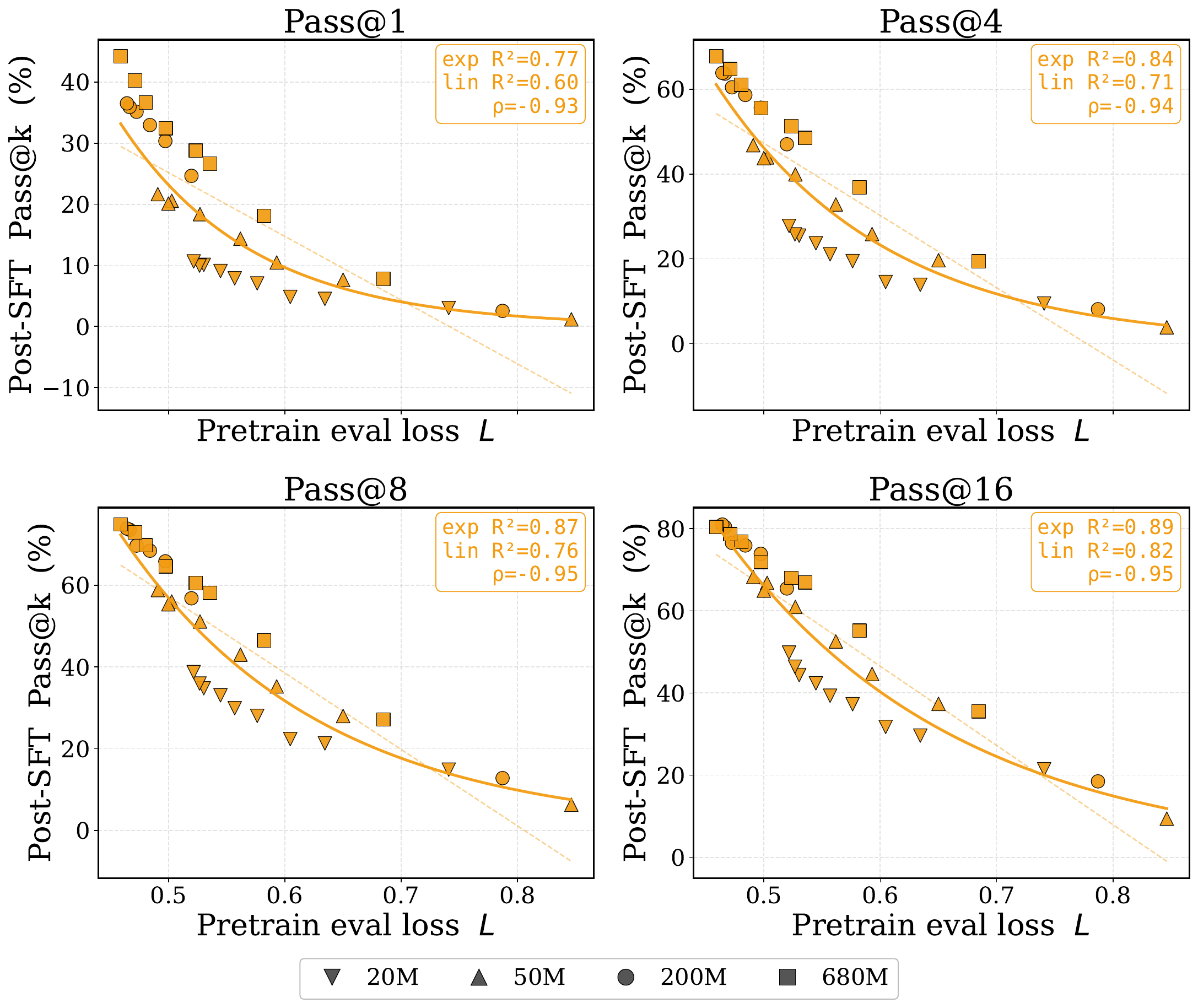}
    \caption{\textbf{SFT pass@$k$ vs. loss curves.}}
    \label{fig:sft_passk_vs_loss}
\end{figure}

In Fig.~\ref{fig:sft_passk_vs_loss}, we also evaluate whether post-SFT pass@$k$ metrics show the same dependence on pretraining loss.  Across 36 runs (the thinking-track SFT models), pass@$k$ is well fit by an exponential function of pretraining evaluation loss, with stronger fits at larger $k$.  This supports the interpretation that pretraining evaluation loss is a useful summary variable not only for the post-RL reward reference point, but also for post-SFT downstream capability.

\begin{figure}
    \centering
    \includegraphics[width=0.55\linewidth]{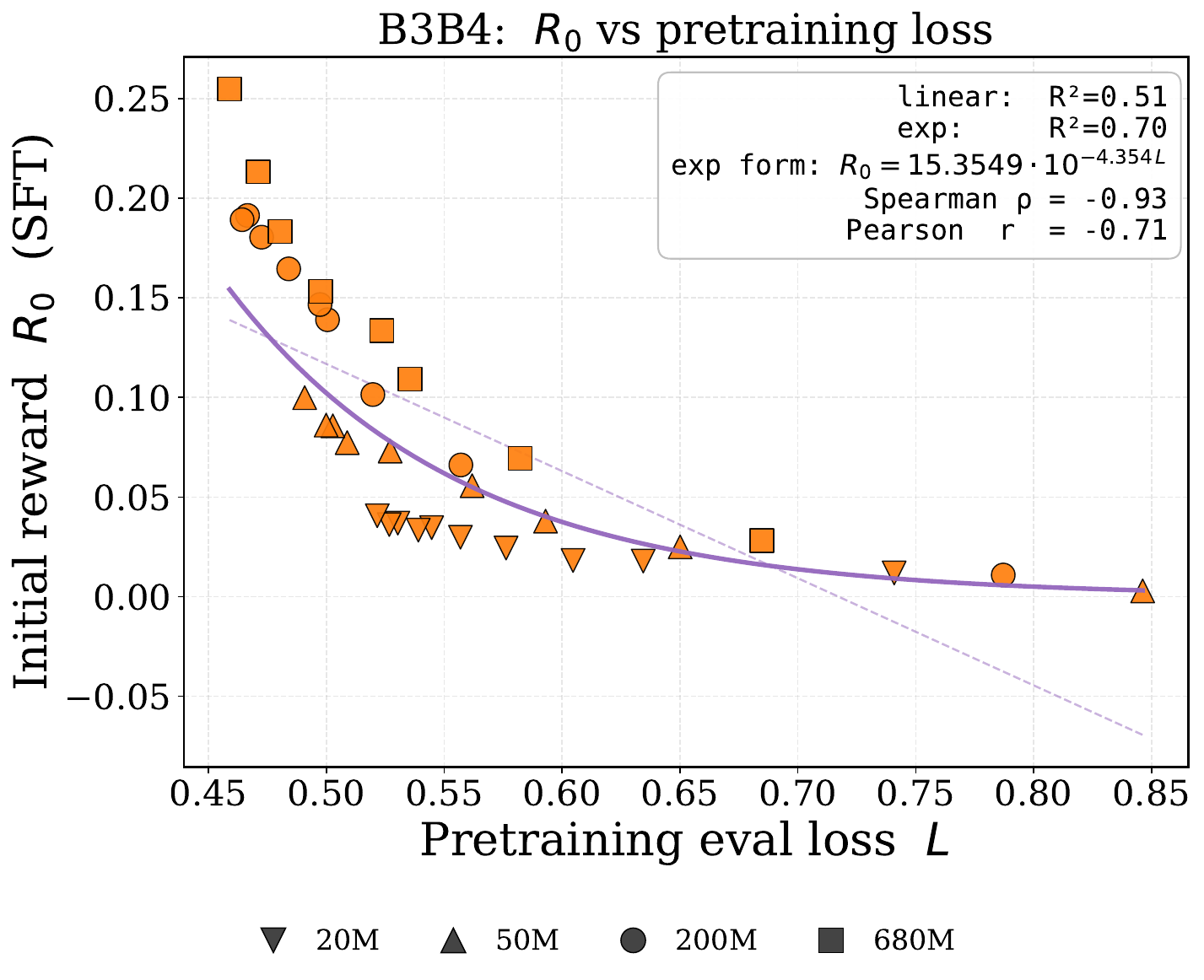}
    \caption{\textbf{Initial (post-SFT) reward $R_0$ versus pretraining eval loss $L$ on the B3B4 benchmark.}  Each marker is one of the 36 runs used in the joint-law fit.  The exponential fit $R_0 = 14.62\cdot 10^{-4.30\,L}$ (solid) attains $R^2 = 0.70$; a linear fit (dashed) attains $R^2 = 0.51$.  Spearman $\rho = -0.92$.  This fit is used as the physical floor $R \geq R_0(L)$ in the frontier optimiser.}
    \label{fig:R0_vs_L}
\end{figure}

The SFT-baseline reward $R_0$ itself (the pre-RL reward mean on the benchmark) shows the same qualitative dependence on $L$.  Fig.~\ref{fig:R0_vs_L} plots $R_0$ against pretraining eval loss for the B3B4 population: an exponential fit gives $R^2 = 0.70$ (with Spearman $\rho = -0.92$, near-perfectly monotone), meaningfully outperforming a linear fit ($R^2 = 0.51$).  We also use this fit as the physical floor $R \geq R_0(L)$ in the frontier optimisation of Sec.~\ref{sec:exp_pre_to_rl_scaling}, preventing the log-linear RL extrapolation from predicting rewards below the SFT baseline.

\begin{table}[t]
\centering
\caption{\textbf{Exponential fits of post-SFT pass@$k$ versus pretraining evaluation loss.}  The decay rates are broadly consistent with the reference reward and SFT-baseline reward fits. The pretraining loss predicts pass@$k$ with larger $k$ better.}
\label{tab:sft-passk}
\begin{tabular}{lrrrr}
\toprule
Metric & $\alpha$ & $\beta_{10}$ & $R^2$ & Spearman $\rho$ \\
\midrule
pass@1 & 1838 & $-3.80$ & 0.77 & $-0.93$ \\
pass@4 & 1421 & $-2.98$ & 0.84 & $-0.94$ \\
pass@8 & 1060 & $-2.54$ & 0.87 & $-0.95$ \\
pass@16 & 789 & $-2.15$ & 0.89 & $-0.95$ \\
\bottomrule
\end{tabular}
\end{table}

\subsection{Asymptote Ceiling Fitting}
\label{app:20m_ceiling}
Following \citet{khatri2025art}, we also estimate the asymptote of RL improvement by fitting a per-run logistic curve,
\begin{align*}
R(C) = R_0 + (A_\infty - R_0)\sigma\bigl(\gamma(\log_{10} C - \log_{10} C_{\rm mid})\bigr),
\end{align*}
with $R_0$ fixed at the SFT baseline and $(A_\infty, \log_{10} C_{\rm mid}, \gamma)$ estimated by weighted nonlinear least squares; 90\% confidence intervals are obtained from a 500-sample parametric bootstrap. We restrict this fit to the 20M-parameter family on B1-B4 benchmark, the only sweep whose RL trajectories enter the saturation regime within our compute budget. Resolving $A_\infty$ for larger models would demand roughly an order of magnitude more RL compute per run.

\begin{figure}
    \centering
    \includegraphics[width=1.0\linewidth]{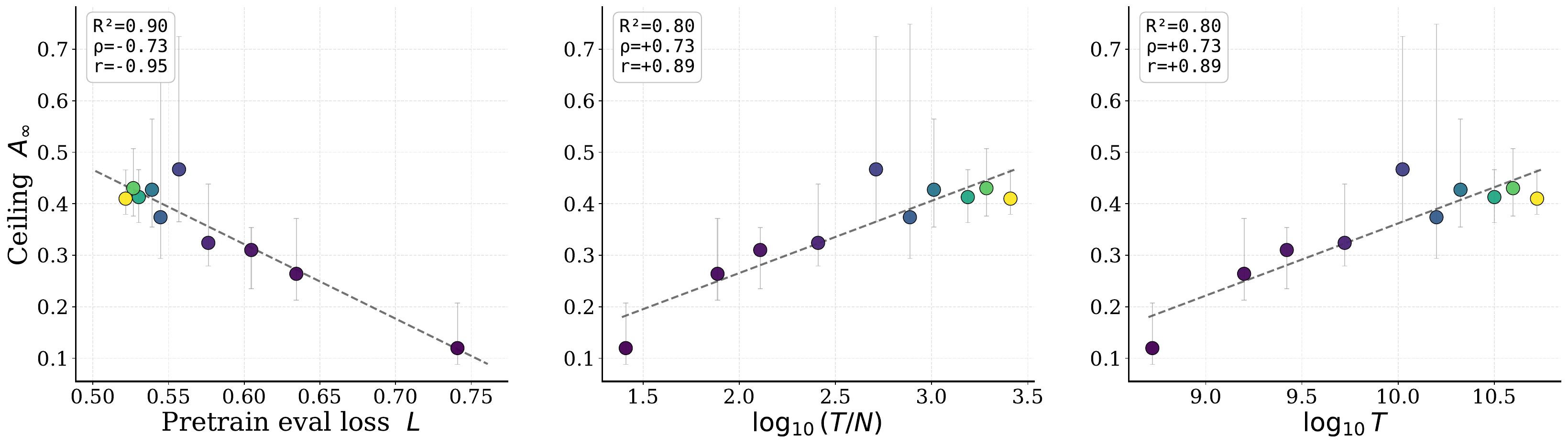}
    \caption{\textbf{Asymptote $A_\infty$ of the per-run logistic RL learning curve for the 20M-parameter family, plotted against three pretraining covariates.} Each point is one of the ten 20M runs; markers are coloured by the pretrain-tokens fraction $\alpha$, vertical bars are 90\% parametric-bootstrap CIs (500 resamples per run), and the dashed line is an ordinary-least-squares fit on the displayed sample. \textbf{(left)} ceiling versus pretrain eval loss $L$: better-pretrained checkpoints support strictly higher RL asymptotes (Spearman $\rho{=}{-}0.73$, $R^2{=}0.90$). \textbf{(middle, right)} ceiling versus $\log_{10}(T/N)$ and $\log_{10} T$: both show a positive trend ($\rho{=}{+}0.73$, $R^2{=}0.80$ each)}
    \label{fig:20m_ceiling}
\end{figure}

Across the ten 20M runs the estimated ceiling spans $A_\infty \in [0.12, 0.47]$ and is well predicted by the pretraining eval loss $L$ (Spearman $\rho = -0.73$, linear $R^2 = 0.90$): SFT initialisations from better-pretrained checkpoints support strictly higher RL asymptotes. The relationship with $\log_{10}(T/N)$ is weaker but goes in the same direction ($\rho = +0.73$, $R^2 = 0.80$), as expected if $L$ is the proximal mediator of both effects. Fig.~\ref{fig:20m_ceiling} visualises these dependencies together with per-run bootstrap CIs.

\subsection{Extrapolating the Compute-Optimal Frontier}

\label{sec:sim-frontier}
The fitted law lets us score \emph{any} hypothetical training recipe without
running it. A recipe is fully specified by a triple $(N, T, C_{\mathrm{RL}})$:
the model size $N$, the number of pretraining tokens $T$, and the RL compute
$C_{\mathrm{RL}}$. 

\paragraph{Per-size frontier.}
For a fixed model size $N$ and a total-compute budget $C$, the recipe still has
one free degree of freedom: how to split $C$ between pretraining
($T$) and RL ($C_{\mathrm{RL}}$). We resolve it by maximizing the predicted
reward subject to the budget constraint,
\begin{equation}
  R^\star_N(C) \;=\;
  \max_{T,\,C_{\mathrm{RL}}}\;
  \widehat{R}(N,T,C_{\mathrm{RL}})
  \quad\text{s.t.}\quad
  C_{\mathrm{tot}}(N,T,C_{\mathrm{RL}}) = C,
  \label{eq:persize}
\end{equation}
sweeping $T$ over a dense grid and solving for the residual
$C_{\mathrm{RL}} = C - 6NT - C_{\mathrm{SFT}}$. This traces out the
compute-optimal reward curve of a single model size as a function of its total
budget.

\begin{figure}
    \centering
    \includegraphics[width=0.5\linewidth]{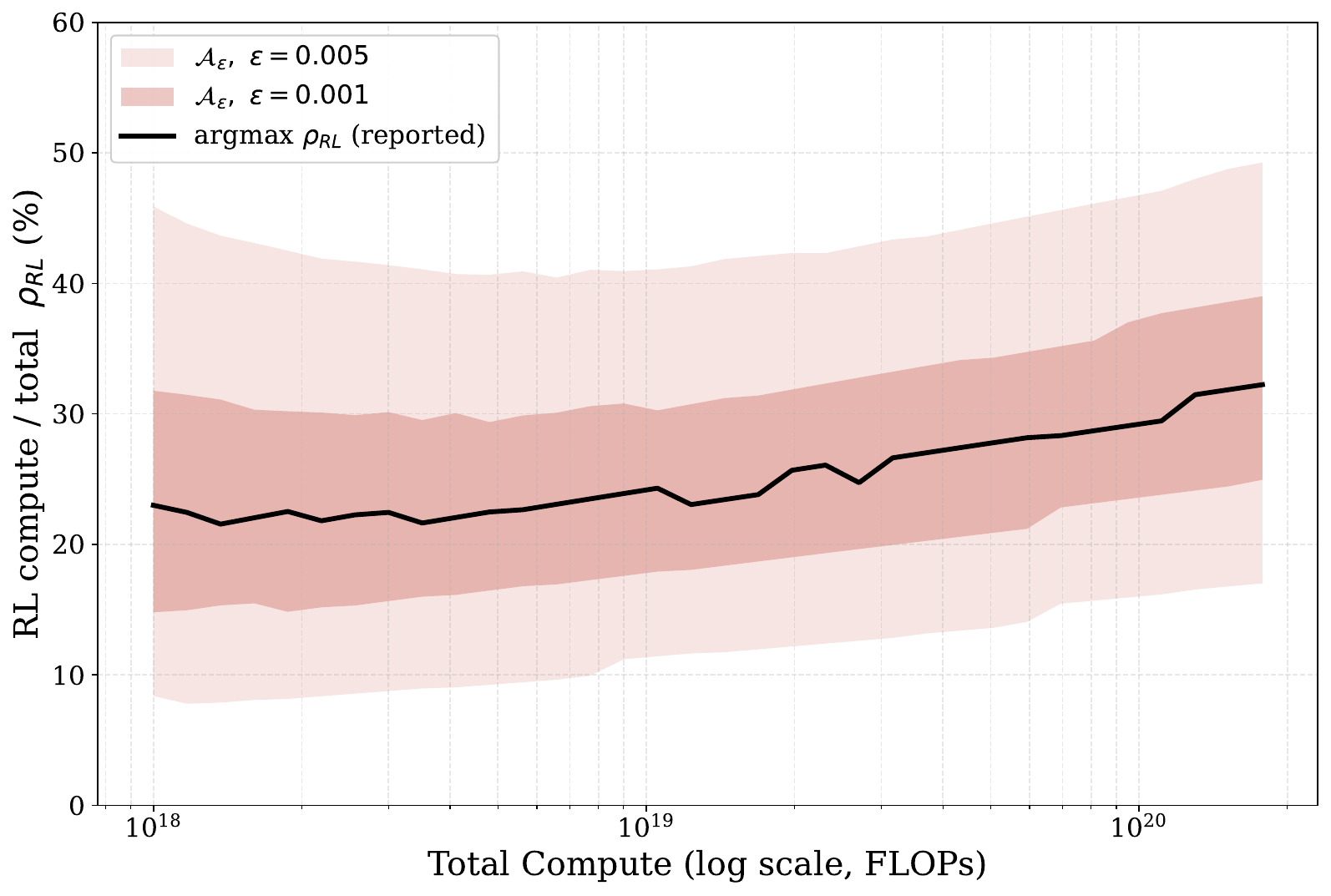}
    \caption{\textbf{$\epsilon$-sensitivity of the compute-optimal RL fraction.} Black: argmax $\rho_{\mathrm{RL}}$ along the continuous-$N$ optimum. Shaded: the spread of $\rho_{\mathrm{RL}}$ over allocations within $\epsilon$ reward of the optimum, $\mathcal{A}\epsilon = {\widehat{R} \ge \widehat{R}{\max}-\epsilon}$, for $\epsilon=0.001$ (dark) and $0.005$ (light). The reward surface is flat near its peak; however, the whole band shifts upward with compute, so the increasing-RL-share trend is robust.}
    \label{fig:rho_epsilon_band}
\end{figure}

We identify the optimum by grid search over allocations for which $N$, $T$, and the implied $C_{\mathrm{RL}}$ lie within the ranges supported by the fitted pretraining and RL scaling laws. For each value of $C_{\mathrm{tot}}$, we first evaluate approximately 400 feasible allocations on a coarse grid in $(\log_{10} N,\log_{10} T)$. Since the candidate set is finite, exhaustive enumeration identifies the global maximizer over this grid. We then evaluate a finer grid around the best-performing region of the coarse grid. We report the sensitivity of the frontier in Fig.~\ref{fig:rho_epsilon_band}. This refinement reduces discretization error but does not provide an additional guarantee of global optimality over the unrestricted continuous domain. The reported solution is therefore optimal over the evaluated candidate set and within the empirical support of the fitted scaling laws. We do not interpret it as a guarantee of global optimality under extrapolation beyond the compute ranges considered. 

\paragraph{Global frontier.}
We evaluate \eqref{eq:persize} over a ladder of model sizes, i.e., the sizes we
actually trained, augmented with hypothetical fill-in sizes, and take the
upper envelope over $N$ at each budget:
$R^\star(C) = \max_N R^\star_N(C)$. The size attaining the maximum
changes as the budget grows, so the global frontier is a sequence of
\emph{takeovers} in which progressively larger models become compute-optimal.
For each frontier point we also record the compute-optimal RL share
$\rho_{\mathrm{RL}} = C_{\mathrm{RL}}^\star / C$, which quantifies how the budget
should be divided between pretraining and RL along the frontier. Taking the continuous limit of the size ladder (optimizing $N$ jointly with $T$) yields the smooth continuous-$N$ optimum, which we compute at each budget with the Nelder--Mead simplex method~\citep{avriel2003nonlinear}, initialized from a coarse grid search and warm-started from the previous budget's solution. Fig.~\ref{fig:law_runs_frontier}
shows the resulting frontier and the RL share $\rho_{\mathrm{RL}}$ along it. Because our law is fit locally, we restrict the extrapolation to the range of compute we actually observe. Within this range, the global frontier predicted by the law closely tracks the empirical frontier of measured runs, confirming that the fitted law faithfully recovers the compute-optimal trade-off. Along this frontier, the RL-optimal compute share increases from ${\sim}20\%$ at 50M to ${\sim}28\%$ at 680M.

This suggests that in the lower-compute regime, additional pretraining remains the more valuable use of compute, whereas as the total compute budget grows, RL should receive a proportionally larger share of the allocation.

\subsection{Limitations}
\label{app:limitations_law_fitting}

Several limitations are important for interpreting the fitted law.

First, the law assumes that RL reward is approximately linear in $\log_{10}C_{\mathrm{RL}}$ over the measured compute range.  This is an empirical local approximation.  It should not be interpreted as evidence that RL improvement is unbounded.

Second, the strict extrapolation setting compounds two sources of error: prediction error in the Chinchilla loss surface and residual error in the $f$ and $g$ maps.  The observed-$L$ LOO setting shows that the latter is smaller, but the practical setting requires predicting $L$ from $(N,T)$.

Third, $g(N,T)$ is materially noisier than $f(L)$.  The reference reward is strongly controlled by pretraining loss, whereas the RL slope has additional unexplained variation.  This may reflect optimizer details, SFT/RL data differences, reward-model variation, or other run-level factors not included in the current parameterization.

Finally, the data are densest near the observed model sizes, token counts, and RL-compute range.  The frontier analysis is therefore best viewed as a diagnostic for allocation trends under the fitted assumptions, not as a claim that the same exponents hold arbitrarily far beyond the training distribution.

\section{Move Policy Evolution}
\label{sec:appendix_policy_evolution}
\subsection{From Token Space to Move Space: Policy Evolution Metrics}

Although training is performed in token space, all policy-evolution metrics are defined in move space. This is possible because each valid token prefix corresponds to a legal board state, and each legal move at that state has a token serialization.

\paragraph{Induced move policies.}
For pretraining, define the raw move score
$\widetilde{\pi}_{\theta_{\mathrm{pre}}}(a\mid s)=\prod_{j=1}^{|\tau(a)|} p_{\theta_{\mathrm{pre}}}(x^{(a)}_j\mid s,x^{(a)}_{<j})$
and the induced move policy
$\pi_{\theta_{\mathrm{pre}}}(a\mid s)=\frac{\widetilde{\pi}_{\theta_{\mathrm{pre}}}(a\mid s)}{\sum_{a'\in\mathcal{A}(s)}\widetilde{\pi}_{\theta_{\mathrm{pre}}}(a'\mid s)}$.

For post-training stages $m\in\{\mathrm{sft},\mathrm{rl}\}$, the trajectory policy $\pi_{\theta_m}(\zeta\mid s)=\pi_{\theta_m}(r,a_1,o_1,\dots,a_T\mid s)$ induces a root-conditioned trace policy $\pi_{\theta_m}(r\mid s_0)$. Conditional on a fixed trace $r$, define the raw move score
$\widetilde{\pi}_{\theta_m}(a\mid s,r)=\prod_{j=1}^{|\tau(a)|} p_{\theta_m}(x^{(a)}_j\mid s,r,x^{(a)}_{<j})$
and the conditional move policy
$\pi_{\theta_m}(a\mid s,r)=\frac{\widetilde{\pi}_{\theta_m}(a\mid s,r)}{\sum_{a'\in\mathcal{A}(s)}\widetilde{\pi}_{\theta_m}(a'\mid s,r)}$.
Marginalizing over traces gives the root-conditioned marginal move policy
$\pi_{\theta_m}(a\mid s)=\sum_r \pi_{\theta_m}(r\mid s)\,\pi_{\theta_m}(a\mid s,r)$. Since enumerating all possible reasoning traces is intractable, we estimate the marginalized move policy by Monte Carlo sampling in practice. For each prompt, we sample $K=128$ rollouts from the trace policy $\pi_{\theta_m}(r\mid s_0)$ and approximate the marginal as
\begin{align*}
\widehat{\pi}_{\theta_m}(a\mid s)
=
\frac{1}{K}\sum_{k=1}^K
\pi_{\theta_m}(a\mid s,r^{(k)}),
\qquad r^{(k)}\sim \pi_{\theta_m}(r\mid s_0).
\end{align*}

\subsection{Fitting Power-Sharpening Transformations}
\label{app:alpha_power_fitting}

We use two complementary diagnostics to test whether RL primarily transforms the SFT marginal policy by simple probability sharpening. 
For a state $s$, let
\[
p_s(a)=\pi_{\mathrm{sft}}(a\mid s),
\qquad
q_s(a)=\pi_{\mathrm{rl}}(a\mid s),
\]
where $a\in\mathcal A(s)$ denotes a legal move. 
Given a coefficient $\alpha$, define the $\alpha$-power transform of the SFT policy as
\[
p_{s,\alpha}(a)
=
\frac{p_s(a)^\alpha}{\sum_{b\in\mathcal A(s)} p_s(b)^\alpha}.
\]
When $\alpha>1$, this transform sharpens the SFT distribution by increasing relative mass on high-probability moves; when $\alpha<1$, it flattens the distribution.

\paragraph{KL power fit.}
The first diagnostic fits a single global sharpening coefficient by projecting the RL policy onto the SFT power family. 
Specifically, we choose
\[
\alpha^\star
=
\arg\min_{\alpha\in[0,\alpha_{\max}]}
\sum_s w_s \,
D_{\mathrm{KL}}\!\left(q_s \,\|\, p_{s,\alpha}\right),
\]
where $w_s$ is a state weight. 
By default, we use uniform weighting, $w_s=1/N$ for $N$ states.

The fitted value $\alpha^\star$ measures the degree of distribution-level sharpening. 
If $\alpha^\star>1$, the best power approximation sharpens the SFT policy toward the RL policy. 
The remaining divergence measures how much of the RL update is not explained by simple power sharpening. 
We summarize fit quality using
\[
\mathrm{ExplainedSharp}
=
1-
\frac{
\sum_s w_s \,
D_{\mathrm{JS}}\!\left(q_s, p_{s,\alpha^\star}\right)
}{
\sum_s w_s \,
D_{\mathrm{JS}}\!\left(q_s, p_s\right)
}.
\]
A high value of $\mathrm{ExplainedSharp}$ indicates that the RL policy is well approximated by a power-sharpened SFT policy. 
A low value indicates that RL reshapes the distribution in ways not captured by a uniform sharpening transform.

\paragraph{Centered-logit linear fit.}
The second diagnostic tests the logit-geometry implied by power sharpening. 
Taking logs of the power transform shows that, up to a state-dependent normalization constant, power sharpening scales centered log-probabilities linearly. 
For each state, define
\[
x_s(a)
=
\log p_s(a)
-
\frac{1}{|\mathcal A(s)|}
\sum_{b\in\mathcal A(s)}
\log p_s(b),
\]
and
\[
y_s(a)
=
\log q_s(a)
-
\frac{1}{|\mathcal A(s)|}
\sum_{b\in\mathcal A(s)}
\log q_s(b).
\]
We then fit the linear relation
\[
y_s(a) \approx \beta x_s(a)
\]
across states and actions, and report the fitted slope $\beta$ and coefficient of determination $R^2$. 
A slope $\beta>1$ indicates sharpening in centered log-probability space, while the $R^2$ measures how well the RL policy is explained by a pure scaling of the SFT logits.

We fit the results on the 50M model pretrained with 4.6B tokens. Table~\ref{tab:alpha_fitting_statistics} and Table~\ref{tab:beta_fitting_statistics} show the fitting results. Both $\alpha^*$ and $\beta$ tend to increase during RL, though their distributions reveal substantial sample-level heterogeneity.

\begin{table}[t]
\centering
\small
\caption{\textbf{Estimated global $\alpha$ and $\alpha$ per state.} $\alpha^*$ appears to increase across RL stages, but the widening IQR suggests substantial variability across samples.}
\begin{tabular}{lcccc}
\toprule
Stage
& $\alpha^*$ (global)
& $\alpha^*$ (median [IQR])
& ExplainedSharp (global)
& ExplainedSharp (median [IQR]) \\
\midrule
pretrain $\to$ SFT
& 0.57 & 0.63 [0.44, 0.88] & -0.01 & 0.09 [-0.01, 0.26] \\
SFT $\to$ RL\_50
& 1.05 & 1.12 [0.91, 1.41] & 0.03 & 0.16 [0.02, 0.54] \\
SFT $\to$ RL\_100
& 1.15 & 1.25 [0.98, 1.64] & 0.09 & 0.24 [0.03, 0.68] \\
SFT $\to$ RL\_250
& 1.27 & 1.47 [1.07, 2.06] & 0.16 & 0.42 [0.07, 0.84] \\
SFT $\to$ RL\_500
& 1.27 & 1.52 [1.03, 2.39] & 0.14 & 0.39 [0.04, 0.88] \\
SFT $\to$ RL\_750
& 1.35 & 1.72 [1.14, 2.81] & 0.17 & 0.49 [0.08, 0.93] \\
\bottomrule
\end{tabular}
\label{tab:alpha_fitting_statistics}
\end{table}

\begin{table}[t]
\centering
\small
\caption{\textbf{Estimated global $\beta$ and $\beta$ per state. }}
\begin{tabular}{lcccc}
\toprule
Stage
& $\beta$ (global)
& $\beta$ (median [IQR])
& $R^2$ (global)
& $R^2$ (median [IQR]) \\
\midrule
pretrain $\to$ SFT
& 0.60 & 0.62 [0.47, 0.76] & 0.40 & 0.44 [0.30, 0.58] \\
SFT $\to$ RL\_50
& 0.99 & 1.03 [0.89, 1.16] & 0.68 & 0.75 [0.62, 0.85] \\
SFT $\to$ RL\_100
& 1.03 & 1.07 [0.92, 1.21] & 0.63 & 0.71 [0.57, 0.81] \\
SFT $\to$ RL\_250
& 1.07 & 1.12 [0.94, 1.29] & 0.60 & 0.68 [0.53, 0.80] \\
SFT $\to$ RL\_500
& 1.06 & 1.12 [0.94, 1.29] & 0.57 & 0.65 [0.49, 0.77] \\
SFT $\to$ RL\_750
& 1.13 & 1.18 [0.99, 1.38] & 0.56 & 0.63 [0.48, 0.76] \\
\bottomrule
\end{tabular}
\label{tab:beta_fitting_statistics}
\end{table}

\begin{figure}[htbp]
    \centering
    \includegraphics[width=\linewidth]{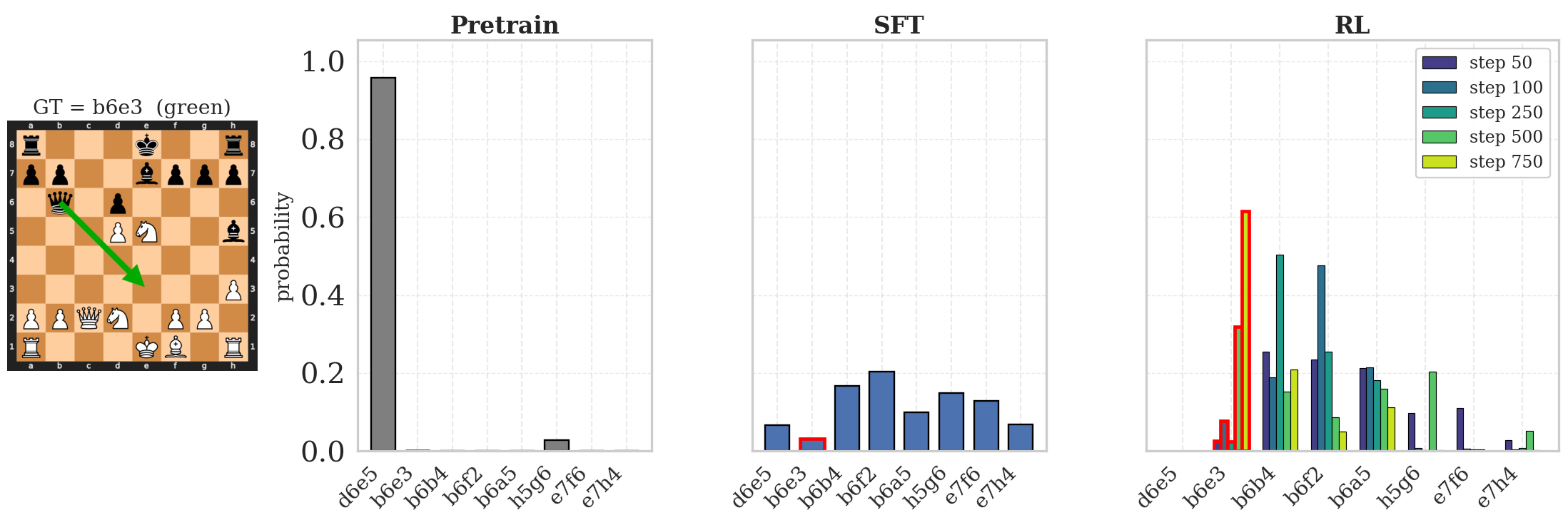}
    \caption{\textbf{Policy evolution on a hard puzzle (B5) across 
training stages.} Left: board position with ground-truth move 
b6e3 (green arrow). Pretrain: the model assigns ${\sim}95\%$ 
mass to a wrong move (d6e5), with b6e3 nearly absent. SFT: 
probability spreads across many candidates but b6e3 (red) 
remains low. RL: over training steps 50--750, b6e3 rises to 
become the top move (tail discovery), while a competing wrong 
move (b6b4) also retains significant mass (wrong-mode 
amplification).
}
    \label{fig:puzzle_example}
\end{figure}

\subsection{Policy Categorization}
\label{app:policy_evolution}

Table~\ref{tab:policy_update_taxonomy} gives the formal definitions for all categories we consider. Intuitively, the categories distinguish the following update types. 
\textit{Ground-truth amplification} captures cases where the correct move is already top-ranked and is further reinforced. 
\textit{Tail discovery} captures cases where training promotes a correct move from the low-probability tail into the top-$k$ set. 
\textit{Top-$k$ correction} captures promotion of a correct move that was initially plausible but not top-ranked. 
\textit{Ground-truth regression} captures cases where a previously top-$k$ correct move is demoted. 
\textit{Wrong-mode amplification} captures cases where the correct move remains outside the top-$k$ set while the initially preferred wrong move is further reinforced. 
All remaining transitions are grouped as \textit{Other}. We set $\epsilon_{\mathrm{tail}}$ as 0.05. We fit the results on the 50M model pretrained with 4.6B tokens. 

\begin{table}[t]
\centering
\footnotesize
\caption{
Taxonomy of policy-update effects based on changes in top-$k$ membership and probability between the initial policy $\pi_{\theta_0}$ and trained policy $\pi_{\theta_1}$.
}
\setlength{\tabcolsep}{4pt}
\renewcommand{\arraystretch}{1.25}
\begin{tabularx}{\linewidth}{@{}
>{\raggedright\arraybackslash}p{0.22\linewidth}
>{\raggedright\arraybackslash}X
>{\raggedright\arraybackslash}p{0.43\linewidth}
@{}}
\toprule
\textbf{Category} & \textbf{Description} & \textbf{Condition} \\
\midrule

\textit{Ground-truth amplification}
&
The ground-truth move remains in the top-$k$ set and gains probability.
&
$\begin{aligned}[t]
a^\star(s) \in \mathcal T_{\theta_0}^k(s), a^\star(s) \in \mathcal T_{\theta_1}^k(s), \Delta p(a^\star;s) > 0.
\end{aligned}$ \\

\midrule

\textit{Tail discovery}
&
A low-probability ground-truth move is promoted into the top-$k$ set.
&
$\begin{aligned}[t]
& a^\star(s) \notin \mathcal T_{\theta_0}^k(s), a^\star(s) \in \mathcal T_{\theta_1}^k(s), \\
& \pi_{\theta_0}(a^\star(s)\mid s) < \epsilon_{\mathrm{tail}}.
\end{aligned}$ \\

\midrule

\textit{Top-$k$ correction}
&
A non-tail ground-truth move is promoted into the top-$k$ set.
&
$\begin{aligned}[t]
&a^\star(s) \notin \mathcal T_{\theta_0}^k(s), a^\star(s) \in \mathcal T_{\theta_1}^k(s),\\ 
&\pi_{\theta_0}(a^\star(s)\mid s) \ge \epsilon_{\mathrm{tail}}.
\end{aligned}$ \\

\midrule

\textit{Ground-truth regression}
&
A previously top-$k$ ground-truth move is demoted out of the top-$k$ set.
&
$\begin{aligned}[t]
a^\star(s) &\in \mathcal T_{\theta_0}^k(s), \quad a^\star(s) \notin \mathcal T_{\theta_1}^k(s).
\end{aligned}$ \\

\midrule

\textit{Wrong-mode amplification}
&
The ground-truth move remains outside the top-$k$ set, while the initial top-1 wrong move is reinforced.
&
$\begin{aligned}[t]
&w_{\theta_0}(s) = \arg\max_a \pi_{\theta_0}(a\mid s), \\
&a^\star(s) \notin \mathcal T_{\theta_0}^k(s), a^\star(s) \notin \mathcal T_{\theta_1}^k(s), \\
&w_{\theta_0}(s) \in \mathcal T_{\theta_1}^k(s), \quad \Delta p(w_{\theta_0};s) > 0.
\end{aligned}$ \\

\midrule

\textit{Other}
&
All remaining cases, including stable top-$k$ structure, switches among wrong modes, and partial changes that do not promote the ground-truth move into the top-$k$ set.
&
Otherwise. \\

\bottomrule
\end{tabularx}
\label{tab:policy_update_taxonomy}
\end{table}

\subsection{CoT Evolution Analysis}
\label{app:cot_evolution}

As a complementary lens, we examine how the structure of reasoning traces evolves over the course of RL training. Because our CoT format comprises explicit move sequences, each rollout can be reconstructed as a prefix tree rooted at the puzzle state (Section~\ref{sec:sft}), with each node corresponding to a move, enabling us to probe both the structure and quality of the model's reasoning. 

We characterize properties of reasoning traces along three axes: (1) \emph{Search shape} defines structural properties of the parsed tree: number of nodes, maximum depth $D$, average branching factor, and the width-to-depth ratio $|L|/D$, where $|L|$ is the number of leaves. (2) \emph{Move quality} scores candidate moves against Stockfish via the normalized rank $(r-1)/(n-1)$, where $r \in \{1, \ldots, n\}$ is the Stockfish rank of the move (with $r=1$ best) among the $n \geq 2$ legal moves at that position; lower values indicate stronger moves. We report this metric separately for the player's first-move candidates and the model's proposed opponent replies. (3) \emph{Search behavior} captures traversal patterns by measuring the consistency with DFS and fractions of revisiting nodes. All metrics are computed per rollout, averaged across rollouts within the same prompt, and then aggregated across prompts.
\begin{figure}
    \centering
    \includegraphics[width=1.0\linewidth]{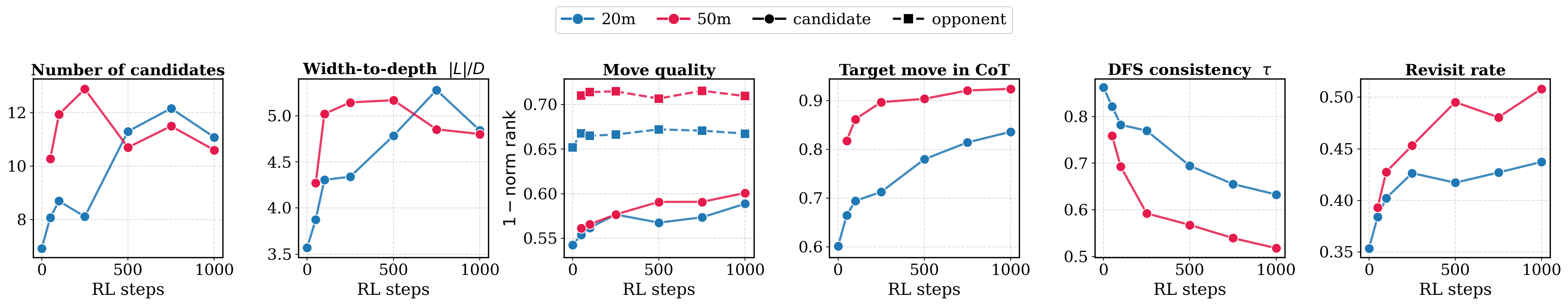}
    \caption{ \textbf{Structured reasoning traces evolve during RL training.}
For the 20M and 50M models, panels report per-step prompt means on the puzzle benchmark. RL modestly increases search breadth, improves move quality scored by Stockfish rank for both model moves (solid) and predicted opponent replies (dashed), and increases ground-truth move coverage in the parsed CoT tree. DFS consistency decreases and revisit rate increases, indicating more re-exploration of earlier prefixes.
}
    \label{fig:cot_evolution}
\end{figure}

\begin{figure}
    \centering
    \includegraphics[width=0.8\linewidth]{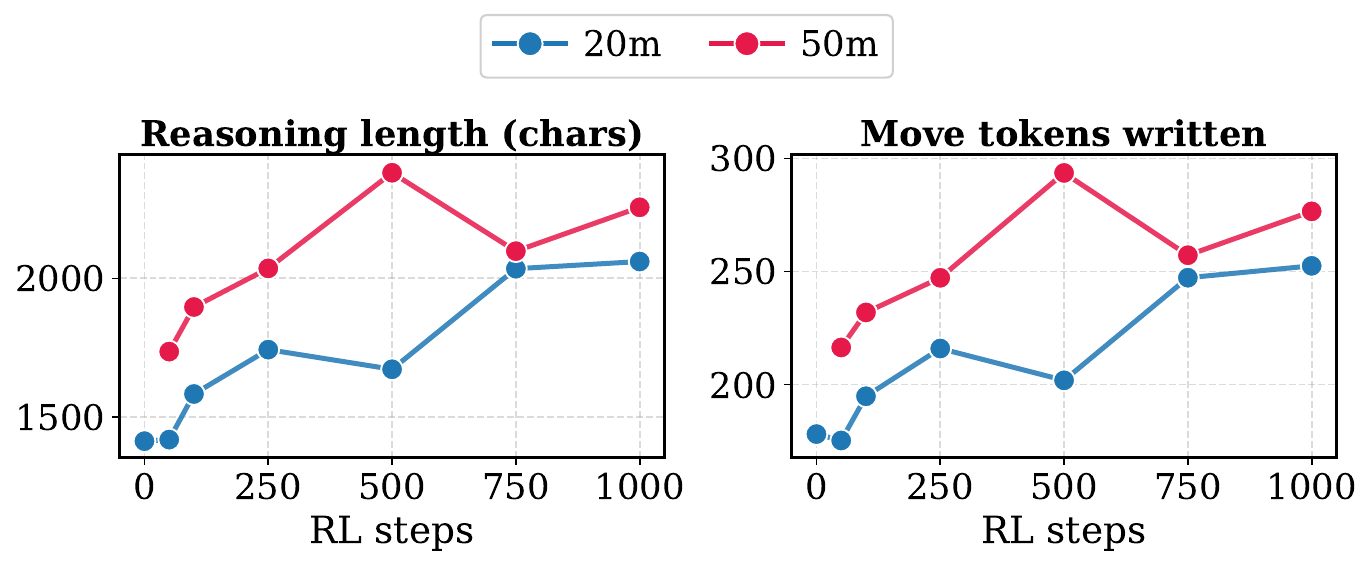}
    \caption{\textbf{Reasoning length dynamics over RL training.}}
    \label{fig:cot_length_app}
\end{figure}

\textbf{Structured CoT exposes fine-grained search dynamics and weakness in deeper search.} In Fig.~\ref{fig:cot_evolution}, we compare two representative RL runs for the 20M and 50M models pretrained under matched compute. Additional metrics are shown in Fig.~\ref{fig:cot_structure_app} and Fig.~\ref{fig:cot_target_continuation_app} in Appendix~\ref{app:cot_evolution}. These structured-CoT metrics provide a finer-grained view of search behavior than reasoning-token counts alone (Fig.~\ref{fig:cot_length_app}). The parsed search traces show that, on average, the models primarily expand search breadth rather than depth, as the width-to-depth ratio and branching factor increase while maximum search depth remains roughly flat. The 20M model tends to propose more distinct candidate moves over training, which may also help explain the improvement in its pass@$k$. Meanwhile, the quality of moves proposed in the CoT improves for both the model's own moves and its predicted opponent responses, with larger gains on the model-move side. The model also becomes more likely to mention the ground-truth move in its CoT and to commit to the best candidate it has considered. Interestingly, the generated search traces become less aligned with a strict DFS serialization order, showing more revisits to previously considered lines. Despite these improvements, Fig.~\ref{fig:cot_target_continuation_app} reveals that the model still struggles to recover continuations that require deeper search, suggesting that current RL training may improve candidate generation and selection faster than it improves long-horizon search. 
Understanding how these structured search features affect performance may help guide future SFT data filtering and construction toward examples that encourage deeper, more systematic search.

\begin{figure}
    \centering
    \includegraphics[width=1.0\linewidth]{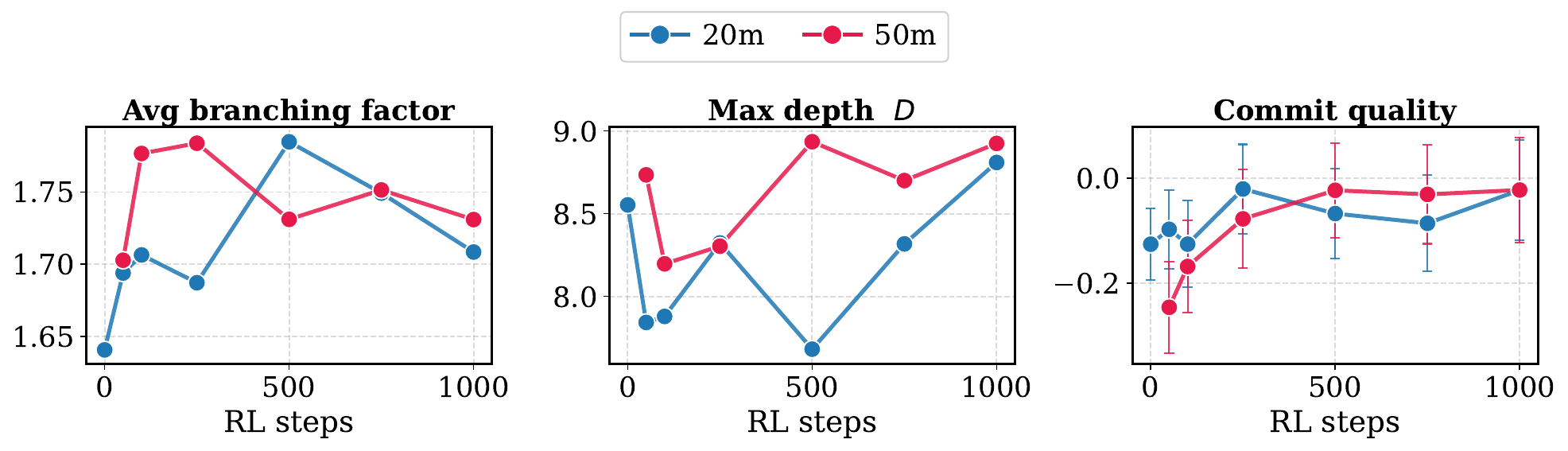}
    \caption{\textbf{Search shape and selected move quality over RL training for the 20M and 50M models}. Curves show mean metrics for the 20M and 50M models, with standard error across prompts for the selected moves quality. Columns report effective branching factor $b_{\mathrm{eff}} = N^{1/D}$, maximum tree depth $D$, and commit quality, a per-prompt z-score measuring the Stockfish rank of the committed move relative to both the best move and the best considered move. Branching increases modestly, depth decreases slightly, and commit quality improves monotonically.}
    \label{fig:cot_structure_app}
\end{figure}

\begin{figure}
    \centering
    \includegraphics[width=1.0\linewidth]{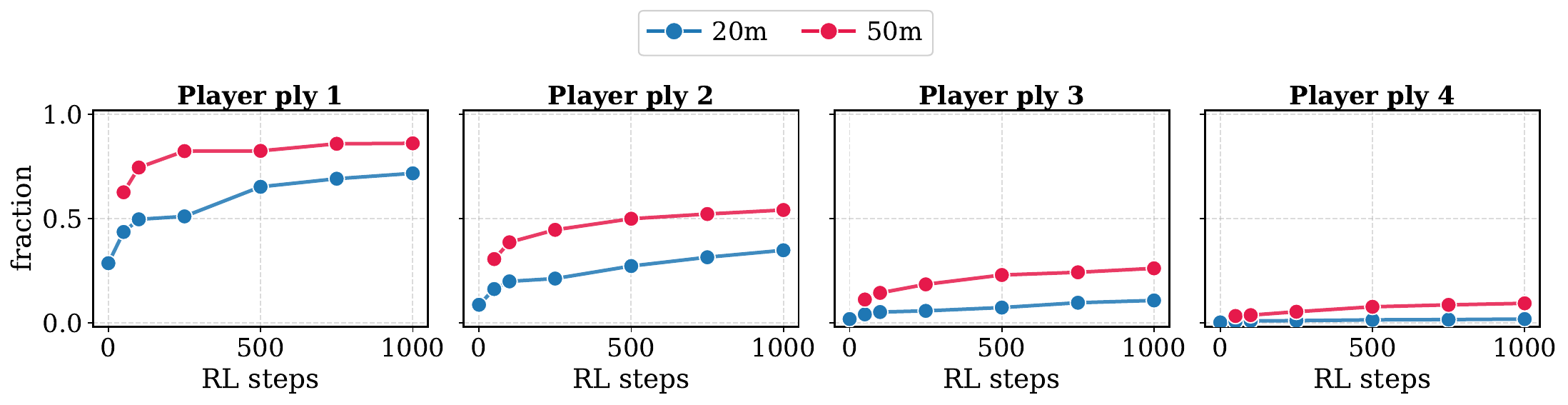}
    \caption{\textbf{Search-tree coverage of ground-truth continuations by depth for the 20M and 50M models.} For each rollout, we measure whether the parsed CoT search tree contains a root-to-depth-$k$ path whose player moves match the first $k$ target plies, allowing arbitrary legal opponent replies. Panels report coverage for $k=1,2,3,4$, excluding rollouts with shorter targets. RL improves coverage at all depths, but gains decay sharply with depth; even at step 1000 for the 50M model, few rollouts recover the full 4-ply target line.}
    \label{fig:cot_target_continuation_app}
\end{figure}

\subsection{Entropy}

We report policy entropy over the induced move distribution and token entropy over CoT tokens below (for the 50M model pretrained with 4.6B tokens). Policy entropy drops substantially (1.370 → 0.705), while CoT token entropy decreases modestly (0.536 → 0.492), indicating RL concentrates move selection with a smaller reduction in token-level variability. This concentration is heterogeneous across states: the policy-entropy distribution remains broad, and harder B3–B5 puzzles retain higher entropy than easier B1–B2 puzzles. These results support our conclusion that RL does not sharpen uniformly across states.

\begin{figure}
    \centering
    \includegraphics[width=\linewidth]{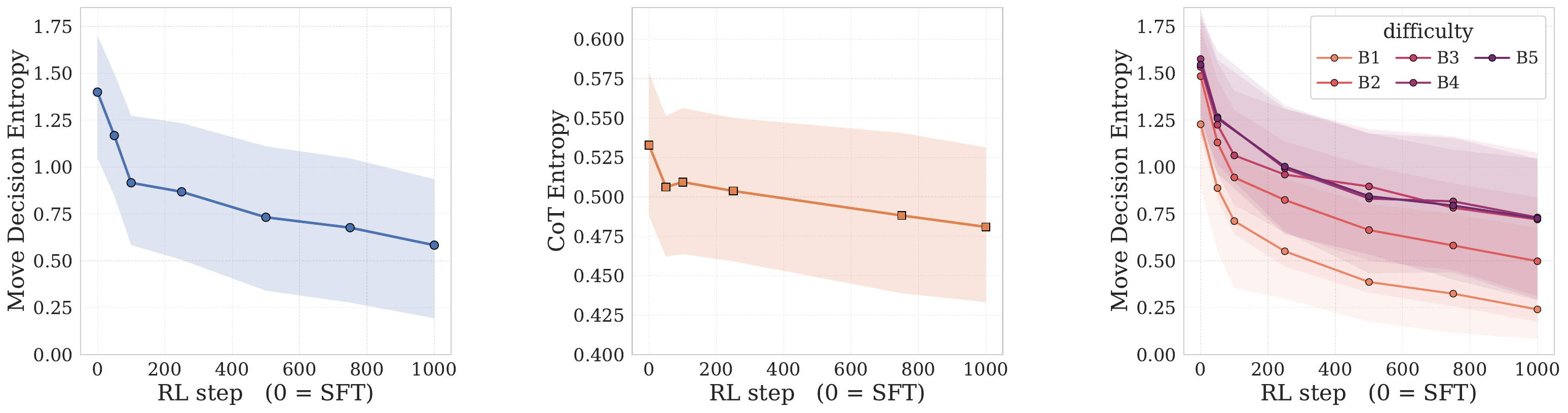}
    \caption{\textbf{Entropy dynamics during RL training.}}
    \label{fig:entropy}
\end{figure}

\section{Olmo Experiment Additional Details}
\label{app:olmo}

\subsection{Implementation Details}
We pretrain a 1B-parameter OLMo-2 language model, fork intermediate
checkpoints as pretraining-scale anchors, anneal each to convergence,
supervised fine-tune on math reasoning traces, and finally run GRPO
reinforcement learning. Architecture is fixed across all runs; the only
variable across anchors is the number of pretraining tokens.
Tables~\ref{tab:arch} and~\ref{tab:hparams} summarize the setup.

\begin{table}[ht]
\centering
\small
\caption{Model architecture. Identical for every pretraining anchor,
supervised fine-tuning, and reinforcement-learning run.}
\begin{tabular}{@{}ll@{}}
\toprule
\textbf{Architecture (OLMo-2)} & \textbf{Value} \\
\midrule
Total parameters            & 1.48\,B \\
Non-embedding parameters    & 1.07\,B \\
Layers                      & 16 \\
Hidden size                 & 2048 \\
FFN intermediate size       & 8192 \\
Attention heads             & 16 \\
KV heads (no GQA)           & 16 \\
Vocabulary size             & 100{,}278 \\
Max position embeddings     & 8192 \\
Positional encoding         & RoPE ($\theta=10^4$) \\
Tied input/output embeddings & No \\
Precision                   & bf16 \\
\bottomrule
\end{tabular}
\label{tab:arch}
\end{table}

\begin{table}[t]
\centering
\small
\setlength{\tabcolsep}{5pt}
\caption{Hyperparameters for each stage of the pipeline. Pretraining uses a
warmup--stable--decay (WSD) schedule; intermediate stable-phase checkpoints
(every 5{,}000--10{,}000 steps) define the pretraining-scale anchors. Each
anchor is annealed independently (linear LR decay to zero over 5\,B tokens)
before supervised fine-tuning and reinforcement learning. RL uses GRPO
(no KL penalty, entropy coefficient 0, dual-clip ratios
$0.2/0.26$, $c=10$) with 8 rollouts per prompt at temperature 1.0.}
\begin{tabular}{@{}lllll@{}}
\toprule
& \textbf{Pretrain (stable)} & \textbf{Anneal} & \textbf{SFT} & \textbf{RL (GRPO)} \\
\midrule
Data
  & Dolma3/Dolmino mix & (same corpus) & NuminaMath-CoT & GSM8K + MATH + DeepScaler mix \\
Examples / tokens
  & 200\,B tokens & 5\,B tokens / anchor & 859{,}490 ex.\ ($\approx$0.46\,B tok) & up to 3000 steps \\
Sequence length
  & 4096 & 4096 & 4096 (cutoff) & 512 prompt / 3584 resp. \\
Global batch
  & 512 seq ($\approx$2.1\,M tok) & 512 seq & 512 examples & 128 prompts $\times$ 8 \\
Optimizer
  & AdamW & AdamW & AdamW & AdamW \\
$(\beta_1,\beta_2)$
  & (0.9, 0.95) & (0.9, 0.95) & --- & --- \\
Weight decay
  & 0.033 (0 on emb.) & 0.033 & --- & --- \\
Peak LR
  & $4\times10^{-4}$ & $4\times10^{-4}\!\to\!0$ & $1\times10^{-5}$ & $1\times10^{-6}$ \\
LR schedule
  & WSD (warmup+const) & linear decay to 0 & cosine & constant \\
Warmup
  & 2\,B tokens & --- & 3\% of steps & 50 steps \\
Epochs / steps
  & 95{,}368 steps & --- & 1 epoch (1679 steps) & 3000 steps \\
Packing
  & --- & --- & No (one ex.\ / seq) & --- \\
Loss mask
  & all tokens & all tokens & assistant only & --- \\
\bottomrule
\end{tabular}
\label{tab:hparams}
\end{table}

\subsection{Additional Results}

\begin{figure}
    \centering
    \includegraphics[width=1.0\linewidth]{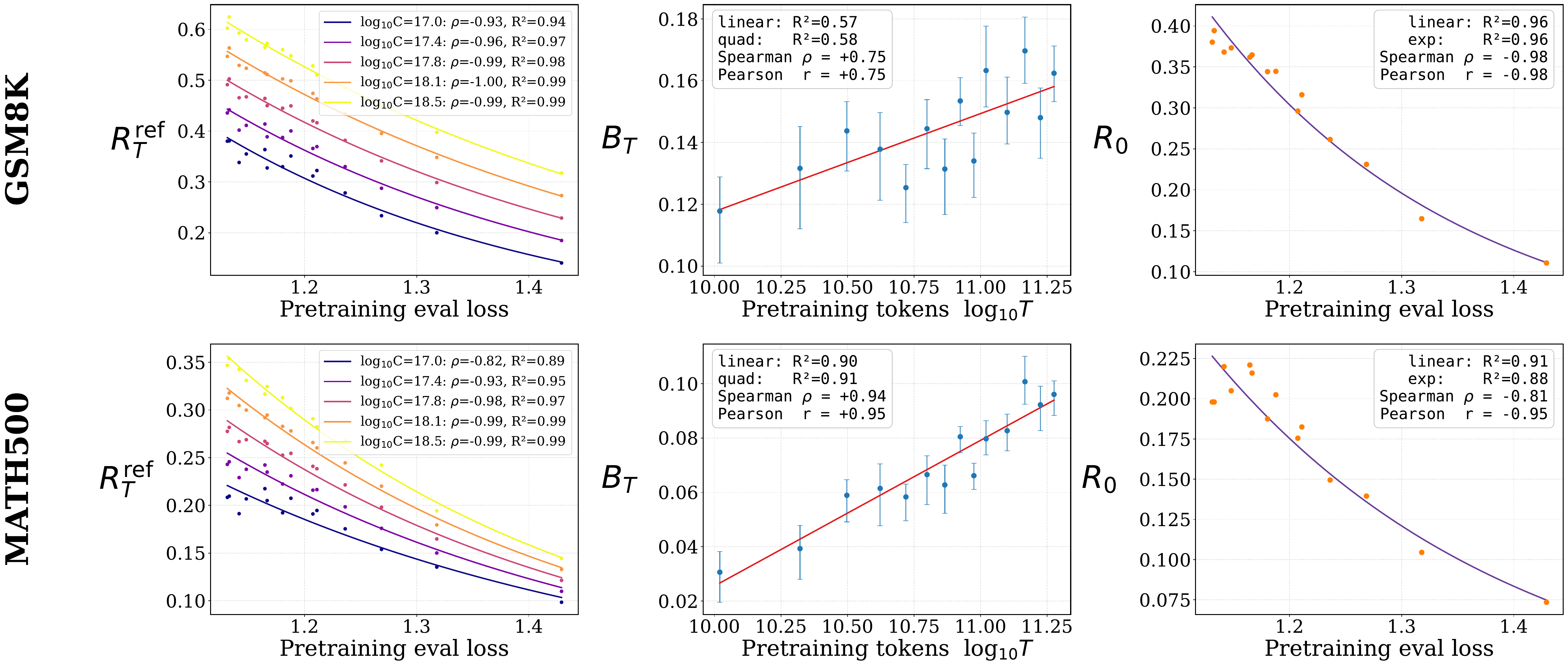}
    \caption{\textbf{RL local performance and slope with pretraining properties on
GSM8K and MATH500 for pretrained 1B models.}}
    \label{fig:gsm_math_law_fitting}
\end{figure}

Fig.~\ref{fig:gsm_math_law_fitting} reports the fitting results on downstream benchmarks.

\end{document}